\documentclass[
12pt, 
english, 
onehalfspacing, 
headsepline, 
]{MastersDoctoralThesis} 

\usepackage[utf8]{inputenc} 
\usepackage[T1]{fontenc} 
\usepackage{amsmath}
\usepackage{mathpazo} 
\usepackage{algpseudocode,algorithm,algorithmicx} 
\newcommand*\Let[2]{\State #1 $\gets$ #2}
\algrenewcommand\algorithmicrequire{\textbf{Precondition:}}
\algrenewcommand\algorithmicensure{\textbf{Postcondition:}}
\usepackage{euscript}
\usepackage{subfig}
\usepackage{longtable}
\usepackage[most]{tcolorbox}
\usepackage{caption}
\usepackage{array}
\usepackage{bm}
\usepackage{marginnote}
\usepackage{tikz}
\usetikzlibrary{positioning,angles,quotes,calc,arrows,arrows.meta,positioning, decorations.pathreplacing}
\usepackage{url}

\usepackage{mathtools}
\usepackage{makecell}

\usepackage{soul}

\newcommand*{\rom}[1]{\expandafter\@slowromancap\romannumeral #1@}

\newcommand{\bea}{\begin{eqnarray}}
\newcommand{\eea}{\end{eqnarray}}
\newcommand{\bc}{\begin{center}}
\newcommand{\ec}{\end{center}}

\newcolumntype{M}[1]{>{\centering\arraybackslash}m{#1}}
\newcolumntype{C}{>{\centering\arraybackslash}m{8em}}

\algnewcommand\algorithmicinput{\textbf{Input:}}
\algnewcommand\algorithmicoutput{\textbf{Output:}}
\algnewcommand\Input{\item[\algorithmicinput]}%
\algnewcommand\Output{\item[\algorithmicoutput]}%

\usepackage[autostyle=true]{csquotes} 

\thesistitle{Deep Reinforcement Learning for Complex Manipulation Tasks with Sparse Feedback} 
\supervisor{Dr. Armin \textsc{Biess}} 
\examiner{} 
\degree{Master of Industrial Engineering and Management} 
\author{Binyamin \textsc{Manela}} 
\addresses{} 

\subject{Deep Reinforcement Learning} 
\university{\href{http://in.bgu.ac.il/Pages/default.aspx}{Ben-Gurion University of the Negev}} 
\department{\href{http://department.university.com}{Department of Industrial Engineering and Manegment}} 
\group{\href{http://researchgroup.university.com}{Robotics and Machine Intelligence Lab}} 
\faculty{\href{http://in.bgu.ac.il/en/engn/Pages/default.aspx}{Faculty of Engineering Sciences}} 

\AtBeginDocument{
\hypersetup{pdftitle=\ttitle} 
\hypersetup{pdfauthor=\authorname} 
\hypersetup{pdfkeywords=\keywordnames} 
}

\begin{document}

\frontmatter 

\pagestyle{plain} 


\begin{titlepage}
\begin{center}

\vspace*{.06\textheight}
{\scshape\LARGE \univname\par}\vspace{1.5cm} 
\textsc{\Large Thesis}\\[0.5cm] 

\HRule \\[0.4cm] 
{\huge \bfseries \ttitle\par}\vspace{0.3cm} 
\HRule \\[1.5cm] 
 
\begin{minipage}[t]{0.4\textwidth}
\begin{flushleft} \large
\emph{Author:}\\
{\authorname} 
\end{flushleft}
\end{minipage}
\begin{minipage}[t]{0.4\textwidth}
\begin{flushright} \large
\emph{Advisor:} \\
{\supname} 
\end{flushright}
\end{minipage}\\[2.5cm]
 

\large \textit{A thesis submitted in fulfillment of the requirements\\ for the degree of \degreename}\\[0.3cm] 
\textit{in the}\\[0.3cm]
\deptname\\[1.5cm] 
 

{\large \today}\\[4cm] 

\vfill
\end{center}
\end{titlepage}


\begin{declaration}
\addchaptertocentry{\authorshipname} 
\noindent I, \authorname, declare that this thesis titled, \enquote{\ttitle} and the work presented in it are my own. I confirm that:

\begin{itemize} 
\item This work was done wholly or mainly while in candidature for a research degree at this University.
\item Where any part of this thesis has previously been submitted for a degree or any other qualification at this University or any other institution, this has been clearly stated.
\item Where I have consulted the published work of others, this is always clearly attributed.
\item Where I have quoted from the work of others, the source is always given. With the exception of such quotations, this thesis is entirely my own work.
\item I have acknowledged all main sources of help.
\item Where the thesis is based on work done by myself jointly with others, I have made clear exactly what was done by others and what I have contributed myself.\\
\end{itemize}
 
\noindent Signed:\\
\rule[0.5em]{25em}{0.5pt} 
 
\noindent Date:\\
\rule[0.5em]{25em}{0.5pt} 
\end{declaration}

\cleardoublepage


\vspace*{0.2\textheight}

\noindent\enquote{\itshape Thanks to my solid academic training, today I can write hundreds of words on virtually any topic without possessing a shred of information, which is how I got a good job in journalism.}\bigbreak

\hfill Dave Barry


\begin{abstract}
\addchaptertocentry{\abstractname} 
Learning optimal policies from sparse feedback is a known challenge in reinforcement learning. Hindsight Experience Replay (HER) is a multi-goal reinforcement learning algorithm that comes to solve such tasks. The algorithm treats every failure as a success for an alternative (virtual) goal that has been achieved in the episode and then generalizes from that virtual goal to real goals. HER has known flaws and is limited to relatively simple tasks. In this thesis, we present three algorithms based on the existing HER algorithm that improves its performances. First, we prioritize virtual goals from which the agent will learn more valuable information. We call this property the \textit{instructiveness} of the virtual goal and define it by a heuristic measure, which expresses how well the agent will be able to generalize from that virtual goal to actual goals. Secondly, we reduce existing bias in HER by the removal of misleading samples during learning. Lastly, we enable the learning of complex, sequential, tasks using a form of curriculum learning combined with HER. To test our algorithms, we built three challenging manipulation environments with sparse reward functions. Each environment has three levels of complexity. Our empirical results show vast improvement in the final success rate and sample efficiency when compared to the original HER algorithm.
\end{abstract}


\begin{acknowledgements}
\addchaptertocentry{\acknowledgementname} 
This research was supported in part by the Helmsley Charitable Trust through the Agricultural, Biological and Cognitive Robotics Initiative and by the Marcus Endowment Fund both at the Ben-Gurion University of the Negev. This research was supported by the ISRAEL SCIENCE FOUNDATION (grant no. 1627/17).\\
I also want to thank my advisor Dr. Armin Biess who helped me throughout the entire process and was there to provide useful and insightful advises any time needed.
\end{acknowledgements}


\hypersetup{linkcolor=black}
\setcounter{tocdepth}{1}

\tableofcontents 

\listoffigures 

\listoftables 


\begin{abbreviations}{ll} 
	
\textbf{MDP} & Markov Decision Process \\
\textbf{DP} & Dynamic Programming \\
\textbf{ML} & Machine Learning \\
\textbf{ANN} & Artificial Neural Network \\
\textbf{DRL} & Deep Reinforcement Learning \\
\textbf{HER} & Hindsight Experience Replay \\

\end{abbreviations}


\begin{symbols}{ll} 

$S$ & state space \\
$A$ & action space \\
$V(s)$ & estimated value of state s \\
$V_{real}(s)$ & real value of state s \\
$Q(s,a)$ & estimated value of action a at state s \\
$Q_{real}(s,a)$ & real value of action a at state s \\
$r(s)$ & reward for state s \\
\addlinespace 

$\pi$ & policy \\
$\alpha$ & learning rate \\
$\lambda$ & discount factor \\

$\Psi$ & multi layered task\\
$\psi$ & sub-task in a multi layered task\\

\end{symbols}


\dedicatory{Dedicated to my wife, Hilit, how bravely tolerated all my craziness and been there for me when things did not go as expected. After all those long hours you spent trying to look like you are interested in my research, I genuinely think your name should be on the title page, next to mine.} 


\mainmatter 

\pagestyle{thesis} 



\chapter{Introduction} 

\label{introduction} 


\paragraph{}One central vision of robotics is to ease human life by automating repetitive and daily processes \cite{mayor2018gods}. Robots can nowadays perform a variety of tasks from washing the floor \cite{fujiwara1995floor,chang2011floor}, to building a car \cite{mortimer2003mix} or playing soccer \cite{kitano1998robocup,muller2007making}. Teaching a robot new skills by manually programming is a long and exhausting process that requires significant manpower and time.\\
In recent years machine learning algorithms have increasingly been used for robot skill learning. The most commonly used type of algorithms is \emph{supervised learning} due to its relatively simple implementation and robust performances. For example, behavioral cloning has been used to transfer various skills from expert demonstrations to a robot, such as driving a car \cite{zhang2018behavioral}, helicopter aerobatics \cite{abbeel2010autonomous}, or robot ball paddling \cite{kober2009learning}. Nevertheless, supervised learning has several drawback as it requires, first, a lot of demonstrations, and second, limits the ability to surpass the demonstrator performances.\\
A different machine learning approach towards robot skill learning is \textit{reinforcement learning}, in which an agent learns through trial and error by interacting with the environment \cite{sutton2018reinforcement}. Deep reinforcement learning, the combination of reinforcement learning with deep learning \cite{goodfellow2016deep} has led to many breakthroughs in recent years in generating goal-directed behavior in artificial agents ranging from playing Atari games without prior knowledge and human guidance \cite{mnih2013playing}, to teaching an animated humanoid agent to walk \cite{todorov2012mujoco, lillicrap2015continuous, schulman2017proximal}, and defeating the best GO player in the world \cite{silver2018general}, just to name a few. All reinforcement learning problems are based on the {\it{reward hypothesis}}, stating that any goal-directed task can be formulated in terms of a reward function. The reward function needs to be informative and guide the agent towards the optimal policy. However, the engineering of such a reward function is often challenging. The difficulties in shaping suitable dense reward functions limit the application of reinforcement learning to real-world tasks, particularly in robotics \cite{kober2013reinforcement}. One way to overcome the problem of reward shaping has been presented in Hindsight Experience Replay (HER) \cite{andrychowicz2017hindsight}, which uses sparse reward signals to indicate whether a task has been completed or not. The algorithm uses failures to learn how to achieve alternative goals that have been achieved in the episode (virtual goals) and uses the latter to generalize to actual goals.

\section{Problem Statement}
Although HER is considered to be the state-of-the-art algorithm for manipulation tasks with sparse reward function, it is still not able to efficiently learn complex tasks. When sampling virtual goals from failures, HER does not consider which are most instructive for the agent but instead randomly samples from future states. Secondly, using HER induces bias, which may hinder the learning process. Lastly, HER cannot be applied for sequential manipulation tasks, which significantly limits its practical application.

\section{Research Objective}
This thesis is about enabling manipulators to learn new challenging skills from sparse feedback using deep reinforcement learning algorithms.
We aim to overcome the limitations of HER by introducing three novel algorithms based on HER. As we will show, these modifications lead to vast improvement over the vanilla algorithm on a variety of challenging manipulation tasks. In this thesis we will focus on throwing tasks because they provide a challenging test bed for reinforcement learning using sparse feedback and consist of a sequence of sub-tasks, such as picking a ball and throwing it, and thus, cannot be solved using the vanilla-HER algorithm.

\section{Thesis Structure}
The thesis is organized as follows. In chapter 2, the background of our work is provided. Chapter 3 describes the simulations we designed and built to test our algorithms. In chapters 4,5 and 6, we present our novel algorithms. Chapter 4 introduces our algorithm for improving the virtual-goals generation strategy of the vanilla-HER and show its performances. Chapter 5 introduces our algorithm for reducing the bias induced by the vanilla-HER and show its performances. In chapter 6, we introduce our algorithm that extends the vanilla-HER algorithm for sequential manipulation tasks and show its performances. 
Finally, in chapter 7, we conclude this thesis by summarizing our work and findings and proposing possible directions for future work.



\chapter{Background} 

\label{background} 


\newcommand{\keyword}[1]{\textbf{#1}}
\newcommand{\tabhead}[1]{\textbf{#1}}
\newcommand{\code}[1]{\texttt{#1}}
\newcommand{\file}[1]{\texttt{\bfseries#1}}
\newcommand{\option}[1]{\texttt{\itshape#1}}


In this chapter we will introduce the theoretical background for reinforcement learning and deep reinforcement learning. We will first introduce the general theory of traditional reinforcement learning with emphasis on Q-learning and multi-goals tasks. We will then introduce artificial neural networks and deep learning. We will describe the basic elements of artificial neural networks and their extension to deep learning. Finally, we will introduce deep reinforcement learning, which is the combination of reinforcement learning and deep learning. In particular, we will focus on DRL from sparse reward functions.


\section{Traditional Reinforcement Learning} \label{s:traditional_reinforcement_learning}
\begin{figure}[th]
\centering
\includegraphics{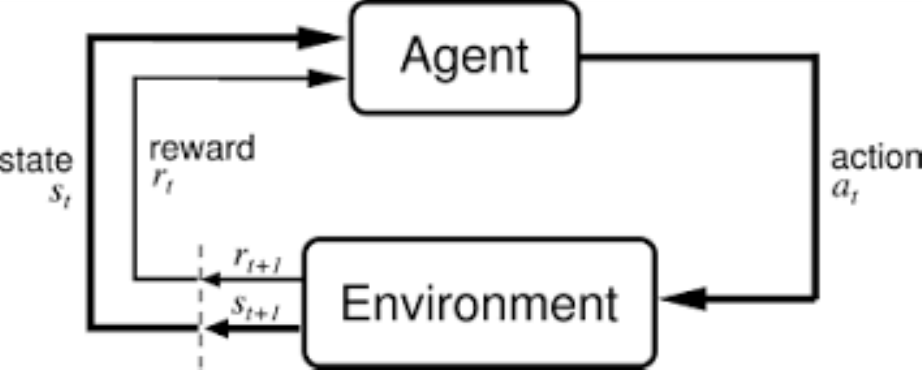}
\decoRule
\caption[RL diagram]{The reinforcement learning diagram. The agent receives the state of the environment, chooses an action accordingly, and gets back the new state of the environment and a reward which indicates how good was this action\footnotemark.}
\label{fig:RL_diagram}
\end{figure}

In reinforcement learning (RL) an agent tries to solve a task by trial and error through interaction with an environment whose dynamics are unknown to the agent. The agent can change the state of the environment by its actions while receiving immediate feedback from the environment. The objective of the agent is to to solve the task by finding an optimal chain of actions. \\
Although reinforcement learning is an area within machine learning, it is fundamentally different from standard machine learning methods (supervised or unsupervised) in several aspects. First, reinforcement learning does not depend on data acquisition. Instead, in reinforcement learning the agent learns from its own experience created during the interaction with the environment and does not depend on a supervisor. Secondly, reinforcement learning focuses on finding an optimal policy rather than analyze data. 
Reinforcement learning can be described by the diagram in Fig \ref{fig:RL_diagram}.\\
Reinforcement learning is usually modeled as a Markov Decision Process.
\footnotetext{Source: \quad\url{https://i.stack.imgur.com/eoeSq.png}}
\paragraph{Markov Decision Processes:} \label{ss:markov decision process}
The Markov decision processes (\emph{MDPs}) is a stochastic mathematical model for a decision-making scenario. At each step, the decision-maker (or \emph{agent}) chooses an action, and the outcomes are partly random and partly as a result of the action \cite{bellman1957markovian}. MDPs are used for modeling a variety of optimization problems and solved via dynamic programming (\emph{DP}) and reinforcement learning (\emph{RL}).\\
A Markov decision process is defined by 5 elements, $\langle $S$,$A$,$P$,$R$,\gamma \rangle$, where
\begin{itemize}
\item \textbf{S} is a set of states
\item \textbf{A} is a set of actions
\item \textbf{$P(s,a,s')$}$=Pr(s_{t+1}=s'|s_t=s, a_t=a)$ is the transition matrix which gives the probability that action $a$ in state $s$ at time $t$ will lead to state $s’$ at time $t+1$
\item \textbf{$R(s,a,s')$} is the reward (or expected reward) the agent received after applying action $a$ at state $s$ and getting to state $s'$.\\ alternatively, $R(s)$ is the reward received after entering state $s$. 
\item \textbf{$\gamma$} is the discount factor, which represents the difference in importance between short- and long term rewards.
\end{itemize}

The problem in \emph{MDPs} is to find a "policy" for the decision-maker: a function $\pi$ that maps states to actions $a=\pi(s)$. The policy can be either deterministic or stochastic. Once a Markov decision process is combined with a policy in this way, it fixes the action for each state, and the resulting combination behaves as a Markov Reward Process.\\

The goal is to find a policy $\pi$ that will maximize the expected discounted sum of rewards from every state $s_t$ onwards (also known as the \textit{return} $G_t$):
\begin{equation}
G_t = \sum_{i=t}^{\infty}\gamma^{i}\cdot R(s_i,a_i,s_{i+1})
\end{equation}

where $a_i=\pi(s_i)$ and $\gamma$ is the discount factor and satisfies $0\leq\gamma\leq1$. \\

Now that we have defined our goal and what defines an optimal behavior, we can start introducing methods to find that behavior.
For the cases in which the dynamics of the \emph{MDP} and the \emph{reward functions} are known (\textbf{model-based}), this problem can be solved using dynamic programming.
For the cases in which the dynamics of the \emph{MDP} or the \emph{reward function} is not available (\textbf{model-free}), there is a need for Reinforcement-Learning methods.\\
Reinforcement learning methods can be divided into two main categories: \textit{Value-Function based algorithms} and \textit{Policy-based algorithms}

\paragraph{Value-Function based algorithms} \label{ss:Value-Function based algorithms}
The value-function $V^\pi(s)$ assigns values to states. The value of state $s$ under policy $\pi$ is, by definition, the expected return $G_t$ from state $s$ onwards, following policy $\pi$ and defined by Bellman's equation as:
\begin{equation} \label{eq:value-function}
\begin{aligned}
V^\pi(s) &= \mathop{\mathbb{E}}[G_t | S_t=s] \\
&= \sum_{a\in A}\pi(a|s)\sum_{s'\in S}P(s'|s,a)[R(s,a,s')+\gamma V^\pi(s')]
\end{aligned}
\end{equation} 
Similarly, the value of action $a_t$ in state $s_t$ (or $Q(s_t,a_t)$) is the expected return $G_t$ after executing action $a_t$ in state $s_t$ and defined by Bellman's equation as:
\begin{equation}
\begin{aligned}
Q^\pi(s,a) &= \mathop{\mathbb{E}}[G_t | S_t=s, A_t=a] \\
&= \sum_{s' \in S} P(s'|s,a)[R(s,a,s')+\gamma\sum_{a'\in A} \pi(a'|s')Q^\pi (s',a')]
\end{aligned}
\end{equation}
An optimal policy $\pi^*$ is such that the value of any state $s$ under $\pi^*$ is greater or equal to the value of state $s$ under any other policy $\pi'$ for all $s\in S$
\begin{equation}
V^{\pi^*}(s) = V^*(s) \geq V^{\pi'}(s) \quad \forall\quad s\in S,\, \pi'
\end{equation}
In value-function based algorithms we try to learn the value of all states $s\in S$ and actions $a\in A$. Then, we can derive $\pi^*$ from the $Q$ function.\\
The core of value-function based algorithms is to update the Q function parameters $\theta$ iteratively so that the squared error $\delta$ between the estimated Q value and the real Q value decreases. 
Since the real Q value is unknown, it is estimated using the experience.\\
The best practice to evaluate the Q value is given by \emph{temporal difference methods} (or \textit{TD}).
The TD evaluation of Q value uses the immediate reward and the evaluation of the next state: $\tilde{Q}(s,a) = r + \gamma\hat{Q}(s',a^*)$ \\
and $\delta = \hat{Q}(s,a)-\tilde{Q}(s,a) = \hat{Q}(s,a)-r-\gamma\hat{Q}(s',a^*)$, where $\hat{Q}$ denotes the value in the table and $\tilde{Q}$ denotes the estimated target value.\\
The optimization process can be formalized as follows:
\begin{equation}
\theta_{i+1} = \theta_{i} + \Delta\theta_i
\end{equation}
when $\Delta\theta_i$ is the update to the parameters and is learned from $\delta$.

\paragraph{Policy-based algorithms} \label{ss:Policy based algorithms}
Policy search methods work in a more direct approach. Instead of finding the value of each possible state and then derive the optimal policy, policy-based methods seeking to directly find a policy $\pi$ that maximizes the expected return $G$.\\
The core of policy-based algorithms is to iteratively update the policy parameters $\theta$, so that the expected return increases. The optimization process can be formalized as follows:
\begin{equation}
\theta_{i+1} = \theta_{i} + \Delta\theta_i
\end{equation}

\paragraph{Exploration vs Exploitation}\label{ss:Exploration vs Exploitation}
While learning the environment, the agent can apply two strategies:
\begin{itemize}
\item Exploration: choose a random action - by following this approach the agent can visit new states and find new, better policies.
\item Exploitation: act greedily – get high total rewards for the task by using known best actions according to existing knowledge.
\end{itemize}

The agent should use exploration when it is not certain that its knowledge is right and exploitation when it is confident that its estimate of $Q(s_t,a_t)$ is close to the real value. \\

If the agent will only do exploration, it may not achieve high scores at the task and may not improve its actions. On the other hand, if it will only apply exploitation, it may get stuck in its current policy not seeing all possible trajectories. Hence, the agent will probably miss the optimal policy. Thus, there must be a fine balance between exploitation and exploration. \\
The most popular combination of exploration and exploitation is an \emph{$\epsilon$-greedy} policy. In an $\epsilon$-greedy policy, a single parameter $\epsilon$ between 0 and 1 ($0\leq \epsilon \leq 1$) controls what fraction of the time the agent deviates from greedy behaviour. Each time the agent selects an action, it chooses probabilistically between exploration and exploitation. With probability $\epsilon$ it explores by selecting randomly from all the available actions and with probability $1-\epsilon$ it exploits by selecting the greedy action.

\begin{equation}
a = \begin{cases}
rand(a_n)       & rand(0,1) \leq \epsilon \\
argmax_a Q      & otherwise \\
\end{cases}
\end{equation}

High values of $\epsilon$ will force the agent to explore more frequently and - as a result - will reduce the probability of taking optimal action, while giving the agent the ability to react rapidly to changes that take place in the environment. Low values of $\epsilon$ will drive the agent to exploit more optimal actions. Often the value $\epsilon$ in an episode is chosen as a decreasing function of time, where $\epsilon \rightarrow 0$ for $t \rightarrow \infty$. For such a choice the agent acts more and more greedily over time. \\

Another method to choose actions is the Boltzmann Distribution (Also known as Gibbs Distribution or softmax) policy.
Boltzmann Distribution (BD) is a learning policy that reduces the tendency for exploration with time. It is based on the assumption that the current model improves as learning progresses. BD assigns a probability to any possible action according to its expected utility and according to a parameter \textit{$T$} called temperature. \\

BD assigns a positive probability for any possible action $a\in A$ using the following Eq.
\begin{equation}
P(a \mid s) = \frac{e^{\frac{Q(s,a)}{T}}}{\sum_{a' \in A} e^{\frac{Q(s,a')}{T}}}
\end{equation}
where
\begin{equation}
T_{new} = e^{-dj} * T_{max} + 1
\end{equation}

Action with high $Q(s,a)$ are associated with higher probability $P$. $T$ decreases as iteration $j$ increases over time. Therefore, as learning progresses, the exploration tendency of the agents reduces and a \emph{BD} learning policy will tend to exploit actions with high $Q(s,a)$. The parameters $T_{max}$ and decay rate $\emph{dj}$ are set at the start. \\

The advantage of using a Boltzmann distribution for action selection is that it produces a stochastic policy. It is commonly used as a way of inducing variability in the behavior that is tied to the action- Q function, and thus, to the actions themselves. It is also used as a model for human decision making.

\paragraph{Multi-Goal environments}\label{ss:Multi-Goal environments}
The standard MDP model can be extended to multi-goal scenarios. That is that the goal changes in every episode. The MDP is then augmented by a set of goals \textbf{G} from which a goal is sampled at the beginning of each episode. For those tasks, the agent needs to consider the current goal while evaluating a state or choosing an action. Thus, for those environments, the notation of the value-function, the $Q$ function and the policy is extendend to $V(s,g)$, $Q(s,a,g)$ and $\pi(s,g)$, respectively

\subsection{RL Algorithms} \label{ss:rl algorithms}
\textbf{Q learning}, also known as \emph{temporal difference method} (TD), is a value-function based algorithm \cite{watkins1992q}. Nowadays, many deep reinforcement learning algorithms are based on Q learning. For example, the algorithm Deep Q Network uses a variation of Q learning with a neural network as a function approximator. Temporal difference methods use the TD evaluation of the Q value from paragraph \ref{ss:Value-Function based algorithms} to update the estimated Q function. The update rule is as follows:
\begin{equation}
Q(s,a) = Q(s,a) - \alpha\cdot(Q(s,a) - r - \gamma Q(s',a'^*))
\end{equation}
\begin{algorithm}[H]
\caption{Q-learning}
\label{alg:Q-learning}
\begin{algorithmic}[1]
	\Require{$\alpha$ - learning-rate, $\gamma$ - discount factor, $\lambda$ - decay rate}
	\Statex Initialize $Q(s,a)$ to 0 $\forall s,a$
	\For {each episode}
	\Let{$s$}{$s_0$}		
	\While{$s$ in not terminal}
	\Let{$a$}{$\pi(s)$} \Comment{according to the policy (e.g. epsilon-greedy)}
	\State play action $a$ and get reward $r$ and next state $s'$ 
	\Let{$a'^*$}{$max_{\tilde{a}}(Q(s',\tilde{a}))$}
	\Let{$\delta$}{$Q(s,a) - r - \gamma Q(s',a'^{*})$} \Comment{error}
	\Let{$Q$}{$Q-\alpha*\delta$} \Comment{update Q}
	\Let{$s$}{$s'$} \Comment{update state}
	\EndWhile
	\EndFor
\end{algorithmic}	
\end{algorithm}

\textbf{Universal Value Function Approximators} or \textit{UVFA} is an extension of the standard Q learning algorithm in which the Q function and the policy is also dependent on the current goal. \\
The update rule changes to:
\begin{equation}
Q(s,a,g) = Q(s,a,g) - \alpha*(Q(s,a,g) - r - \gamma Q(s',a'^*,g))
\end{equation}
\newpage

\section{Artificial Neural Networks} \label{s:artefitial_neural_network}
Machine learning (ML) systems are used for many tasks, including image classification, text translation from one language to another, weather forecasting, and more. Increasingly, machine learning algorithms make use of \emph{deep learning} methods, which rely on deep neural networks. Deep networks can find an efficient representation of intricate patterns within multidimensional data (like images and text). Deep learning has been a breakthrough in many fields, including supervised learning, reinforcement learning, and more.\\

\begin{figure}[th]
\centering
\includegraphics[height=7cm]{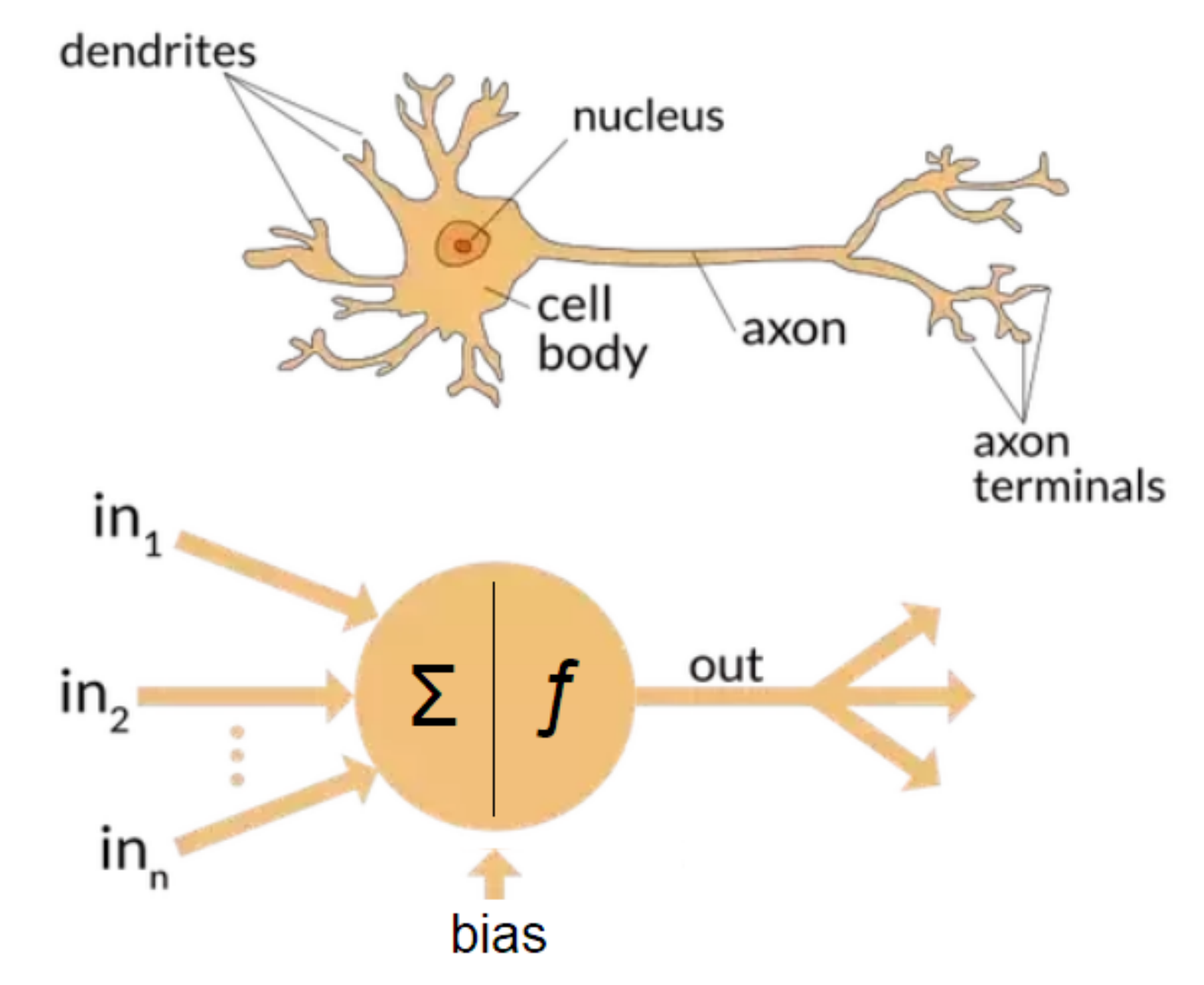}
\decoRule
\caption[neuron diagram]{The biological and artificial neurons. Both neurons receive multiple inputs, process them, and shoot if a certain threshold is crossed\footnotemark.}
\label{fig:neuron_diagram}
\end{figure}

Artificial neural networks (ANN) are a set of algorithms designed to work similarly like the biological brain. The human brain contains, roughly, 100 billion neurons used to process input from the environment such as vision, hearing, smelling, and more. Every single neuron has several inputs coming from other neurons. The neuron processes the input's activities and if a certain threshold is reached, the neuron fires through its single output to all the neurons to which it is connected.\\
The artificial neuron (will be called a \textit{neuron} from now on) is roughly simulating the biological neuron. Each neuron receives input from several neurons. the input is processed using a weighted-sum, adding a bias term and passing through an activation function (or \textit{non-linearity}): $output=f(b+\sum^n\theta_ix_i)$. Figure \ref{fig:neuron_diagram} shows the biological and artificial neurons.
\footnotetext{Source: \quad \url{https://www.quora.com/What-is-the-difference-between-artificial-intelligence-and-neural-networks}}
The artificial neural network (will be called a \textit{neural network} from now on) approximates nonlinear functions using many neurons in a chain-like structure. The weight parameters $\theta$ are, usually, found using an iterative process of small updates such that the network's performances as an approximation function are maximized. The process of finding these parameters is called \emph{learning}. \\

\begin{figure}[th]
\centering
\includegraphics[height=5cm]{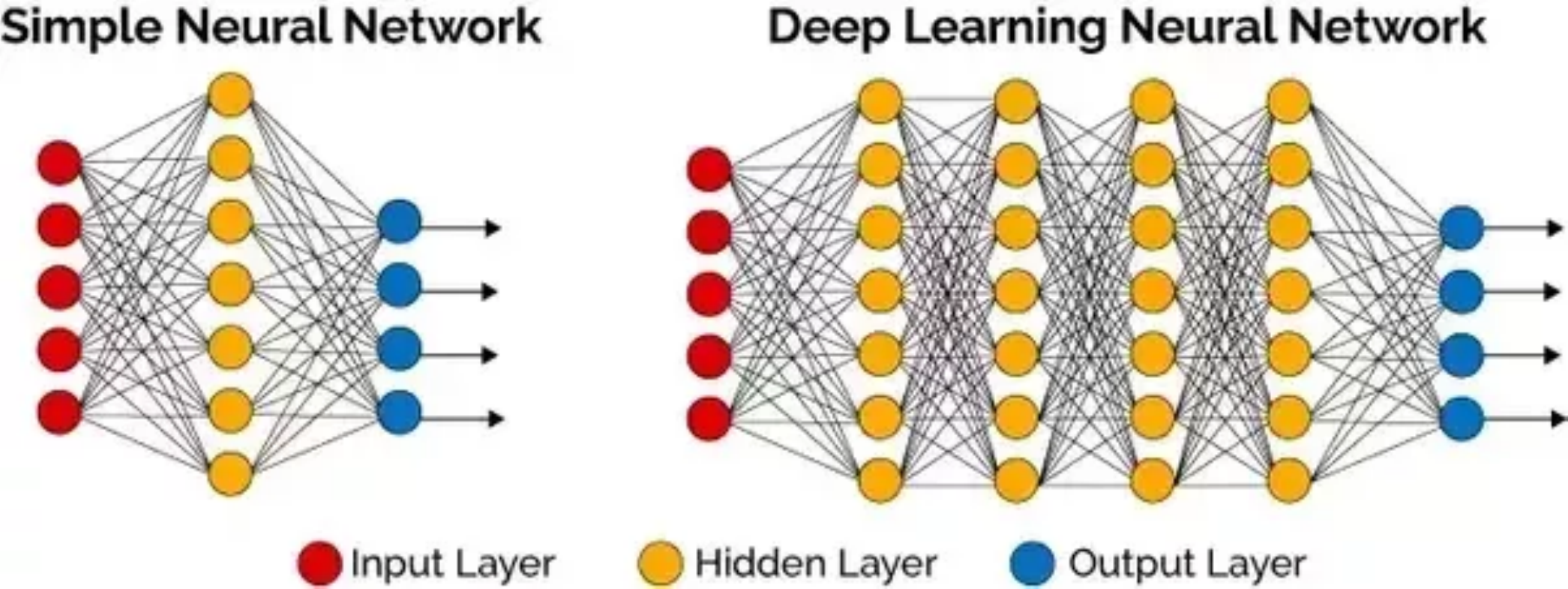}
\decoRule
\caption[FC network]{The fully-connected network. A deep neural network is a neural network with more than one hidden layer. These networks are fully connected since each neuron is connected to all the neurons in the previous layer\footnotemark.}
\label{fig:FC network}
\end{figure}

The simplest form of a neural network is called \emph{multilayer perceptron} (MLP) and consists of several, fully-connected, layers of neurons (Figure \ref{fig:FC network}). This network contains an \textit{input} layer that receives the raw input from the environment, an \textit{output} layer that returns the network's outputs, and one or more hidden layers of neurons in which all the processing is performed. This network is also called a fully-connected network, since each neuron in layer $i$ gets as input the output of all the neurons in layer $i-1$.
\footnotetext{Source: \quad\url{https://www.quora.com/What-is-the-difference-between-Neural-Networks-and-Deep-Learning}}
\subsection{Learning Process} \label{ss:Learning Process}
The goal in the learning process is to find a set of parameters $\theta$ which leads to the best performances. In supervised learning, the real target value $Y$ is known for a given set of inputs. By using the ground-truth values, the network can update the parameters $\theta$ such that the performances on the given set is maximized.\\
The learning process is iterative and includes the following steps:
\begin{enumerate}
\item \textbf{Feed-forward:} The input $x$ is fed to the network and resulting in the network's prediction $\hat{Y}$ of the corresponding $Y$ value. $\hat{Y}=f(x,\theta)$. 
\item \textbf{Loss:} The loss $L(\theta)$ is computed by comparing the predicted value $\hat{Y}$ to the target value $Y$. Different tasks require different loss functions. In our work, we use one of the most commonly used loss function - the mean-squared-error (MSE). The MSE loss function is used, mostly, for regression problems. It calculates the L2 distance between the predicted value $\hat{Y}$ to the target value $Y$.
\begin{equation}
MSE(\hat{Y}, Y) = \frac{\sum_{i=1}^n(\hat{Y}^T-Y)^2}{n} = \frac{||\hat{Y}^T-Y||}{n}
\end{equation} 
The loss function can also be an objective that we like to minimize with no comparison to a given label. 
\item \textbf{Back-propagation:} In order to minimize the loss, we use the back-propagation algorithm, introduced by \cite{rumelhart1988learning}. The back-propagation algorithm uses the chain rule to calculate the local gradient of the loss to all hidden neurons. The chain rule is used to calculate the derivatives of composed functions by multiplying local derivatives. Given the functions $y=f(x)$ and $z=g(f(x))$, the derivative $\frac{\partial z}{\partial x}$ can be computed according to equation \ref{eq:chain rule}.
\begin{equation}
\label{eq:chain rule}
\frac{\partial z}{\partial x} = \frac{\partial z}{\partial y}\times\frac{\partial y}{\partial x}
\end{equation} 
The gradient $\frac{\partial loss}{\partial \theta}$ is propagated back through the network using the chain rule in the opposite direction of the forward pass. Figure \ref{fig:back propagation} shows a single neuron $z=f(x,y,\theta)$. Its local derivatives are $\frac{\partial z}{\partial x}$ and $\frac{\partial z}{\partial y}$ and can be determined during the forward pass. The local gradient of the loss is computed during back-propagation by multiplying the local derivatives with the local gradient over the neuron at the next layer. In case a neuron is connected to several neurons in
the next layer, the gradients are added up.
\begin{figure}[th]
	\centering
	\includegraphics[height=5cm]{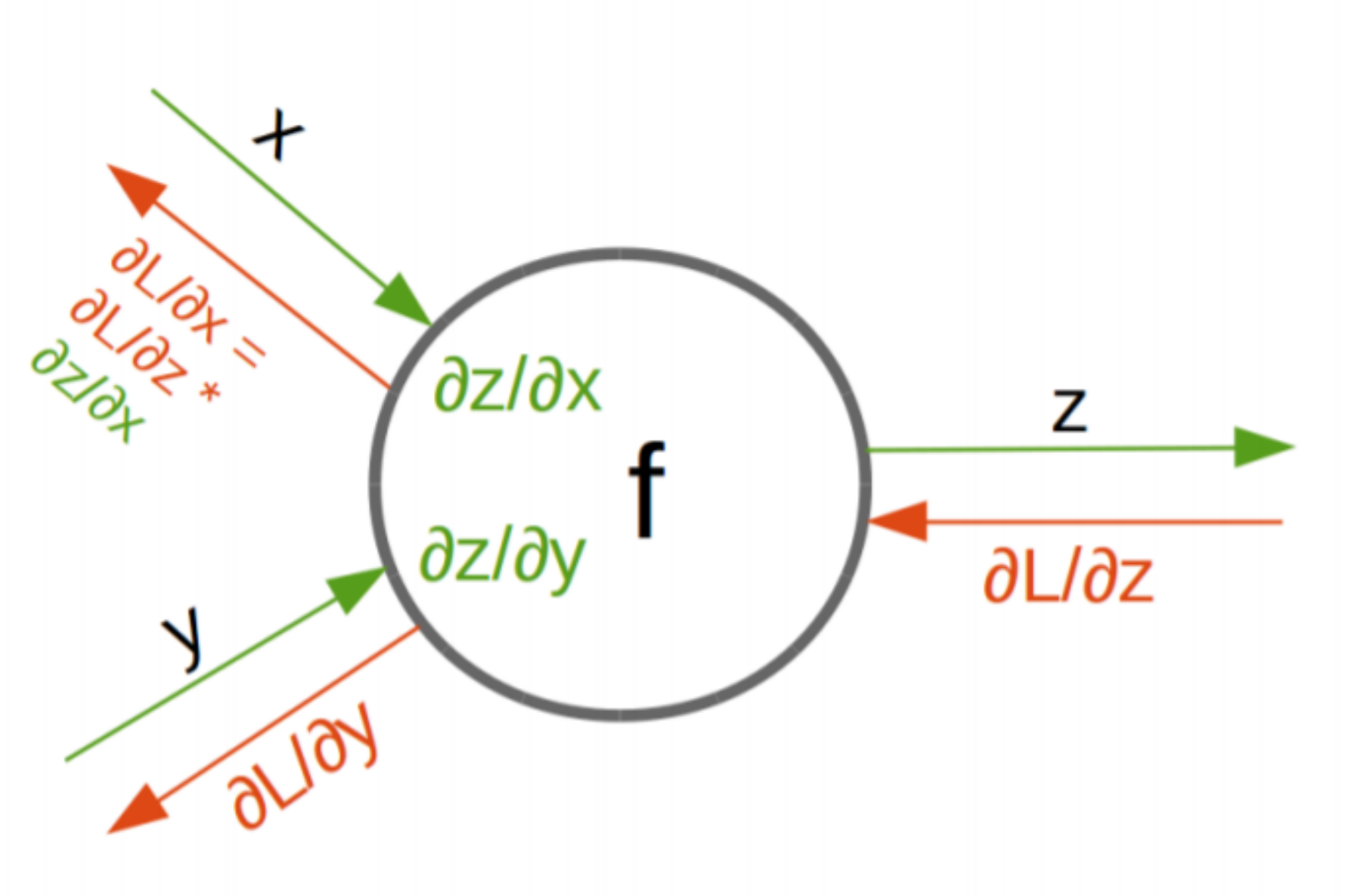}
	\decoRule
	\caption[back propagation]{The back propagation process through a single neuron. The local derivatives can be determined during the forward pass. the local gradient of the loss over the parameters $x$ and $y$ is the multiplication of the local gradient of the next neuron with the local derivatives\footnotemark.}
	\label{fig:back propagation}
\end{figure}
\footnotetext{Source: \quad\url{https://becominghuman.ai/back-propagation-in-convolutional-neural-networks-intuition-and-code-714ef1c38199}}
\item \textbf{Update:} After calculating the gradient, all the parameters are updated accordingly. A commonly used optimizer is stochastic gradient descent (SGD). It calculates the gradient for a random batch of samples and applies one or more gradient descent steps on the parameters. Gradient descent slightly updates the parameters toward the opposite direction of the gradient to minimize the loss, and thus, to reach a local optimum. Calculating the gradient on a batch of sample stabilize the gradient steps and prevent over-fitting. Equation \ref{eq:SGD update} shows the update rule of SGD. The parameter $\alpha$ determines the step size and is predefined. If $\alpha$ is too small, the learning process may take a long time to reach a local minimum. On the other hand, if $\alpha$ is too big, there is a risk that a local minimum will never be reached	 because the taken steps will be too big and overshoot.
\begin{equation}
\label{eq:SGD update}
\theta_{i+1} = \theta_i + \alpha\nabla_{\theta_i} L(\theta_i)
\end{equation}
\end{enumerate}

\subsection{Activation Functions} \label{ss:Activation Functions}
In order to approximate nonlinear function, there needs to be an activation function (also called \textit{non-linearities}) between the layers. The activation functions simulate the threshold of [within the] biological neurons (as showed in figure \ref{fig:neuron_diagram}).\\
Three of the most commonly used activation functions are \emph{Sigmoid}, \emph{Tanh}, and \emph{ReLU}.\\

The Sigmoid function is shown in equation \ref{eq:sigmoid}. This function is a strictly increasing function that maps each $x$ value to a number in the range $(0,1)$. Big negative numbers are mapped to numbers very close to zero, zero is mapped to $0.5$, and big positive numbers are mapped to numbers very close to one.
\begin{equation}
\label{eq:sigmoid}
sigm(x) = \frac{1}{1+e^{-x}}
\end{equation}
\begin{figure}[th]
\centering
\includegraphics[height=3cm]{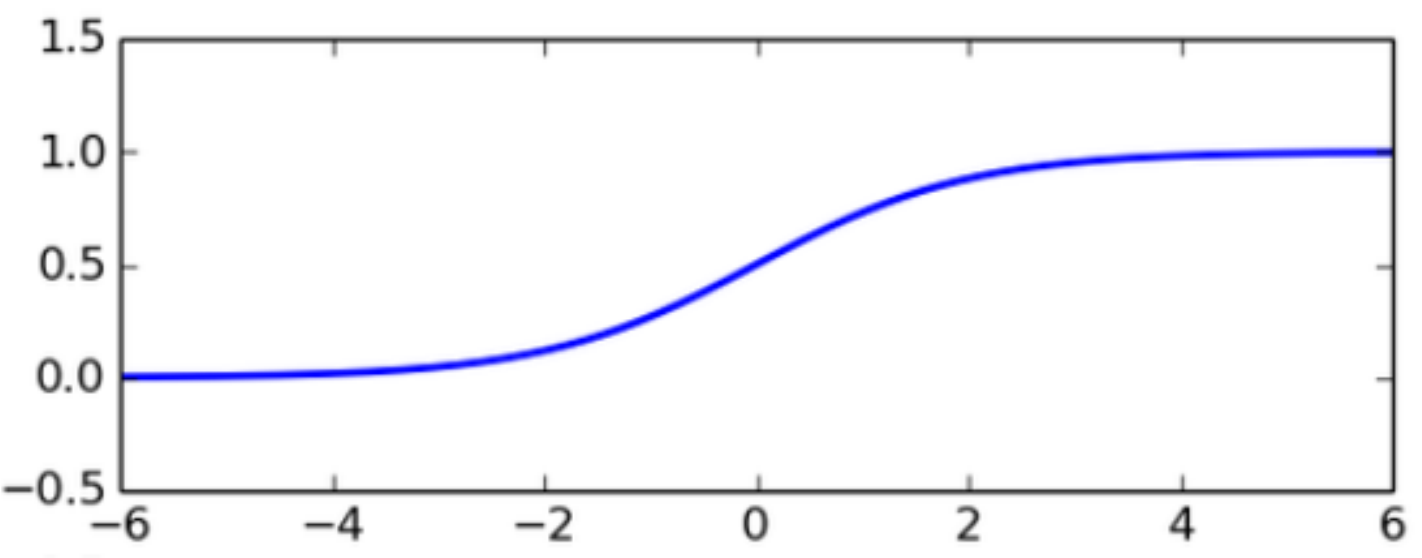}
\decoRule
\caption[Sigmoid function]{The Sigmoid function. Goes to zero when $x$ goes to $-\infty$ and to one when $x$ goes to $\infty$}
\label{fig:sigmoid}
\end{figure}
The Sigmoid function is rarely used for the hidden layers due to a significant disadvantage. As shown in figure \ref{fig:sigmoid}, when the function gets close to zero or one, its derivative goes to zero. During back-propagation the derivatives are multiplied by each other, thus resulting close to zero gradients (also known as the \textit{vanishing gradients} problem), and therefore the parameters barely change. A second disadvantage of the Sigmoid function is that it is not zero-centered.\\
Nowadays, the primary use of the Sigmoid function is for the output of binary classification.\\

The Tanh function is shown in equation \ref{eq:tanh}. The Tanh function looks very much like the Sigmoid function, but is zero-centered.
\begin{equation}
\label{eq:tanh}
tanh(x) = \frac{e^x-e^{-x}}{e^x+e^{-x}}=2*sigm(2x)
\end{equation}

\begin{figure}[th]
\centering
\includegraphics[height=3cm]{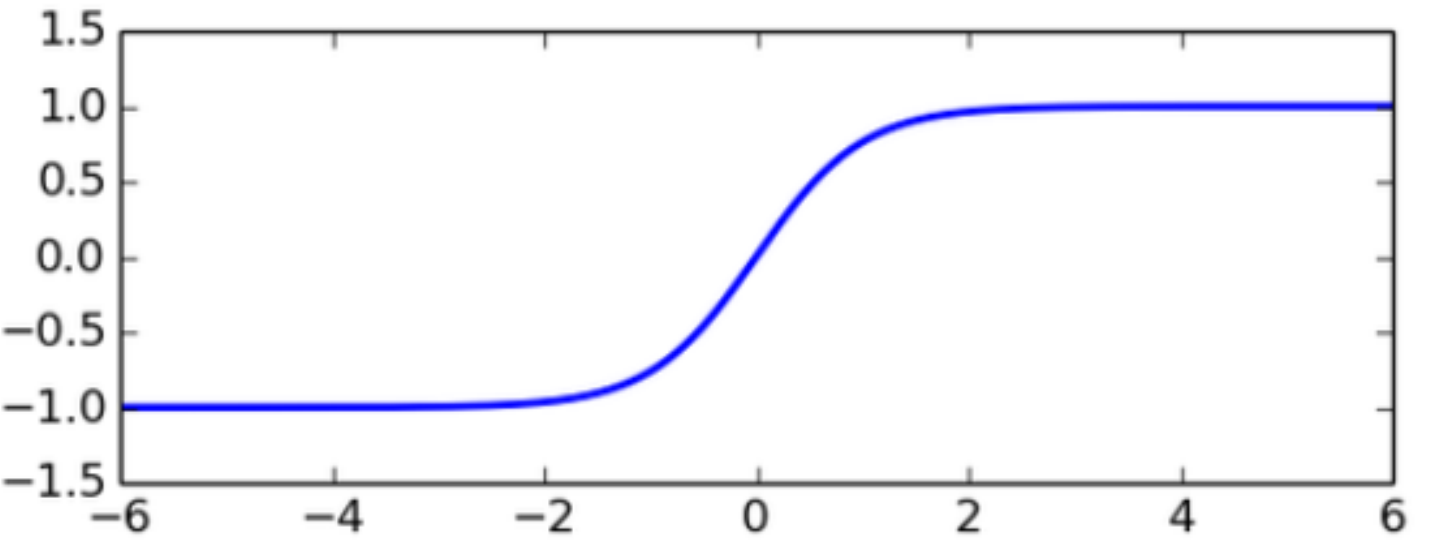} 
\decoRule
\caption[Tanh function]{The Tanh function. Goes to $-1$ when $x$ goes to $-\infty$ and to $1$ when $x$ goes to $\infty$}
\label{fig:tanh}
\end{figure}
As shown in figure \ref{fig:tanh}, the Tanh function does not solve the vanishing gradient problem.\\
Nowadays, the Tanh function is mostly used for the output layer of symmetric output problems (such as robotic control) and then multiplied by a scaling factor to match the output desired range.\\

The ReLU function is shown in equation \ref{eq:relu} and is currently the most popular activation function. The constant, equal to one, gradient allows the gradient to propagate back through the network without vanishing nor exploding. Furthermore, the simplicity of the ReLU function can reduce the training time.
\begin{equation}
\label{eq:relu}
relu(x) = max(0,x)
\end{equation}
\begin{figure}[th]
\centering
\includegraphics[height=3cm]{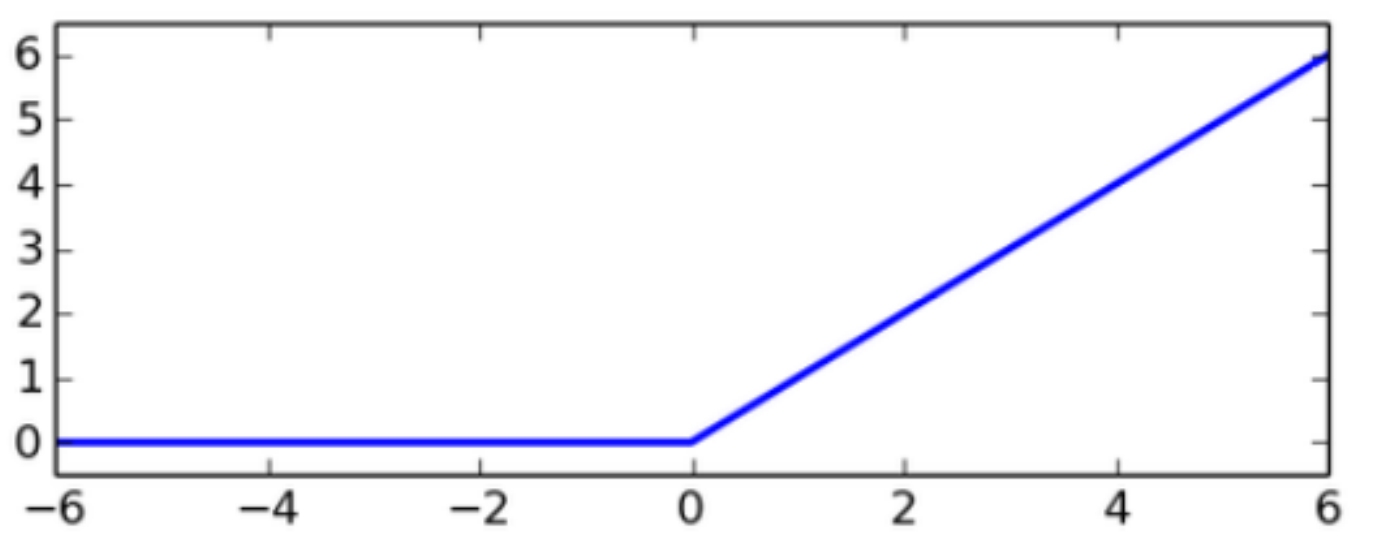}
\decoRule
\caption[ReLU function]{The ReLU function. Returns $x$ for positive $x$ and zero otherwise.}
\label{fig:relu}
\end{figure}
\subsection{Regularization} \label{ss:Regularization}
\begin{figure}[th]
\centering
\includegraphics[height=5cm]{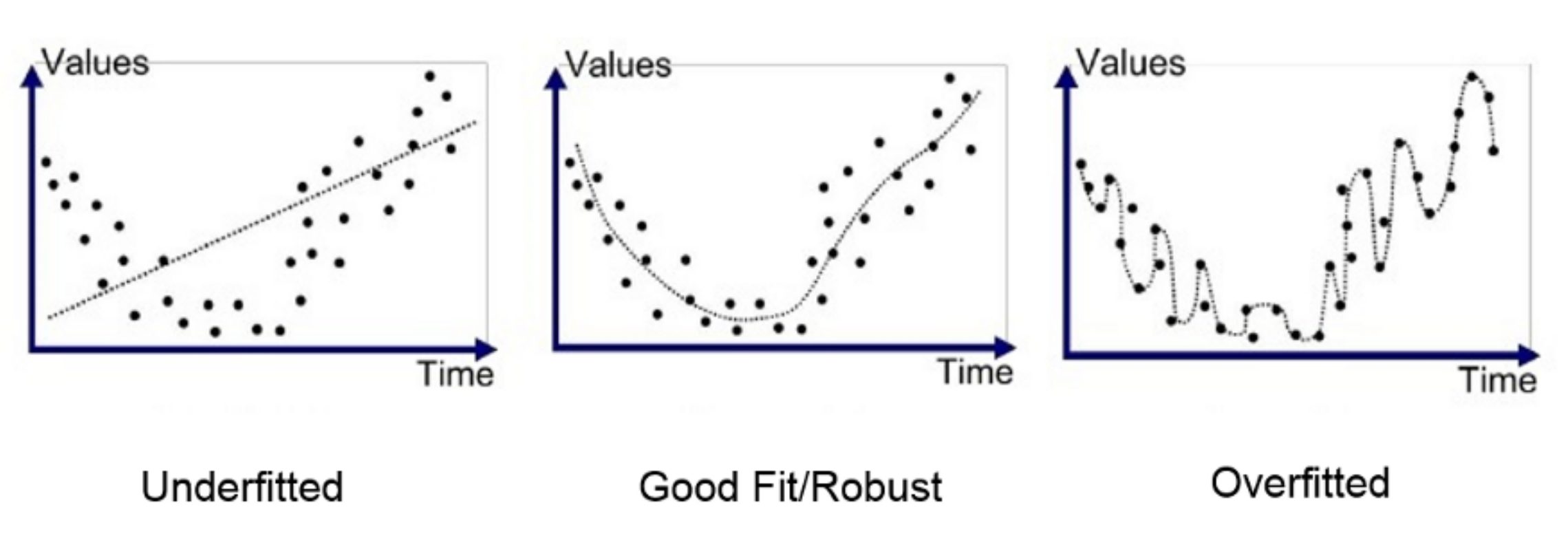}
\decoRule
\caption[overfitting and underfitting]{Underfiiting (left) happens when the model is too simple for the data. Overfitting (right) happens when the model is too complicated for the data\footnotemark.}
\label{fig:under-over-fitting}
\end{figure}

The ultimate goal of the learning process is to find a proper approximation function with valid generalization capabilities. That is, get similar results for new data as for the training data. There is a trade off between under- and over fitting (see figure \ref{fig:under-over-fitting}).\\
Underfitting refers to a model that can neither learn the training data nor generalize to new data. An underfitting machine learning will be easy to detect, as it will have poor performance on the training data. Underfitting is often not discussed as it is relatively easy to detect. The remedy is to move on and try alternative machine learning algorithms. Nevertheless, it does provide a good contrast to the problem of overfitting.\\
Overfitting refers to a scenario where the model learns the training data too well. Overfitting happens when a model learns the detail and noise in the training data to the extent that it negatively impacts the performance of the model on new data. This means that the noise in the training data memorized by the model as part of the data's underlying patterns. The problem is that the concepts learned from the noise do not apply to new data and reduces the model's ability to generalize. Overfitting occurs when the model is too complicated for the training data. One way to tackle overfitting is by penalizing too complex models, also known as regularization.\\
\footnotetext{Source: \quad\url{https://medium.com/greyatom/what-is-underfitting-and-overfitting-in-machine-learning-and-how-to-deal-with-it-6803a989c76}}
Regularization punishes model complexity, thus encourages the model to be as simple as possible. The loss function is extended with a regularization term $\Omega(\theta)$, that tries to keep the approximated function as simple as possible. Equation \ref{eq:regulized loss} shows the new loss function $L'$. $\lambda$ is the regularization factor. The higher $\lambda$, the more weight is given to the regularization term in the extended loss function. For $\lambda=0$, we get back the original loss $L$. 
\begin{equation}
\label{eq:regulized loss}
L'(\theta) = L(\theta) + \lambda\Omega(\theta)
\end{equation}
Equation \ref{eq:L2 regularization} defines a loss function with a L2-regularization. This regularization adds the sum of squared weights to the loss function in order to keep the weights as small as possible, and thus, limiting the model.
\begin{equation}
\label{eq:L2 regularization}
L'(\theta) = L(\theta) + \lambda\frac{1}{2}||\theta||_2
\end{equation}
Equation \ref{eq:L1 regularization} shows the L1-regularization. This regularization punishes the the model linearly by adding the sum of the weights' absolute values to the loss. Therefore, the model can give high values to a few weights, if other weights go to zero. This approach leads to sparse models.
\begin{equation}
\label{eq:L1 regularization}
L'(\theta) = L(\theta) + \lambda\frac{1}{2}||\theta||_1
\end{equation}

\newpage

\section{Deep Reinforcement Learning} \label{s:deep_reinforcement_learning}
The traditional RL algorithms (section \ref{s:traditional_reinforcement_learning}) are tabular. That is, all the values are stored in tables. This approach has several significant disadvantages. First, the tabular structure limits the algorithms to problems with a small number of states and actions. In most real-world problems, the state space is too large to be stored on a regular computer. Even if using computers with enormous storage capabilities, the time it will take to visit all states gets impossible. Secondly, when using a tabular structure, the algorithm cannot use the similarity between states and share knowledge.\\
To overcome these restrictions, a common approach is to replace the tables by function approximators that learn to map between features defining the state of the environment, to the approximated function's values. The most commonly used function approximation for reinforcement learning is a deep neural network (see section \ref{s:artefitial_neural_network}). Their ability to approximate non-linear functions and to extract relevant features from raw inputs makes it possible to generalize to unseen states.
\\ 
\subsection{Deep Q-Network} \label{ss:deep_q_network}
\begin{figure}[th]
\centering
\includegraphics[width=10cm]{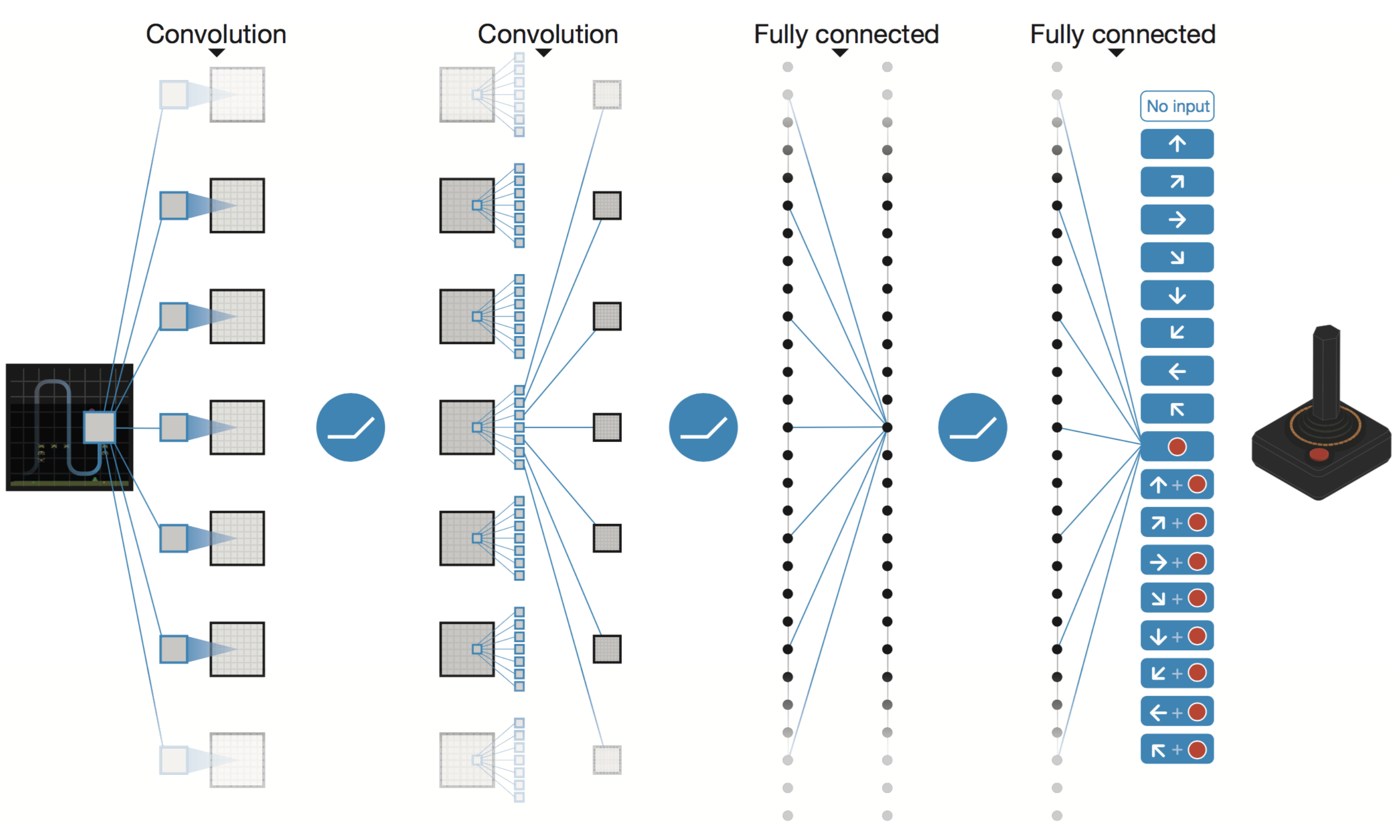}
\decoRule
\caption[DQN architecture]{The input to the network is the last eight frames of the game that is then fed through two convolutional layers, to process the visual information — following by two fully-connected layers for global reasoning. The output of the network is the estimated value of each possible action \footnotemark.}
\label{fig:DQN}
\end{figure}

Combining Q-learning with function approximators has been investigated in the past decades and did not lead to great success due to unstable learning. In 2015, a group of researchers at DeepMind presented a new algorithm - called Deep Q-Network (DQN) \cite{mnih2015human} that combined Q-learning with neural networks and showed a great success in playing Atari games. The inputs of the network are the raw pixels of the game so that the same algorithm can learn multiple games with no need for hand-crafted features. The outputs are the estimated value of each possible action. This end-to-end architecture enabled the network to extract relevant features by itself. The network architecture is shown in figure \ref{fig:DQN}. After estimating all actions, the algorithm uses an epsilon-greedy policy (see section \ref{s:traditional_reinforcement_learning}) to choose an action.
\footnotetext{Source: \cite{mnih2015human}}
To overcome the instability mentioned before, the DeepMind group presented two novel mechanisms: \emph{experience replay} and \emph{frozen target network}.\\

The idea of \emph{experience replay} is to store the agent's experience $S_t, A_t, R_t, S_{t+1}$ in a buffer. At every training step, a mini-batch of experience is uniformly sampled from the buffer and fed to the network for SGD. Neural networks need the data to be independent, or they [might] may overfit the last sequence they see. The frames of the games are highly correlated. The use of experience replay reduces the correlation between samples, and thus, the network can learn without overfitting. Another advantage of experience replay is the reuse of old experience, which make the learning smoother and more sample efficient.\\
The DQN algorithm uses the MSE loss function (see section \ref{ss:Learning Process}). The loss is computed using the squared TD error from the original Q learning algorithm.
\begin{equation}
\label{eq:DQN loss}
L_i(\theta_i) = \hat{\mathop{\mathbb{E}}}_t[(R_{t} + \gamma max_a(Q^-(S_{t+1},a, \theta_i^-))- Q(S_t,A_t, \theta_{i}))^2]
\end{equation} 
Equation \ref{eq:DQN loss} presents also the second mechanism, \textit{frozen target network}. The algorithm uses two networks with an identical architecture, but different weights values: $\theta$ for the Q-network, and $\theta^-$ for the target network.\\
The Q-network is updated using the loss term from equation \ref{eq:DQN loss}, while the target network stays frozen and updated once every C timesteps, by coping the parameters of the Q-network: $\theta^- = \theta$. This method induces a smoothing of oscillating policies and leads to more stabilized learning.

\begin{algorithm}[H]
\caption{Deep Q-Network}
\label{alg:DQN}
\hspace*{\algorithmicindent} \textbf{Input:}\quad the pixels and the game score, a preprocessing map $\phi$\\
\hspace*{\algorithmicindent} \textbf{Output:}\quad Q action value function
\begin{algorithmic}[1]
	\Statex Initialize replay buffer $\EuScript{D}$ to capacity N
	\Statex Initialize action-value function Q with random weights $\theta$
	\Statex Initialize target action-value function $Q(\cdot, \theta^-) = Q(\cdot, \theta) $
	\For {each episode}
	\Let{$s$}{$s_0$}		
	\While{$s$ in not terminal}
	\Statex \textbf{Play}
	\Let{$a$}{$\begin{cases}
		\text{random action} &\text{with prob $\epsilon$}\\
		argmax_a(Q(\phi(s),a|\theta)) &\text{otherwise}
		\end{cases}$}
	\State Execute action $a$ and observe reward $r$ and next state $s'$ 
	\State Store transition ($\phi(s), a, r, \phi(s')$) in $\EuScript{D}$
	\Let{$s$}{$s'$}
	\Statex \textbf{Train}
	\State Sample random minibatch $B$ of transitions ($\phi_j,a_j,r_j,\phi_{j+1}$) from $\EuScript{D}$
	\State Set $y_j=\begin{cases}
	r_j &\text{episode terminated at j+1}\\
	r_j+\gamma\,max_{a'}(Q^-(\phi_{j+1},a'|\theta^-)) &\text{otherwise}
	\end{cases}$
	\State Perform a gradient descent step on $(y_i-Q(\phi_i,a_j|\theta))^2$ on $\theta$
	\State Every $C$ steps set $\theta^- \leftarrow \theta$
	\EndWhile
	\EndFor
\end{algorithmic}	
\end{algorithm}

\subsection{Deep Deterministic Policy Gradient} \label{ss:deep_deterministic_policy_gradient}
When the action space is continuous, the architecture of DQN is impractical, since finding the best action for a given state turns into an optimization problem by itself. The algorithm Deep Deterministic Policy Gradient (DDPG) from Google DeepMind \cite{lillicrap2015continuous} solves this problem using a second network to predict the best action for each state — this architecture is also known as \textit{actor-critic}.
\subsubsection{Actor-Critic Architecture}
\begin{figure}[th]
\centering
\includegraphics[width=10cm]{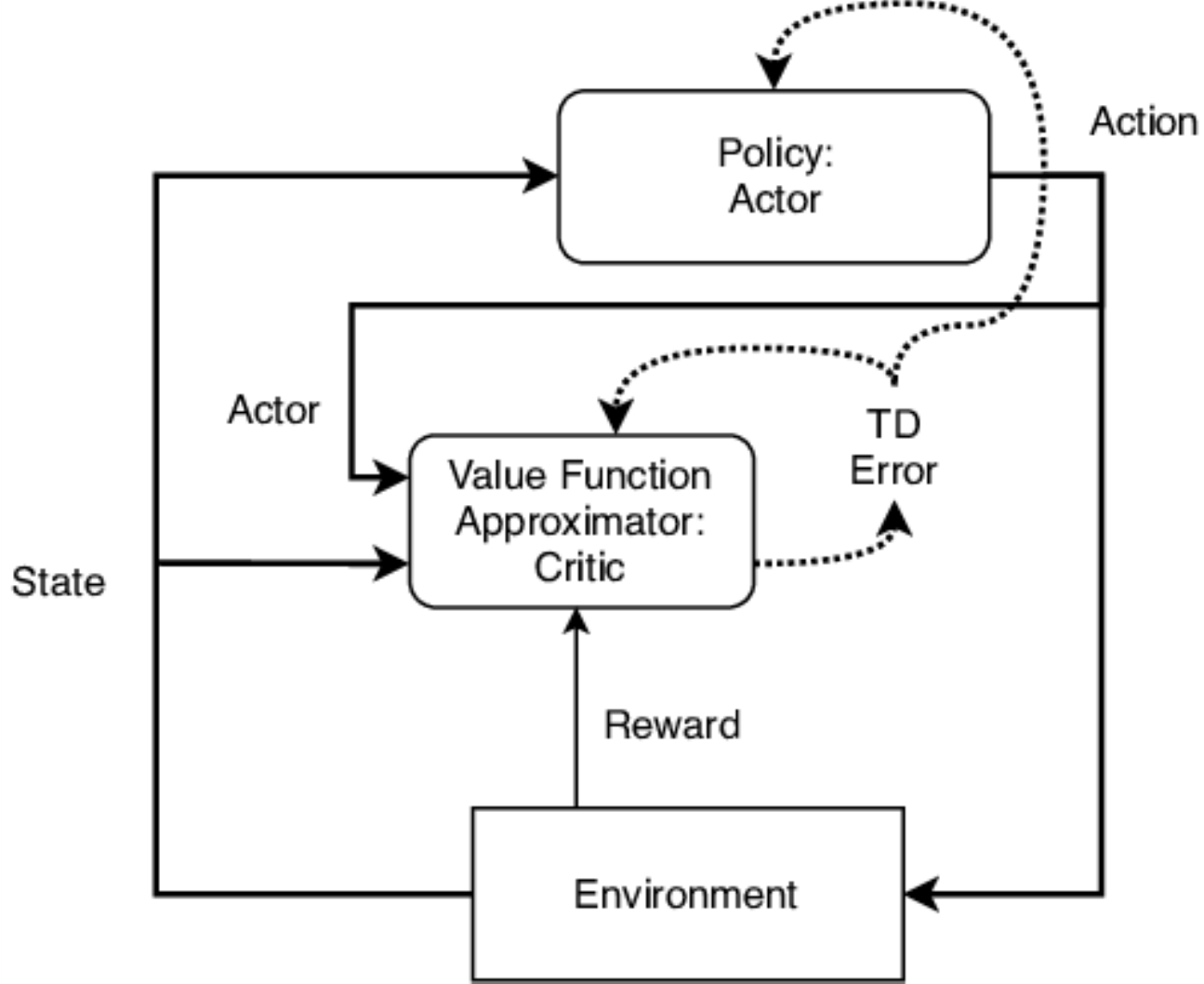}
\decoRule
\caption[Actor-Critic model]{The actor chooses an action given the state. The critic evaluates the state and action. The actor is trained using the critic's evaluation, and the critic is trained using the TD error\footnotemark.}
\label{fig:actor critic model}
\end{figure}

The actor-critic architecture includes two networks (see figure \ref{fig:actor critic model}):
\paragraph{Actor :} The purpose of the actor network, $\mu$, is to pick an action. The network gets the environment's state $S_t$ as input and returns the chosen action $\mu(S_t|\theta^\mu_i)$. The actor's weights $\theta^\mu$ are trained using gradient ascent over the Q value (try to pick an action that will maximize the Q value). Equations \ref{eq:actor loss} and \ref{eq:actor update} shows the actor's loss function and update rule respectively. The loss is negative, so that it will be maximized.
\begin{equation}
\label{eq:actor loss}
L_i(\theta^\mu_i) = \hat{\mathop{\mathbb{E}}}_t[Q(S_t, \mu(S_t|\theta^\mu_i)|\theta^\mu_i)]
\end{equation}
\begin{equation}
\label{eq:actor update}
\theta^\mu_{i+1} = \theta^\mu_{i} + \alpha_a\nabla_{\theta^\mu_i}L_i(\theta^\mu_i)
\end{equation}
Here the following problem arises: The Q value and its derivative are unknown. For that reason another network - the critic network is introduced.
\paragraph{Critic :} The purpose of the critic network is to evaluate the Q value of the state and action picked by the actor. The network gets the state $S_t$ and action $\mu(S_t|\theta^\mu_i)$ as input and returns their estimated Q value. The critic network is trained similarly to the Q network. Equations \ref{eq:critic loss} and \ref{eq:critic update} shows the critic's loss function and update rule respectively.
\begin{equation}
\label{eq:critic loss}
L_i(\theta_i^Q) = \hat{\mathop{\mathbb{E}}}_t[(R_{t} + \gamma Q^-(S_{t+1},\mu^-(S_{t+1}|\theta^{\mu-}_i) | \theta_i^{Q-}))- Q(S_t,A_t | \theta_i^{Q}))^2] 
\end{equation}
\begin{equation}
\label{eq:critic update}
\theta^Q_{i+1} = \theta^Q_{i} + \alpha_a\nabla_{\theta^Q_i}L_i(\theta^Q_i)
\end{equation}
The actor's loss uses the critic's estimation to calculate the derivatives. Using the chain rule, the gradient of the actor's loss can be written as follows:
\begin{equation}
\begin{aligned}
\nabla_{\theta^\mu_i}L_i(\theta^\mu_i) &=  \hat{\mathop{\mathbb{E}}}_t[\nabla_{\theta^\mu_i}Q(S_t,\mu(S_t|\theta^\mu_i)| \theta_i^{Q})]\\
&=\hat{\mathop{\mathbb{E}}}_t[\nabla_{\mu(S_t)}Q(S_t,\mu(S_t)| \theta_i^{Q})\nabla_{\theta^\mu_i}\mu(S_t|\theta^\mu_i)]
\end{aligned}
\end{equation}
\footnotetext{Source: \quad\url{https://www.researchgate.net/figure/Diagram-of-the-actor-critic-architecture-for-DDPG_fig1_333652544}}
The algorithm DDPG is also using \textit{experience replay} as presented in DQN. Additionally, DDPG uses a new version of \textit{frozen target network}, called \textit{soft target update}. Instead of coping the weights every C timesteps, the target network is updated continuously, as shown in equation \ref{eq:soft target update} and slowly approaches the original parameters.
\begin{equation}
\label{eq:soft target update}
\theta^- = \tau\theta + (1-\tau)\theta^-
\end{equation}
The parameter $\tau$ is set between zero and one and defines the tracking speed. That is, the smaller $\tau$, the slower the target weights approach the original weights.\\
In order to apply exploration to continuous action space, noise $\EuScript{N}$ is added to the actor’s output $\mu(s|\theta^\mu)$. The noise is generated by an Ornstein-Uhlenbeck process \cite{uhlenbeck1930theory}.

\begin{algorithm}[H]
\caption{Deep Deterministic Policy Gradient}
\label{alg:DDPG}
\hspace*{\algorithmicindent} \textbf{Input:}\quad the state of the environment
\hspace*{\algorithmicindent} \textbf{Output:}\quad action to perform
\begin{algorithmic}[1]
	\Statex Initialize replay buffer $\EuScript{D}$ to capacity $N$
	\Statex Randomly initialize critic $Q(s,a|\theta^Q)$ and actor $\mu(s|\theta^\mu)$
	\Statex Initialize target network $Q^-$ and $\mu^-$ with weights $\theta^{Q^-}\leftarrow\theta^{Q}$, $\theta^{\mu^-}\leftarrow\theta^{\mu}$
	\For {each episode}
	\State Initialize a random process $\EuScript{N}$ for action exploration
	\Let{$s$}{$s_0$}		
	\While{$s$ in not terminal}
	\Statex \textbf{Play}
	\Let{$a$}{$\mu(s|\theta^\mu)+\EuScript{N}$} according to current policy
	\State Execute action $a$ and observe reward $r$ and next state $s'$ 
	\State Store transition ($s, a, r,s'$) in $\EuScript{D}$
	\Let{$s$}{$s'$}
	\Statex \textbf{Train}
	\State Sample random minibatch $B$ of $N$ samples ($s_j,a_j,r_j,s_{j+1}$) from $\EuScript{D}$
	\State Set $y_j=\begin{cases}
	r_j&\text{$s_{j+1}$ is a termination state}\\
	r_j+\gamma Q^-(s_{j+1},\mu^-(s_{j+1}|\theta^{\mu^-})|\theta^{Q^-})&\text{otherwise}
	\end{cases}$
	\State Update critic by minimizing the loss $L=\frac{1}{N}\sum_i(y_i-Q(s_i,a_i|\theta^Q))^2$ 
	\State Update the actor policy using the sampled policy gradient:
	\Statex \quad\quad\quad $\nabla_{\theta^\mu}L=-\frac{1}{N}\sum_i \nabla_{\mu(s_i)} Q(s_i,\mu(s_i)|\theta_Q)\nabla_{\theta^\mu}\mu(s_i|\theta^\mu)$
	\State update the target networks:
	\Statex \quad\quad\quad $\theta^{Q^-}\leftarrow \tau\theta^Q+(1-\tau)\theta^{Q^-}$
	\Statex \quad\quad\quad $\theta^{\mu^-}\leftarrow \tau\theta^\mu+(1-\tau)\theta^{\mu^-}$
	\EndWhile
	\EndFor
\end{algorithmic}	
\end{algorithm}

\subsection{Prioritized Experience Replay} \label{ss:Prioritized Experience Replay}
\textit{Prioritized Experience Replay} (PER) \cite{schaul2015prioritized} is an improvement to the \textit{experience replay} mechanism in DQN and DDPG. The idea of PER is to prioritize experience from which the agent can learn more valuable information. Each experience is stored with an additional value $\omega_i$ defining the priority of that sample, so that experiences with higher values, have a higher probability $P_i$ of getting samples during the experience replay (equation \ref{eq:PER probabilities}).\\
\begin{equation}
\label{eq:PER probabilities}
P(i) = \frac{\omega_i^\alpha}{\sum_k\omega_k^\alpha}
\end{equation}
The exponent $\alpha$ determines how much prioritization is used, with $\alpha = 0$ corresponding to the uniform case.\\
As an importance measure, the TD-error can be used. It is assumed that the agent can learn more from experience with high TD error since a smaller TD error indicates more familiarity with the corresponding state-action pair.

\subsection{Reinforcement Learning for Sparse Reward Function} \label{ss:sparse reward function}
Reward functions can be divided into two categories: \textit{Dense rewards} and \textit{Sparse rewards} (also known as binary rewards). With dense reward functions, better policies lead to a higher return. Thus, the agent knows whenever it improves. With sparse reward functions on the other hand, most policies lead to the same return, and the agent will only know it is getting better once it will pass some threshold of performances. The reward function will usually be $0$ if the agent achieved the target, and $-1$ otherwise. An example of a binary reward can be a grid-world game, where the agent gets $-1$ for every step until it reaches the target state.
Learning from sparse reward is a known challenge in reinforcement learning since it an extensive amount of exploration to reach the goal and receive some learning signal.
In the following sections, we will introduce two algorithms that can learn from sparse reward functions.
\subsubsection{Demonstration-Initialized Rollout Worker} \label{ss:Demonstration-Initialized Rollout Worker}
Learning from sparse reward requires a lot of exploration. One way to overcome this requirement is to train the agent using expert demonstrations and supervised learning methods. This approach has two main downsides: First, supervised learning requires a vast amount of demonstrations for training the agent. Second, by learning from the expert's action, the agent can never gets better than the expert. To resolve those problems, the algorithm \emph{Demonstration-Initialized Rollout Worker} was introduced \cite{salimans2018learning}. This algorithm uses a form of dynamic programming to reduce learning time. The algorithm gets a single demonstration of an expert and uses it to facilitate the training of the agent. At each episode, the algorithm initializes the game in a state from the given trajectory. 
At the beginning of the training, the episode is initialized close to the termination state at step $t$ out of $T$ steps, and the agent learns how to achieve the target from that state using reinforcement learning. If the agent performs at least as good as the expert, we gradually update the initial state closer to the real initial state. See algorithms \ref{alg:Demonstration-Initialized Rollout Worker} and \ref{alg:Optimizer} for the formal description. By applying this reverse curriculum, the agent can learn how to reach the target state much faster, and can even get better than the expert. Demonstration-Initialized Rollout Worker has two main disadvantages. First, it requires expert demonstrations to learn from. Second, it can only work for environments with a single target.

\begin{algorithm}
\caption{Demonstration-Initialized Rollout Worker}
\label{alg:Demonstration-Initialized Rollout Worker}
\begin{algorithmic}[1]
	\Require
	\Statex \textbullet~ an expert's demonstration $\tau = \{(\tilde{s}_i,\tilde{a}_i,\tilde{r}_i,\tilde{s}_{i+1},\tilde{d}_i)\}_{i=0}^T$
	\Statex \textbullet~ $D$ - number of possible starting points
	\Statex \textbullet~ $M$ - number of rollouts in batch
	\Statex
	\State Initialize max starting point $i_{max} \leftarrow T$

	\While{True}
	\State Get latest policy $\pi(\theta)$ from optimizer
	\State Get latest reset point $i_{max}$ from optimizer
	\State Initialize success counter $\mathcal{W}=0$
	\State Initialize batch $\mathcal{D}=\{\}$
	
	\For{$rollout \leftarrow 1, M$}
	\State Sample starting point $i$ by sampling uniformly from $\{i_{max} - D, ..., i_{max}\}$
	\State $t \leftarrow 0$
	\State Initialize state $s_t$ to state $s_i$ in the trajectory $\tau$
	\State $done \leftarrow False$
	\While{not done}
	\State Sample action $a_t\sim \pi(s_t)$
	\State Take action $a_t$ in the environment
	\State Receive reward $r_t$, next state $s_{t+1}$ and done signal $d_t$
	\State $done \leftarrow d_t$
	\State Add date $\{s_t, a_t, r_t, s_{t+1}, d_t\}$
	\State $t \leftarrow t+1$
	\If{if $done$}
	\If{$\Sigma_{i=0}^t\gamma^i\cdot r_i \geq \Sigma_{i=0}^t\gamma^i\cdot \tilde{r}_i$} \Comment As good as demo
	\State $\mathcal{W} \leftarrow \mathcal{W}+1$
	\EndIf
	\EndIf
	\State $t \leftarrow t+1$
	\EndWhile
	\EndFor
	\State Send batch $\mathcal{D}$ and counter $\mathcal{W}$ to optimizer \Comment see algorithm \ref{alg:Optimizer}
	\EndWhile
\end{algorithmic}
\end{algorithm}

\begin{algorithm}
	\caption{Optimizer}
	\label{alg:Optimizer}
	\begin{algorithmic}[1]
		\Require
		\Statex \textbullet~ $i_{max}$ - max starting point
		\Statex \textbullet~ $\mathcal{D}$ - batch of agent's experience
		\Statex \textbullet~ $\mathcal{W}$ - number of seccessful rollouts
		\Statex \textbullet~ $\Delta$ - starting point shift size,\, $\rho$ - success threshold,\, $\theta$ - agent's parameters,\, $\mathcal{A}$ - learning algorithm,\, $D$ - number of possible starting points
		\Statex
		\If{$\frac{\mathcal{W}}{D} > \rho$} \Comment The agent is successful sufficiently often
		\State $i_{max}\leftarrow i_{max}-\Delta$
		\EndIf
		\State $\theta \leftarrow \mathcal{A}(\theta, \mathcal{D})$
	\end{algorithmic}
\end{algorithm}

\subsubsection{Hindsight Experience Replay} \label{ss:hindsight_experience_replay}
\begin{figure}%
\centering
\subfloat[]{{\includegraphics[width=6cm]{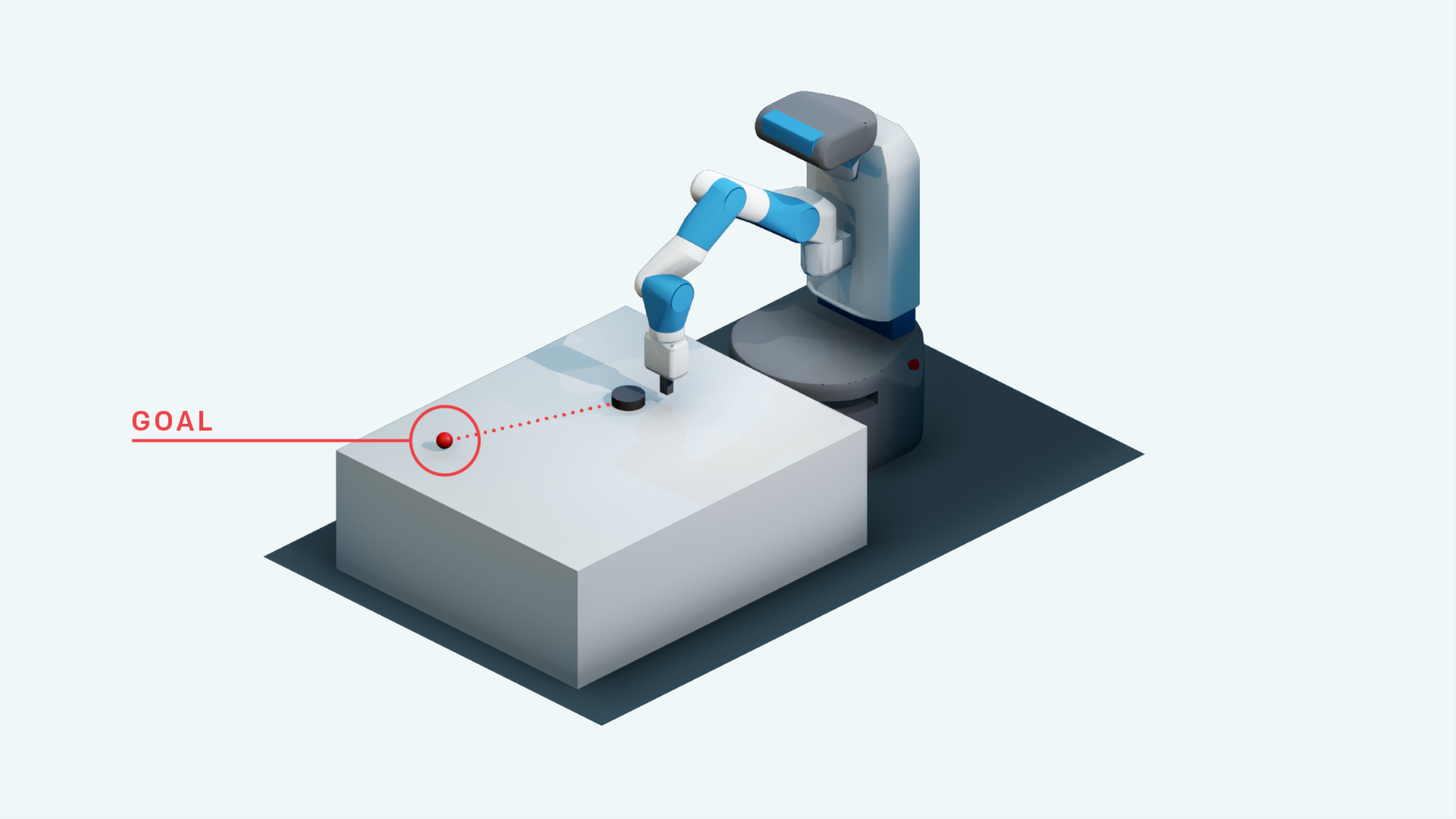}}}%
\qquad
\subfloat[]{{\includegraphics[width=6cm]{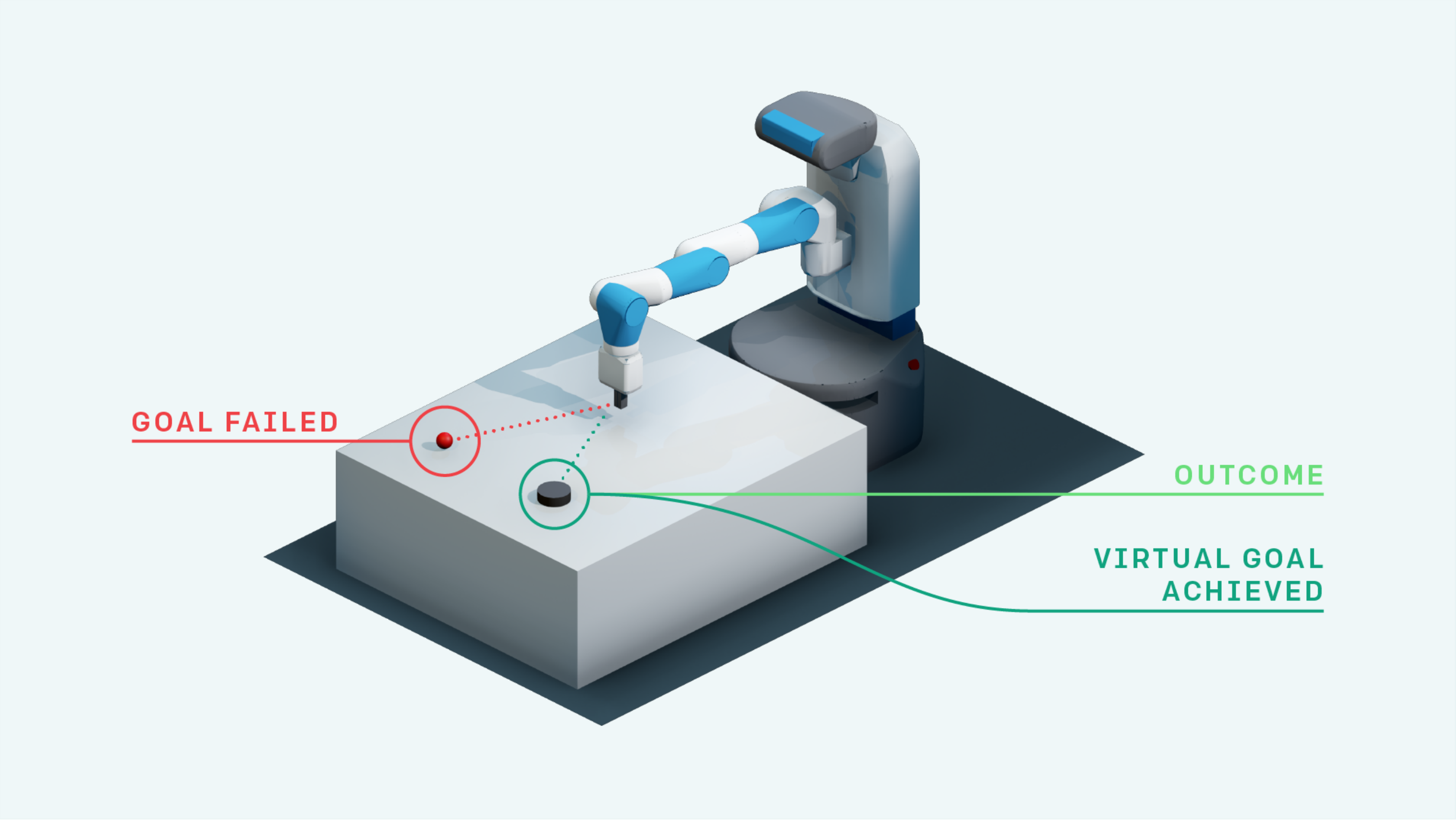}}}%
\caption[HER illustration]{OpenAI's Push Environment used in HER. (A) shows the original goal target for the puck. (B) shows the generation of a virtual goal, after failing the original task\footnotemark}
\label{fig:HER}%
\end{figure}

Many of previous RL achievements are concerned with a particular objective, such as "check mate the opponent in chess". In these problems, the agent picks an action for the given state and gets a reward. However, many real-world problems are not like that. There are cases where we like our agent to achieve many different goals. The agent should get the current goal in the game and pick an action accordingly. These are known as \textit{multi-goal tasks}.\\
DDPG can be extended to multi-goal tasks using \textit{Universal Value Function Approximators} (UVFA) \cite{schaul2015universal}. The key idea behind UVFA is to augment action-value functions and policies by goal states, and thus, every transition contains also the desired goal. This enables generalization not only over states but also over goals when using neural networks as function approximators.\\ 
In multi-goal tasks with \textit{sparse} rewards (that is, 0 for achieving the goal and -1 to all other states), it is challenging to learn the task and achieve any progress. \textit{Hindsight Experience Replay} (HER) \cite{andrychowicz2017hindsight} is a an algorithm from OpenAI to solve this problem. When a traditional algorithm sees a failure, it can only learn that the given trajectory was not successful. Thus, in order to know what is indeed useful, the agent must first reach the goal accidentally.\\
HER addresses this problem by taking a failure as a success to an alternative (or \textit{virtual}) goal (see figure \ref{fig:HER}).
HER applies UVFA and includes additional transitions with virtual goals. Thus, the agent can learn from failures through generalization to actual goals. It has been demonstrated that HER significantly improves the performances in various challenging simulated robotic environments. Every transition $<S_T, A_T, R_t, S_{t+1}, G>$ in the trajectory is also inserted into the buffer with an alternative goal, achieved in the future. For choosing the virtual goal, the algorithm uses a specific strategy( e.g., \textit{Future}), which randomly picks an achieved state for each transition.

\footnotetext{Source: \quad\url{https://openai.com/blog/ingredients-for-robotics-research/}}

\begin{algorithm}
\caption{Hindsight Experience Replay}
\label{alg:HER}
\begin{algorithmic}[1]
	\Require
	\Statex \textbullet~ an off-policy RL algorithm $\mathbb{A}$,\Comment e.g. DQN, DDPG 
	\Statex \textbullet~ a reward function: $\mathcal{S}\times \mathcal{A}\times \mathcal{G} \rightarrow \mathcal{R}$,\Comment e.g. $r(s,a,g)=-1 \text{ if fail, } 0 \text{ if success}$
	\Statex
	
	\State Initialize $\mathbb{A}$
	\State Initialize replay buffer $R$
	
	\While{True}
	\For{$Episode \leftarrow 1, M$}
	\State Sample a goal $g$ and an initial state $s_0$.
	\For{$t \leftarrow 0, T-1$}
	\State Sample an action $a_t$ using the behavioral policy from $\mathbb{A}$: \par
	$\qquad \qquad a_t \leftarrow \pi(s_t||g)$ \Comment || denotes concatenation
	\State Execute the action $a_t$ and observe a new state $s_{t+1}$
	\EndFor
	
	\For{$t \leftarrow 0, T-1$}
	\State $r_t := r(s_{t+1}, g)$
	\State Store the transition $(s_t||g, a_t, r_t, s_{t+1}||g)$ in $R$
	
	\State Sample a set of virtual goals $\tilde{G}$ for replay $\tilde{G} := \mathbb{S}(\text{\textbf{current episode}})$
	\For {$\widetilde{g} \in \tilde{G}$}
	\State $\widetilde{r} = r(s_{t+1}, \widetilde{g})$
	\State Store the transition $(s_t||\widetilde{g}, a_t, \widetilde{r}, s_{t+1}||\widetilde{g})$ in $R$\Comment HER
	\EndFor
	\EndFor
	\EndFor
	
	\For{$t \leftarrow 1, N$}
	\State Sample a minibatch $B$ from the replay buffer $R$
	\State Perform one step of optimization using $\mathbb{A}$ and minibatch $B$
	\EndFor
	\EndWhile
	
\end{algorithmic}
\end{algorithm} 

\chapter{Environments} 

\label{Environments} 


In this chapter, we introduce the environments that we have built for evaluating our algorithms. All our environments has continuous state and action spaces.
For the full description of our environments' layout, see appendix \ref{AppendixA}.


\section{Environments Classes}
For the sake of simplicity, we restricted our research to 2D simulations.
All throwing tasks follows the same structure:
\begin{itemize}
	\item v0 : Simplified version. The hand is holding the ball from the beginning, and thus, the agent only learns to throw.
	\item v1 : Full version. The ball is initialized on the floor, and thus, the agent needs to learn both how to pick the ball and how to throw it towards the target.
\end{itemize}

\subsection{hand\_reach} \label{ss:hand-reach}
In this game, the agent needs to reach the ball with the hand (see figure \ref{fig:hand-reach}).

\subsubsection{Observation}
\begin{table}[H]
	\caption[Hand\_reach observation]{Hand\_reach observation}
	\centering
	\begin{tabular}{||c | c | c ||} 
		\hline
		Num & Observation & Type\\ [0.5ex] 
		\hline\hline
		0 & hand $x$ position & continuous\\ 
		\hline
		1 & hand $y$ position & continuous\\
		\hline
		2 & hand $x$ velocity & continuous\\
		\hline
		3 & hand $y$ velocity & continuous\\
		\hline
		4 & hand state (open/close) & binary\\
		\hline
	\end{tabular}
	\label{tbl:hand-reach observation}
\end{table}

\subsubsection{Actions}
\begin{table}[H]
	\caption[Hand\_reach action]{Hand\_reach action}
	\centering
	\begin{tabular}{||c | c | c||} 
		\hline
		Num & Action & Type\\ [0.5ex] 
		\hline\hline
		0 & hand $x$ velocity & continuous\\
		\hline
		1 & hand $y$ velocity & continuous\\
		\hline
        2 & hand state (open/close) & binary\\
		\hline
    \end{tabular}
	\label{tbl:hand-reach action}
\end{table}

\subsubsection{Goal}
\begin{table}[H]
	\caption[Hand\_reach goal]{Hand\_reach goal}
	\centering
	\begin{tabular}{||c | c | c||} 
		\hline
		Num & Goal & Type\\ [0.5ex] 
		\hline\hline
		0 & ball $x$ position & continuous\\
		\hline
		1 & ball $x$ position & continuous\\
		\hline
	\end{tabular}
	\label{tbl:hand-reach goal}
\end{table}

\subsubsection{Reward function}
The reward is binary, i.e., $0$ if the target is achieved and $-1$ otherwise:\\
\begin{center}
	$R(s_{t}) = \begin{cases}
	0,       & ||ball_{pos}-hand_{pos}|| < \epsilon \\
	-1,      & otherwise \\
	\end{cases}$
\end{center}

\subsection{hand throwing tasks} \label{ss:hand throwing tasks}
These tasks include a hand, a ball and a target. The goal in these tasks is to get the ball close enough to the target
\subsubsection{hand\_v0}
In this game, the hand holds the ball from the beginning and needs to throw the ball towards the target (see figure \ref{fig:hand-v0}).
\subsubsection{hand\_v1}
In this game, the ball is initialized on the ground and the agent needs also to learn how to pick the ball (see figure \ref{fig:hand-v1}).
\subsubsection{hand\_wall\_v0}
This game is like hand\_0, but there is also a wall and the agent needs to throw the ball above the wall (see figure \ref{fig:hand-wall-v0}).
\subsubsection{hand\_wall\_v1}
This game is like hand\_1, but there is also a wall and the agent needs to throw the ball above the wall (see figure \ref{fig:hand-wall-v1}).

\subsubsection{Observation}
\begin{table}[H]
	\caption[Hand\_throw observation]{Hand\_throw observation}
	\centering
	\begin{tabular}{||c | c | c||} 
		\hline
		Num & Observation & Type\\ [0.5ex] 
		\hline\hline
		0 & hand $x$ position & continuous\\ 
		\hline
		1 & hand $y$ position & continuous\\
		\hline
		2 & hand $x$ velocity & continuous\\
		\hline
		3 & hand $y$ velocity & continuous\\
		\hline
		4 & hand state (open/close) & binary\\
		\hline
		5 & ball $x$ position & continuous\\
		\hline
		6 & ball $y$ position & continuous\\
		\hline
		7 & ball $x$ velocity & continuous\\
		\hline
		8 & ball $y$ velocity & continuous\\
		\hline
	\end{tabular}
	\label{tbl:hand-throw observation}
\end{table}

\subsubsection{Actions}
\begin{table}[H]
	\caption[Hand\_throw action]{Hand\_throw action}
	\centering
	\begin{tabular}{||c | c | c||} 
		\hline
		Num & Action & Type\\ [0.5ex] 
		\hline\hline
		0 & hand $x$ velocity & continuous\\
		\hline
		1 & hand $y$ velocity & continuous\\
		\hline
		2 & hand state (open/close) & binary\\
		\hline
	\end{tabular}
	\label{tbl:hand-throw action}
\end{table}

\subsubsection{Goal}
\begin{table}[H]
	\caption[Hand\_throw goal]{Hand\_throw goal}
	\centering
	\begin{tabular}{||c | c | c||} 
		\hline
		Num & Goal & Type\\ [0.5ex] 
		\hline\hline
		0 & black-hole $x$ position & continuous\\
		\hline
		1 & black-hole $x$ position & continuous\\
		\hline
	\end{tabular}
	\label{tbl:hand-throw goal}
\end{table}

\subsubsection{Reward function}
The reward is binary: $0$ if the target is achieved and $-1$ otherwise:\\
\begin{center}
	$R(s_{t}) = \begin{cases}
	0,       & ||goal_{pos}-ball_{pos}|| < \epsilon \\
	-1,      & otherwise \\
	\end{cases}$
\end{center}

\subsection{Robot reach} \label{ss:robot-reach}
In this game, the agent needs to reach the ball with the end-effector of the manipulator (see figure \ref{fig:robot-reach}).

\subsubsection{Observation}
\begin{table}[H]
	\caption[Robot\_reach observation]{Robot\_reach observation}
	\centering
	\begin{tabular}{||c | c | c||} 
		\hline
		Num & Observation & Type\\ [0.5ex] 
		\hline
		0-1 & $\theta$ (joint's angles) & continuous\\ 
		\hline
		2 & end-effector $x$ position & continuous\\
		\hline
		3 & end-effector $y$ position & continuous\\
		\hline
		4-5 & $\dot{\theta}$ (joint's velocity) & continuous\\
		\hline
		6 & end-effector $x$ velocity & continuous\\
		\hline
		7 & end-effector $y$ velocity & continuous\\
		\hline
		8 & end-effector state (open/close) & binary\\
		\hline
	\end{tabular}
	\label{tbl:robot-reach observation}
\end{table}

\subsubsection{Actions}
\begin{table}[H]
	\caption[Robot\_reach action]{Robot\_reach action}
	\centering
	\begin{tabular}{||c | c | c||} 
		\hline
		Num & Action & Type\\ [0.5ex] 
		\hline\hline
		0-1 & $\dot{\theta}$ (joint's velocity) & continuous\\
		\hline
		2 & end-effector state (open/close) & binary\\
		\hline
	\end{tabular}
	\label{tbl:robot-reach action}
\end{table}

\subsubsection{Goal}
\begin{table}[H]
	\caption[Robot\_reach goal]{Robot\_reach goal}
	\centering
	\begin{tabular}{||c | c | c||} 
		\hline
		Num & Goal & Type\\ [0.5ex] 
		\hline\hline
		0 & ball $x$ position & continuous\\
		\hline
		1 & ball $x$ position & continuous\\
		\hline
	\end{tabular}
	\label{tbl:robot-reach goal}
\end{table}

\subsubsection{Reward function}
The reward is binary, i.e., $0$ if the target is achieved and $-1$ otherwise:\\
\begin{center}
	$R(s_{t}) = \begin{cases}
	0,       & ||ball_{pos}-end\_effector_{pos}|| < \epsilon \\
	-1,      & otherwise \\
	\end{cases}$
\end{center}

\subsection{Robot throwing tasks} \label{ss:robot throwing tasks}
These tasks include a hand, a ball and a target. The goal in these tasks is to get the ball close enough to the target
\subsubsection{robot\_v0}
In this game, the end-effector holds the ball from the beginning and needs to throw the ball towards the target (see figure \ref{fig:robot-v0}).
\subsubsection{robot\_v1}
In this game, the ball is initialized within the manipulator's reachable area, and the agent needs also to learn how to pick the ball (see figure \ref{fig:robot-v1}).

\subsubsection{Observation}
\begin{table}[H]
	\caption[Robot\_throw observation]{Robot\_throw observation}
	\centering
	\begin{tabular}{||c | c | c||} 
		\hline
		Num & Observation & Type\\ [0.5ex] 
		\hline
		0-1* & $\theta$ (joint's angles) & continuous\\ 
		\hline
		2 & end-effector $x$ position & continuous\\
		\hline
		3 & end-effector $y$ position & continuous\\
		\hline
		4-5 & $\dot{\theta}$ (joint's velocity) & continuous\\
		\hline
		6 & end-effector $x$ velocity & continuous\\
		\hline
		7 & end-effector $y$ velocity & continuous\\
		\hline
		8 & end-effector state (open/close) & binary\\
		\hline
		9 & ball $x$ position & continuous\\
		\hline
		10 & ball $y$ position & continuous\\
		\hline
		11 & ball $x$ velocity & continuous\\
		\hline
		12 & ball $y$ velocity & continuous\\
		\hline
	\end{tabular}
	\label{tbl:robot-throw observation}
\end{table}
*After scaling, $\theta$ turns to $(cos(\theta), sin(\theta))$.

\subsubsection{Actions}
\begin{table}[H]
	\caption[Robot\_throw action]{Robot\_throw action}
	\centering
	\begin{tabular}{||c | c | c||} 
		\hline
		Num & Action & Type\\ [0.5ex] 
		\hline\hline
		0-1 & $\dot{\theta}$ (joint's velocity) & continuous\\
		\hline
		2 & end-effector state (open/close) & binary\\
		\hline
	\end{tabular}
	\label{tbl:robot-throw action}
\end{table}

\subsubsection{Goal}
\begin{table}[H]
	\caption[Robot\_throw goal]{Robot\_throw goal}
	\centering
	\begin{tabular}{||c | c | c||} 
		\hline
		Num & Goal & Type\\ [0.5ex] 
		\hline\hline
		0 & black-hole $x$ position & continuous\\
		\hline
		1 & black-hole $x$ position & continuous\\
		\hline
	\end{tabular}
	\label{tbl:robot-throw goal}
\end{table}

\subsubsection{Reward function}
The reward is binary, i.e., $0$ if the target is achieved and $-1$ otherwise:\\
\begin{center}
	$R(s_{t}) = \begin{cases}
	0,       & ||goal_{pos}-ball_{pos}|| < \epsilon \\
	-1,      & otherwise \\
	\end{cases}$
\end{center}

\begin{table}[H]\sffamily
	\begin{center}
	\begin{tabular}{c c}
		\toprule
		\subfloat[]{\tcbox{\includegraphics[width=12em]{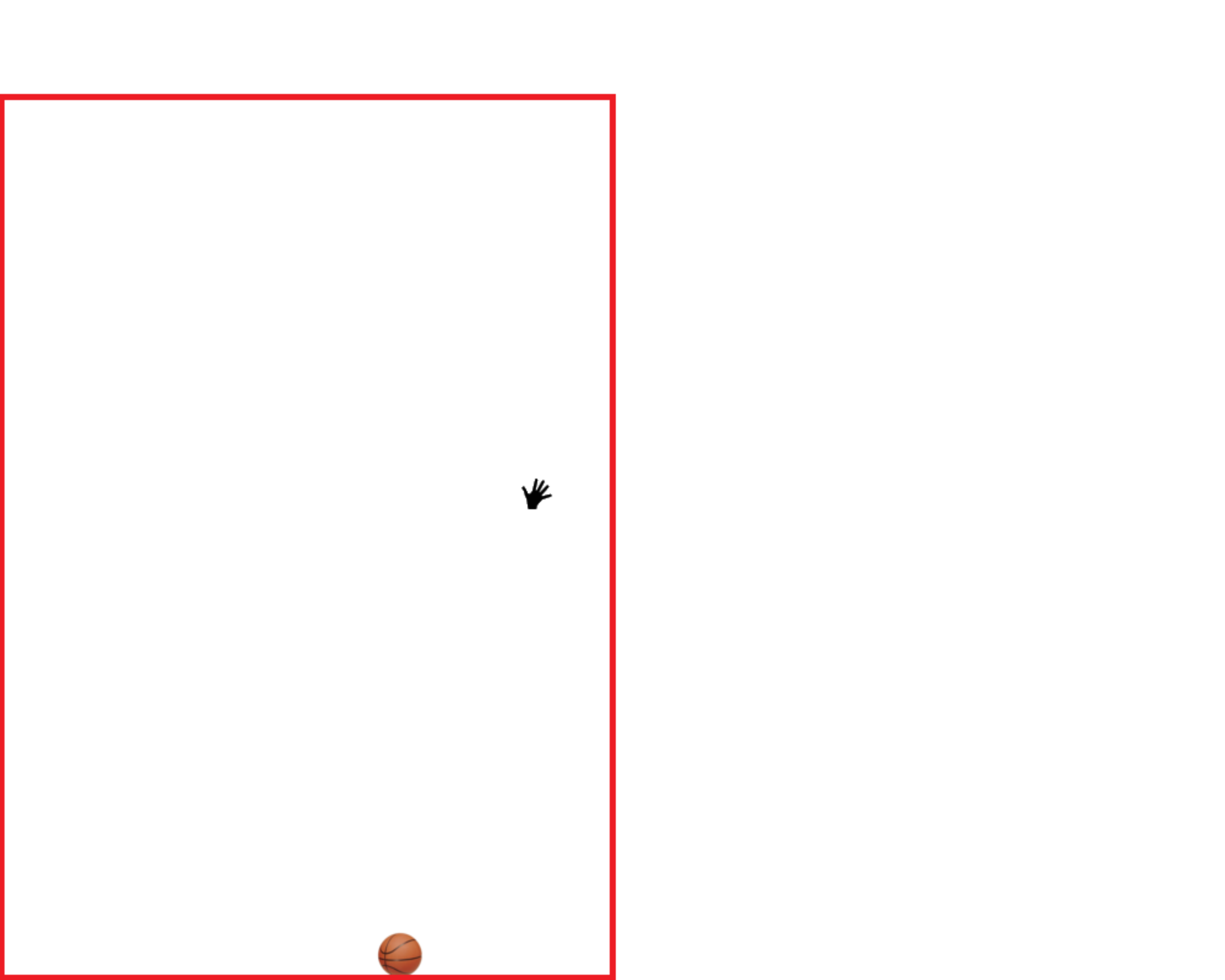}} \label{fig:hand-reach}}
		& \subfloat[]{\tcbox{\includegraphics[width=12em]{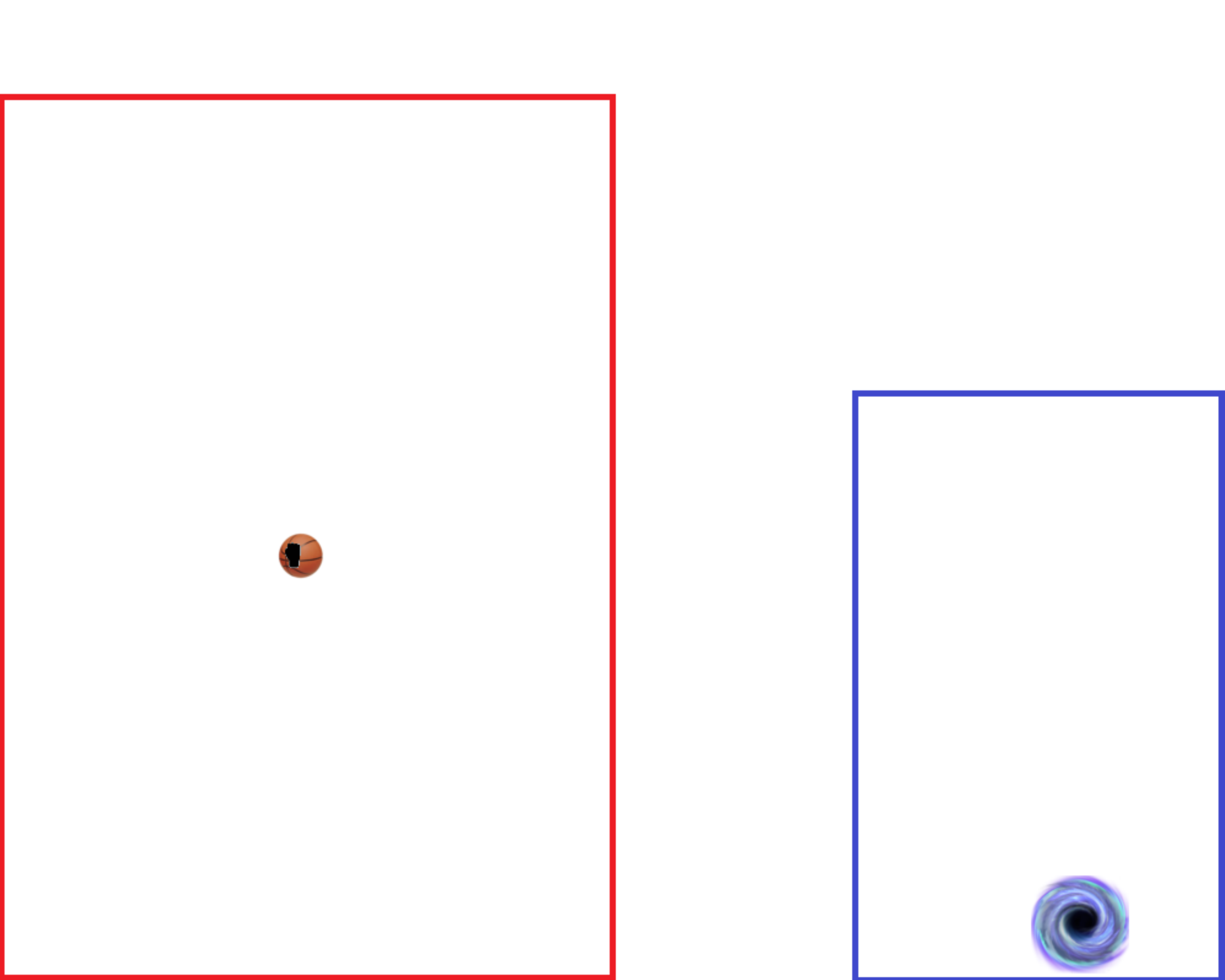}} \label{fig:hand-v0}} \\ \subfloat[]{\tcbox{\includegraphics[width=12em]{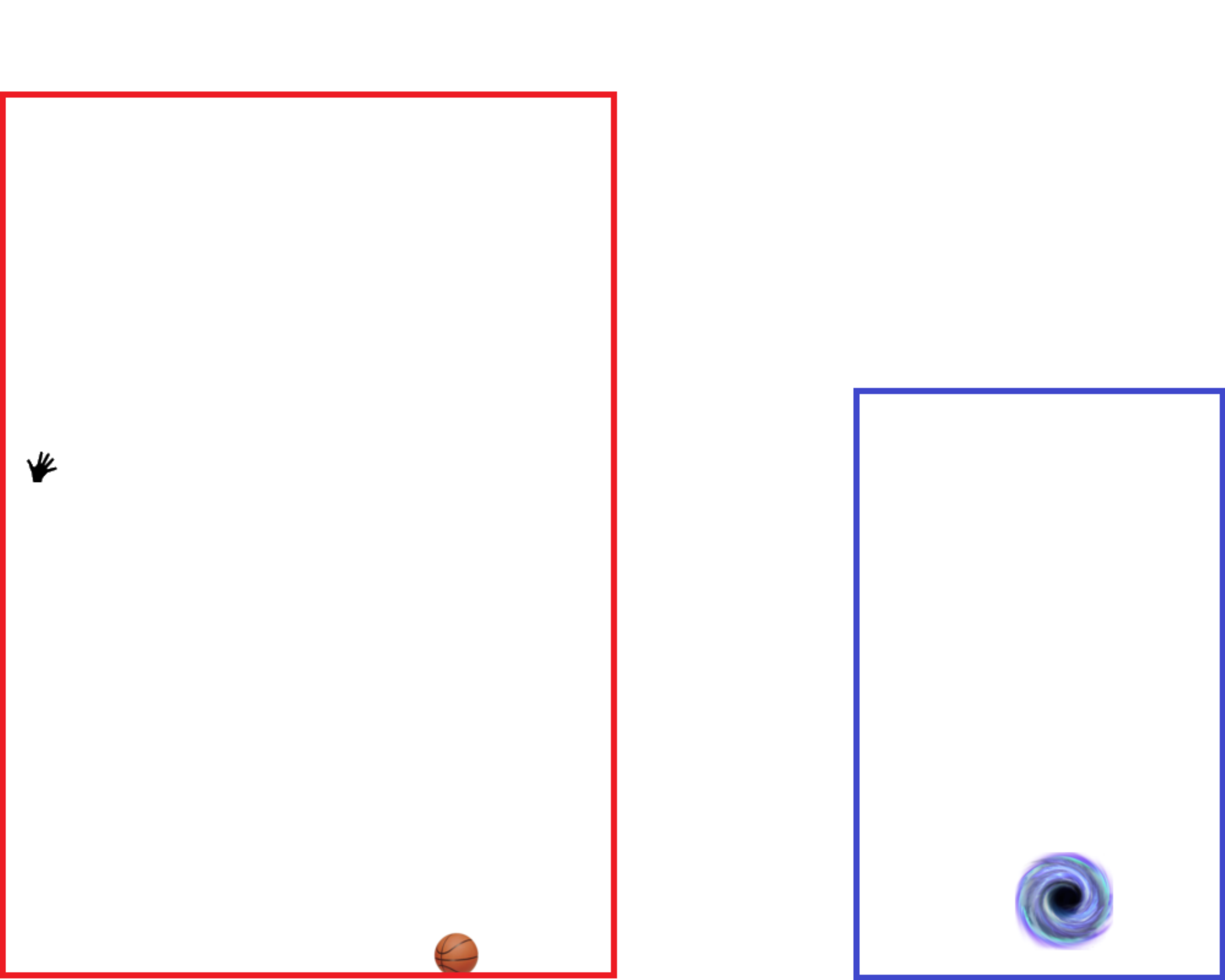}} \label{fig:hand-v1}}
		& \subfloat[]{\tcbox{\includegraphics[width=12em]{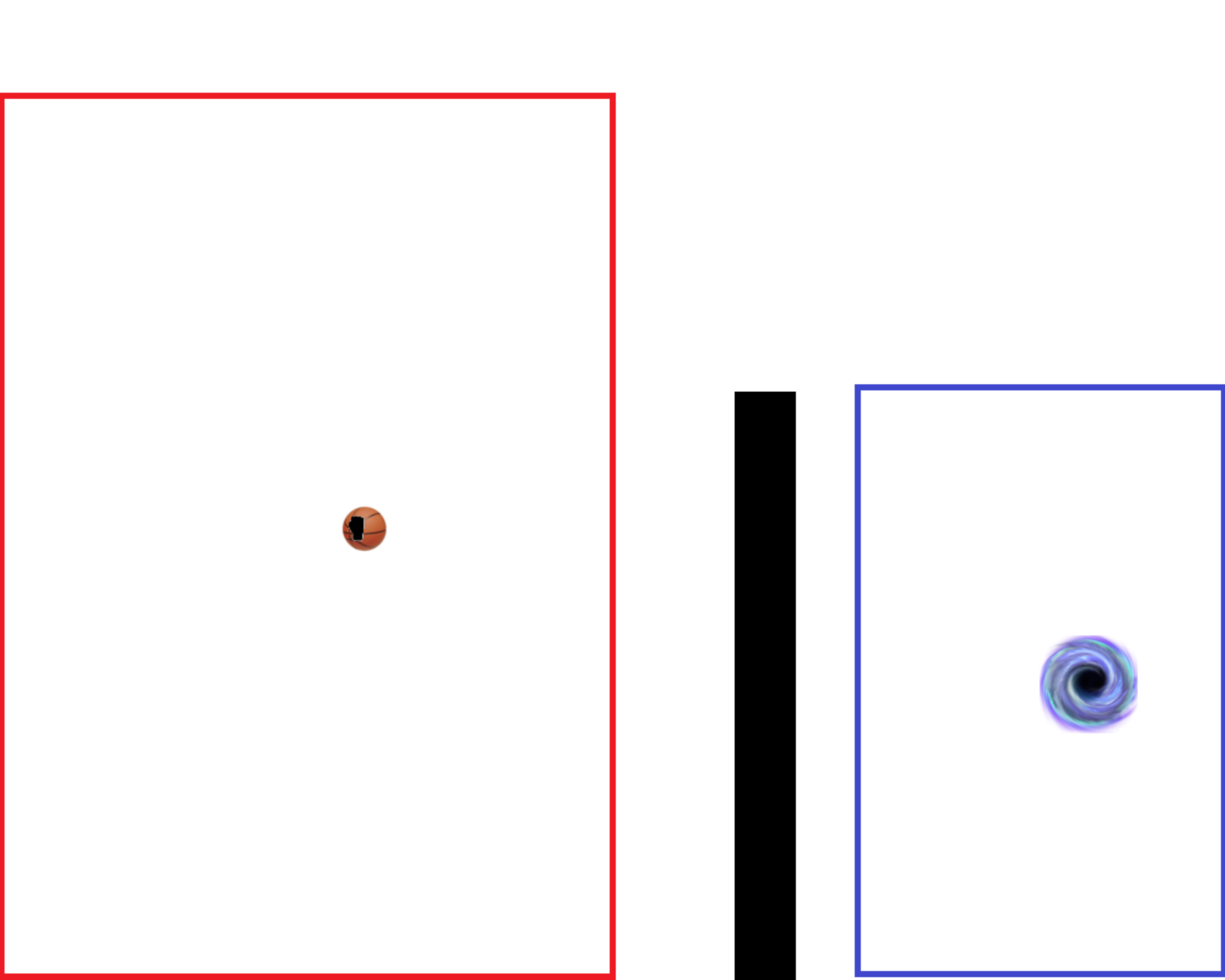}} \label{fig:hand-wall-v0}} \\ 
		\subfloat[]{\tcbox{\includegraphics[width=12em]{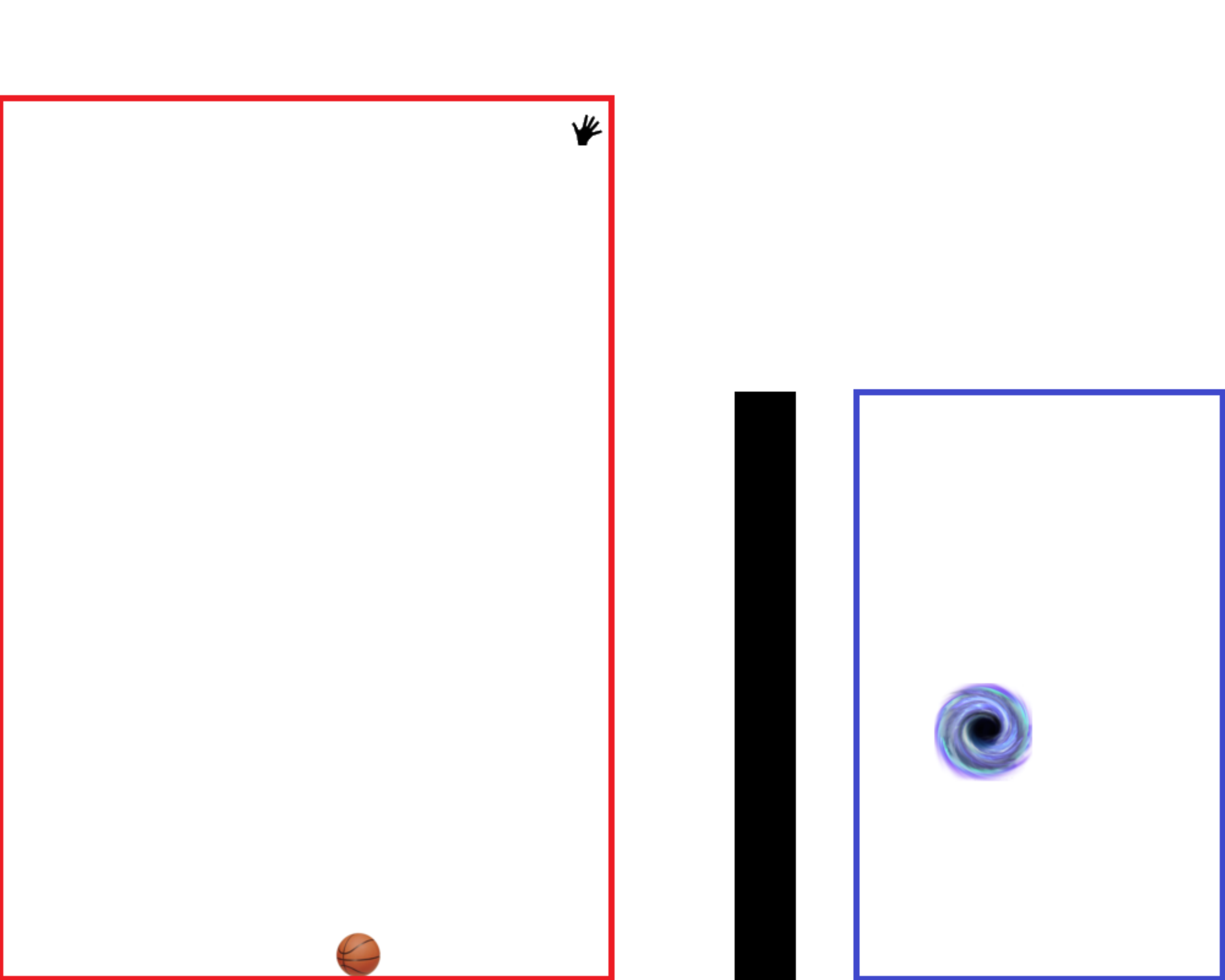}} \label{fig:hand-wall-v1}} 
		& \subfloat[]{\tcbox{\includegraphics[width=12em]{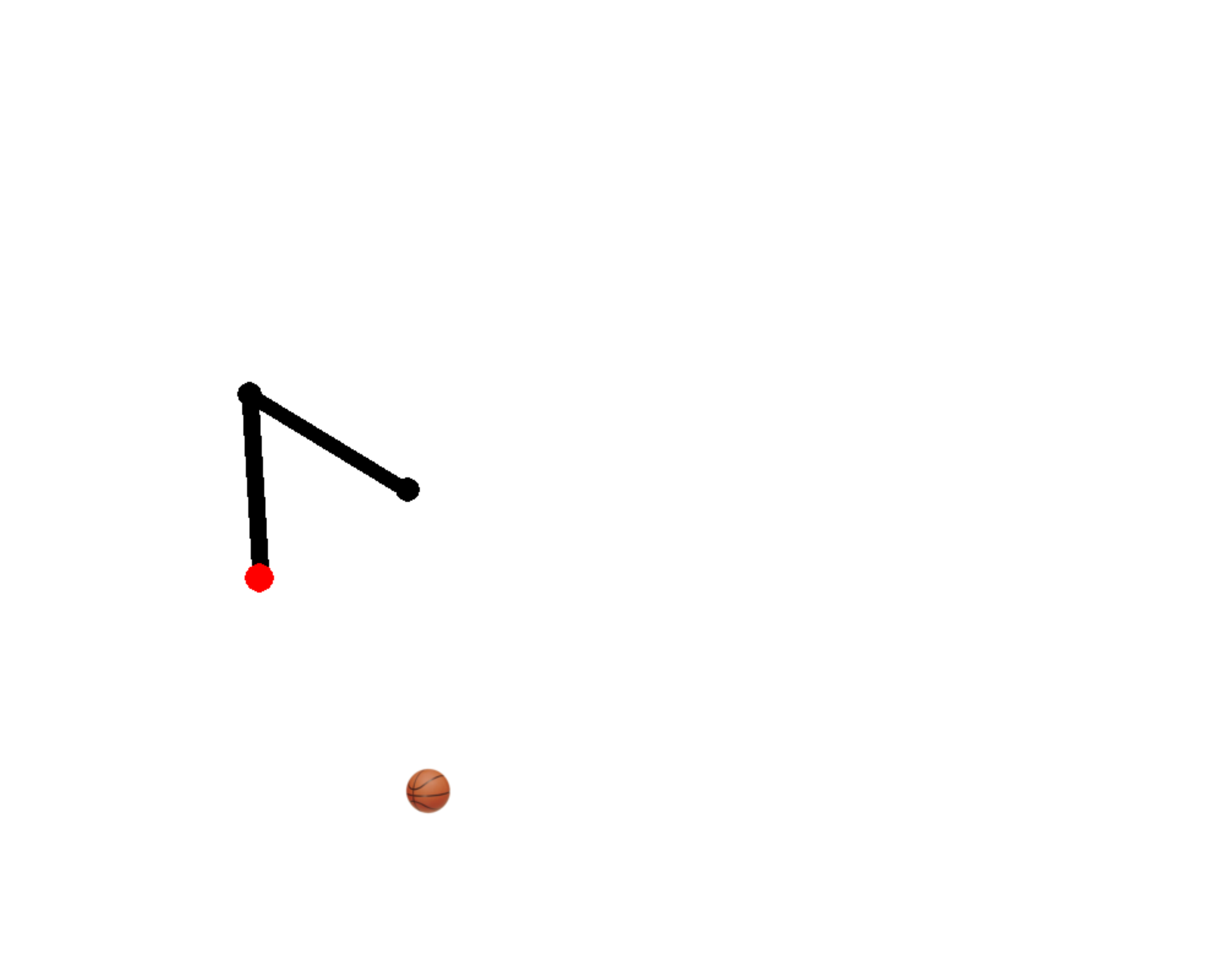}} \label{fig:robot-reach}}\\ 
		\subfloat[]{\tcbox{\includegraphics[width=12em]{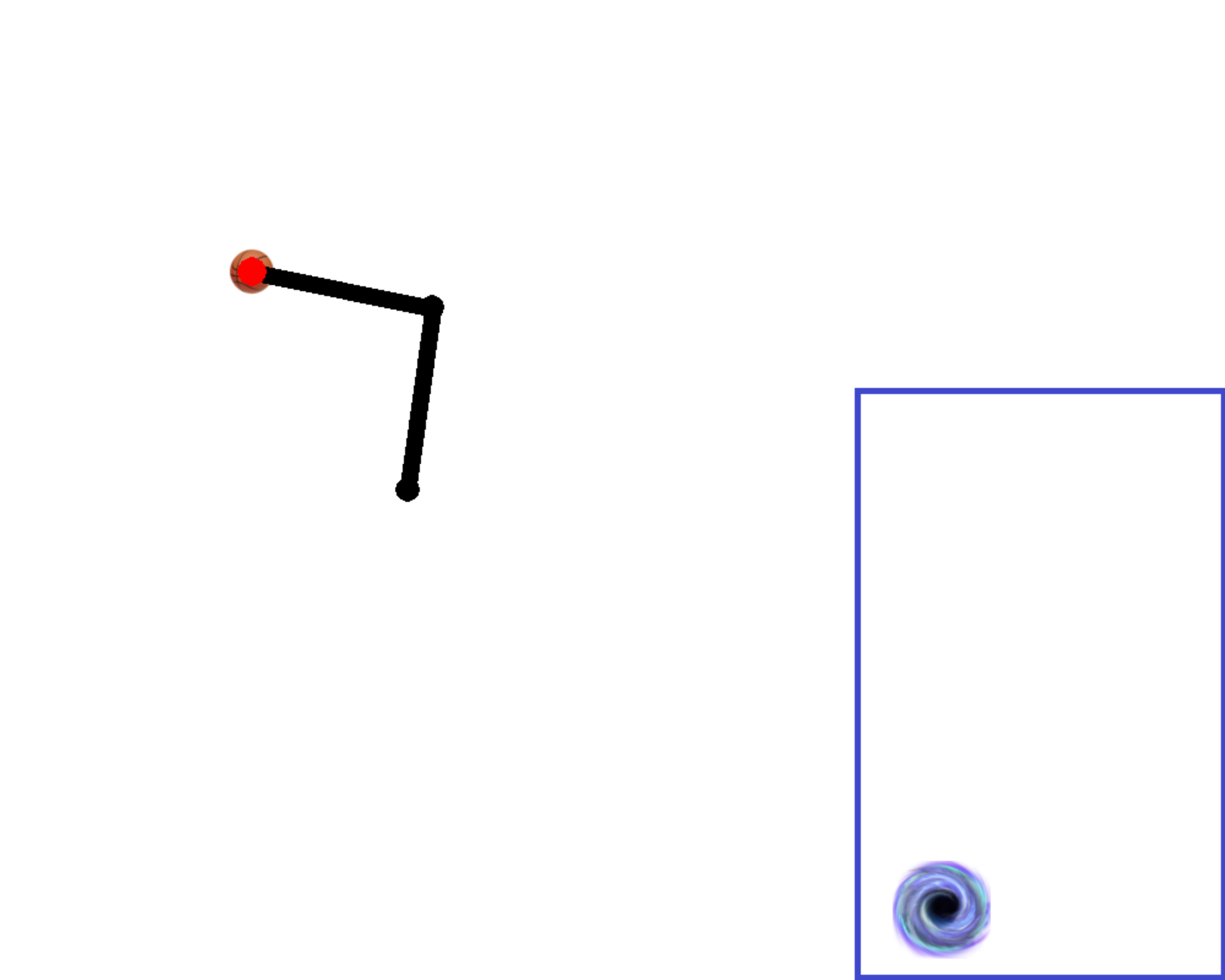}} \label{fig:robot-v0}} 
		& \subfloat[]{\tcbox{\includegraphics[width=12em]{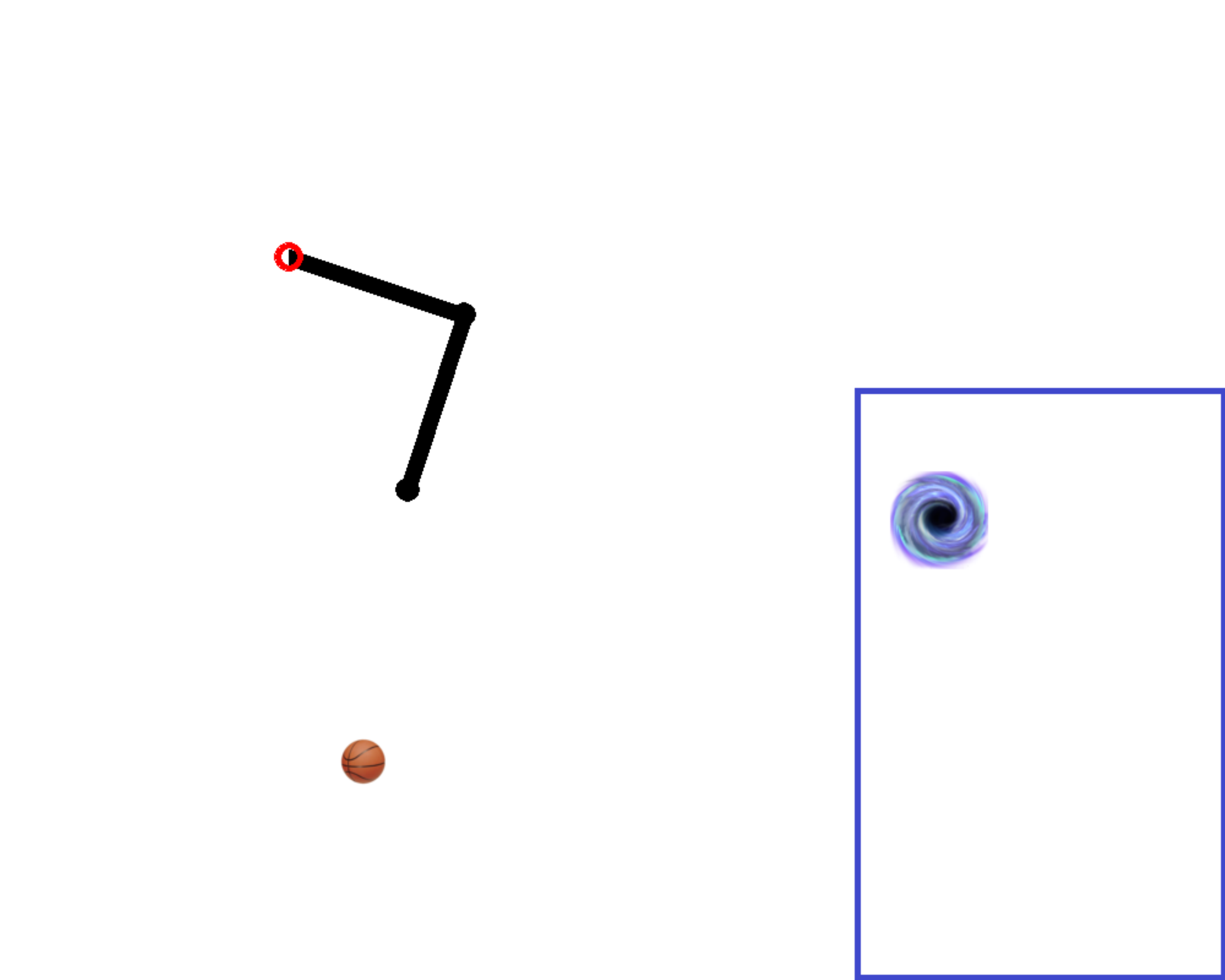}} \label{fig:robot-v1}} \\ 
		\bottomrule 
	\end{tabular}
	\caption[Simulation environments]{\textbf{Simulation environments}: (A) hand\_reach (B) hand\_v0 (C) hand\_v1 (D) hand\_wall\_v0 (E) hand\_wall\_v1 (F) robot\_reach (G) robot\_v0 (H) robot\_v1.\\
	The red and blue rectangles represents the hand's boundaries and goal distribution respectively}
	\end{center}
\end{table} 


\chapter{Virtual Goal Prioritization} 

\label{InstructionalBasedStrategy} 



In this chapter we present our algorithm for virtual goal prioritization, which is using an \textit{Instructional Based Strategy} for selecting virtual goals, and thus, provides an extension of the original HER algorithm.


\section{Motivation}
HER is based on generalizing from previous failures to the desired target. In this chapter, we address the question of how these failures should be taken into consideration in the learning process. Is every failure equally instructive as any other as has been proposed by the original HER algorithm? To analyze this question, consider the following soccer scenario: a player takes two penalty kicks. In a first kick, the goal was missed by a small distance to the right, whereas in a second kick the goal was missed by far to the left. The question arises which of these experiences is more instructive to the soccer player for learning the task of hitting the goal. It seems that nearly missing the goal is more instructive for achieving the goal. However, it might be that the player has experienced many kicks of the first type whereas none of the second. In this case, the latter kick may be more instructive for learning the task. On this \textit{Instructional-Based Strategy} (IBS), we based our heuristic approach towards virtual goal prioritization. 


\section{Method}
Prioritizing virtual goals is guided by three heuristic principles, which define (i) what the agent needs to learn (ii) what the agent can learn from an individual virtual goal and (iii) what is unknown to the agent. The first two principles define the relevance of the sample. Combining with the last principle (which represents inventiveness), they define the \textit{instructiveness} of the virtual goal.\\
For the prioritization we considered two architectures:
\paragraph{Sequential prioritization:} Choose samples to the buffer using the relevance of the sample, then sample them to the mini-batch using PER (PER approximates the inventiveness of the sample). See algorithm \ref{alg:IBS_arch1} for a more formal description of this architecture.
\begin{algorithm}[H]
	\caption{Prioritize for mini-batch selection} \label{alg:IBS_arch1}
	\begin{algorithmic}[1]
		\While{True}
		\For{$Episode \leftarrow 1, M$}
		\State play episode
		\State Store episode in buffer $<S||g, A, R, S'||g>$ \Comment experience replay
		
		\For{$t \leftarrow 0, T-1$}
		\State Sample virtual goals $\tilde{G}$ based on an \textbf{relevance} measure
		\For{$\widetilde{g} \in \tilde{G}$}
		\State $\widetilde{r} := r(s_{t+1}, \widetilde{g})$
		\State Store the transition $(s_t||\widetilde{g}, a_t, \widetilde{r}, s_{t+1}||\widetilde{g})$ in buffer\Comment HER
		\EndFor
		\EndFor
		\EndFor
		
		\For{$t \leftarrow 1, N$}
		\State Sample a mini-batch $B$ from the replay buffer using \textbf{PER}
		\State Perform one step of optimization using mini-batch $B$
		\EndFor
		\EndWhile
	\end{algorithmic}
\end{algorithm}
This architecture has led to poor empirical results (see experiments is section \ref{s:IBS experiments}) due to a theoretical flaw - when the agent picks virtual goals it does not consider what is already known, thus the agent keeps flooding the buffer with samples from the same area. Then, when sampling for the mini-batch, all the experience is from the same area. In other words,\\
$max_g[relevance(g)*inventiveness(g)] \ne max_g[inventiveness(max_g[relevance(g)])]$.\\
Hence we built a new architecture:
\paragraph{Simultaneous prioritization:} Choose samples to the buffer using the instructiveness of the sample (use simultaneously all three principles from the beginning). See algorithm \ref{alg:IBS_arch2} for a more formal description of this architecture.

\begin{algorithm}[H]
	\caption{Prioritize for virtual goal generating} \label{alg:IBS_arch2}
	\begin{algorithmic}[1]
		\While{True}
		\For{$Episode \leftarrow 1, M$}
		\State play episode
		\State Store episode in buffer $<S||g, A, R, S'||g>$ \Comment experience replay
		
		\For{$t \leftarrow 0, T-1$}
		\State Sample virtual goals $\tilde{G}$ based on an \textbf{instructiveness} measure  
		\For{$\widetilde{g} \in \tilde{G}$}
		\State $\widetilde{r} := r(s_{t+1}, \widetilde{g})$
		\State Store the transition $(s_t||\widetilde{g}, a_t, \widetilde{r}, s_{t+1}||\widetilde{g})$ in buffer\Comment HER
		\EndFor
		\EndFor
		\EndFor
		
		\For{$t \leftarrow 1, N$}
		\State Sample a mini-batch $B$ from the replay buffer
		\State Perform one step of optimization using mini-batch $B$
		\EndFor
		\EndWhile
	\end{algorithmic}
\end{algorithm}

\subsection{Definitions}
In this sub-section we elaborate on the three principles defining the \textit{instructiveness} of the virtual goal.
\paragraph{What the agent needs to learn: \label{par:need-to-learn}}The task of the agent is to learn the behavior, which achieves the actual goals. The goals are described by a goal distribution $g(\bm{x})$, where ${\mbox{Pr}}(\bm{x} \in {{G}})= \int_{{G}}g(\bm{x})d\bm{x}$ for any measurable set $G \in {\cal{G}}$. In most cases the goal distribution can be described by a uniform distribution over the range ${\cal{G}}$:
\bea \label{eq:uniform}
\label{uniform} 
g(\bm{x}) = \begin{cases}
	const\,, & \mbox{if} \; \bm{x} \in {\cal{G}} \label{1}\\
	0\,, & \mbox{otherwise}\,.
\end{cases}
\eea 

\paragraph{What can be learned from a virtual goal: \label{par:can-learn}} The selection of a virtual goal $\tilde{\bm{g}}$ teaches the agent of how to reach that goal as well as goals that are in the near surrounding of $\tilde{\bm{g}}$. The latter is due to the generalization capabilities of the underlying neural networks \cite{schaul2015universal,zhang2016understanding}. To approximate the relevance of the virtual goal $\tilde{\bm{g}}$ to other neighboring goals $\tilde{\bm{g}}'$, we use a Gaussian radial basis function (RBF) kernel, $k(\tilde{\bm{g}},\tilde{\bm{g}}')$.

Thus, the relevance of virtual goal $\tilde{\bm{g}}$ to point $\tilde{\bm{g}}'$ is defined by the Mahalanobis distance, where the covariance matrix is set to $\bm{\Sigma} = \sigma^2 \bm{I}$ and the variance $\sigma^2$ is a hyperparameter that can be tuned to maximize the performances. Using kernel regression we score virtual goals given the goal distribution by
\bea \label{eq:value}
\mu({\tilde{\bm{g}}}|\bm{\Sigma})= \int_{\bm{x} \in \mathbb{R}^n}
k(\tilde{\bm{g}},\bm{x}) g(\bm{x}) d\bm{x}\,.
\eea
For a uniform goal distribution equation (\ref{eq:value}) simplifies to
\bea \label{eq:simplify_weight}
\mu({\tilde{\bm{g}}}|\bm{\Sigma},{\cal{G}})= const \cdot \int_{\bm{x} \in {\cal{G}}} k(\tilde{\bm{g}},\bm{x})d\bm{x}\,.
\eea
Thus, virtual goals that are closer to the goal distribution center receive a higher score. For this reason this strategy will not work for environments where the initial state distributions is within the goal distribution, because initial states will get the highest scores. Scores can be turned into a probability distribution over the possible virtual goals by normalization, resulting in the target distribution $q^\ast$ of virtual goals
\bea \label{eq:optVG}
q^\ast({\tilde{\bm{g}}}|\bm{\Sigma},{\cal{G}}) = \frac{\mu({\tilde{\bm{g}}}|\bm{\Sigma},{\cal{G}})}{\int_{\tilde{\bm{g}} \in \tilde{\cal{G}}} \mu({\tilde{\bm{g}}}|\bm{\Sigma},{\cal{G}})d \tilde{\bm{g}}}\,,
\eea
where $ \tilde{\cal{G}}$ denotes the range of all virtual goals. 

\paragraph{What is unknown to the agent: \label{par:unknown}} This principle is implemented differently for each architecture. For the first architecture, this principle is represented by the usage of PER in the mini-batch sampling. For the second architecture it is more elaborated. The agent's current knowledge about the goal distribution is represented by the proposal distribution $q(\tilde{\bm{g}})$ of virtual goals and is initialized with zero. The mismatch between the proposal and target distribution is calculated using the local difference:
\bea \label{eq:weight2}
w(\tilde{\bm{g}})=clip\big[q^*(\tilde{\bm{g}})-q(\tilde{\bm{g}}), min=0\big]\,,
\eea
which by normalization leads to the probability used for prioritization
\bea \label{eq:priority}
p(\tilde{\bm{g}}) = \frac{w(\tilde{\bm{g}})} {\sum_{\tilde{\bm{g}}\in \tilde{\cal{G}}} w(\tilde{\bm{g}}') }\,, \forall \, \tilde{\bm{g}} \in \tilde{\cal{G}} \,,
\eea
where $\tilde{{\cal{G}}}$ denotes the set of virtual goals. In practice, we find it useful to clip the weights to some small value larger than zero (we used $0.002$), so that all virtual goals have some probability of getting sampled. This trick makes learning more stable.

\subsection{Implementation}
For the implementation of the algorithm we discretize the range of virtual goals $\cal{\tilde{G}} = \cal{S}$ into $M\times N$ grid cells and approximate the target and proposal distributions of virtual goals over the grid cells:
\bea \label{eq:discrete_optVG}
q^\ast_{ij}(\bm{\Sigma},{\cal{G}}) &=& \frac{\mu((i,j)|\bm{\Sigma},{\cal{G}})}{ \sum_{i,j=1}^{M,N}\mu((i,j)|\bm{\Sigma},{\cal{G}})}\,,\quad i=1,\dots,M\,, j=1,\dots, N\,,
\eea
\bea \label{eq:discrete_VG}
q_{ij} &=& \frac{1}{|R|}\sum_{\tilde{\bm{g}} \in R} [ \tilde{\bm{g}} \in \mbox{cell}(i,j)]\,,\quad i=1,\dots,M\,, j=1,\dots, N\,, 
\eea
where $(i,j)$ denotes the center of the grid cells, $[\cdot]$ is the indicator function and $R$ the replay buffer of virtual goals with size $|R|$. To stabilize the learning, we start with a high variance, $\Sigma = \sigma^2 I$, and gradually decrease it to its final value with the decay schedule $\Sigma \leftarrow 0.9\cdot\Sigma$. The weight of the virtual goal $\tilde{\bm{g}}$ is the weight of its bin
\bea \label{eq:discrete_weight}
w(\tilde{\bm{g}}) &=&
clip\big[q^\ast_{bin(\tilde{\bm{g}})}(\bm{\Sigma},{\cal{G}})-{q_{bin(\tilde{\bm{g}})}} \,,\, min=0\big]\,,
\eea
and the prioritization probability is defined as in equation (\ref{eq:priority}). See Alg.\ref{alg:IBS_HER} for a more formal description of the algorithm.

\begin{algorithm}
	\caption{Instructional-Based HER} \label{alg:IBS_HER}
	\begin{algorithmic}[1]
		\Require
		\Statex \textbullet~ an off-policy RL algorithm $\mathbb{A}$,\Comment e.g. DQN, DDPG 
		\Statex \textbullet~ a reward function: $\mathcal{S}\times \mathcal{A}\times \mathcal{G} \rightarrow \mathcal{R}$,\Comment e.g. $r(s,a,g)=-1 \text{ if fail, } 0 \text{ if success}$
		\Statex \textbullet~ [real] goal distribution $\mathcal{G}$
		\Statex \textbullet~ std $\sigma$ for the target distribution $q$
		\Statex
		
		\State Initialize $\mathbb{A}$
		\State Initialize replay buffer $R$
		\State Initialize $q$ \Comment $q_{ij}=0 \quad\forall i,j \in [1\ldots M],[1\ldots N], \quad |R| \leftarrow 0$
		\State Calculate $q^*$ \Comment Using equation (\ref{eq:discrete_optVG})
		\While{True}
		\For{$Episode \leftarrow 1, M$}
		\State Sample a goal $g$ and an initial state $s_0$.
		\For{$t \leftarrow 0, T-1$}
		\State Sample an action $a_t$ using the behavioral policy from $\mathbb{A}$: \par
		$\qquad \qquad a_t \leftarrow \pi(s_t||g)$ \Comment || denonts concatenation
		\State Execute the action $a_t$ and observe a new state $s_{t+1}$
		\EndFor
		
		\For{$t \leftarrow 0, T-1$} \Comment IBS
		\State Calculate the priority $p(\widetilde{g}_t)$ via equation (\ref{eq:priority})
		
		\EndFor
		
		\For{$t \leftarrow 0, T-1$}
		\State $r_t := r(s_{t+1}, g)$
		\State Store the transition $(s_t||g, a_t, r_t, s_{t+1}||g)$ in $R$ \Comment standard experience replay
		
		\State Sample a set of virtual goals $\tilde{G}$ for replay from the future state based on priority $p^*(\widetilde{g})$ 
		\For{$\widetilde{g} \in \tilde{G}$}
		\State $\widetilde{r} = r(s_{t+1}, \widetilde{r})$
		\State Store the transition $(s_t||\widetilde{g}, a_t, \widetilde{r}, s_{t+1}||\widetilde{g})$ in $R$\Comment HER
		\State Update $q$
		\State $|R| \leftarrow |R| + 1$
		\EndFor
		\EndFor
		\EndFor
		
		\For{$t \leftarrow 1, N$}
		\State Sample a minibatch $B$ from the replay buffer $R$
		\State Perform one step of optimization using $\mathbb{A}$ and minibatch $B$
		\EndFor
		\EndWhile
		
	\end{algorithmic}
\end{algorithm}


\section{Related Work}
Prioritizing samples over their relevance to the learning has been used in both \textit{Prioritized Experience Replay} (PER) \cite{schaul2015prioritized} and \textit{Energy-Based Hindsight Experience Prioritization} (EBP) \cite{zhao2018energy}.
Similar to our algorithm, PER gives higher priority to samples that are unknown to the agent. However, unlike IBS, PER uses the \textit{TD-Error} of the sample to measure the agent's knowledge (i.e., a smaller error implies more acquaintance). PER receives the buffer as a given input set and prioritizes when sampling from it for experience replay. In contrast, IBS prioritizes when building the buffer during experiences. Another difference is that unlike IBS, PER only prioritizes over unfamiliar samples and does not take into consideration that some samples might be better towards task completion than others. \\
EBP applies a different prioritization scheme by calculating the amount of (translational and rotational) kinetic energy transferred to the object during an episode. Trajectories associated with a larger kinetic energy transfer are therefore preferred, assuming that the agent can learn more from trajectories in which the object moved significantly. EBR does not differentiate between movement directions and is thus applicable for cases where all directions are equally informative for learning. Similar to PER, EBP receives the buffer as a given input set and prioritizes when sampling from it for experience replay. Since PER and EBP prioritize during experience replay, both methods can be applied with IBS.


\section{Experiments} \label{s:IBS experiments}
In this section, we evaluate the Instructional Based Strategy. 
We tested our algorithm on the following environments:
\begin{enumerate}
\item \textbf{Hand\_v0} - The simplest version of the Hand task, where the ball is initialized within the hand
\item \textbf{Hand\_wall\_v0} - The simplest version of the Hand-Wall task, where the ball is initialized within the hand
\item \textbf{Robot\_v0} - The simplest version of the Robot task, where the ball is initialized within the end-effector
\end{enumerate}
\noindent Training is performed using the DDPG algorithm \cite{lillicrap2015continuous}, in which the actor and the critic were represented using multi-layer perceptrons (MLPs). See Appendix \ref{AppendixB} for more details regarding networks architecture and hyperparameters. In order to test the performance of the algorithms, we ran on each environment the \textbf{vanilla-HER} and \textbf{HER} with \textbf{IBS}. For the Hand and Robot tasks we set $\sigma = 0.2$ (equation \ref{eq:value}), while for the Hand-Wall we set $\sigma = 0.1$. In all algorithms we used prioritized experience replay (PER) \cite{schaul2015prioritized}. The results of the algorithms are evaluated using three criteria:
\begin{itemize}
	\item Virtual goal distributions
	\item Success rate
	\item Distance-to-goal
\end{itemize}
The first criterion analyzes the differences in virtual goal selection for the different algorithms. 
The second and third criteria evaluate the performances of the agent.

\subsubsection{Virtual Goal Distributions}
\begin{figure}
	\centering
	\includegraphics[width=15em]{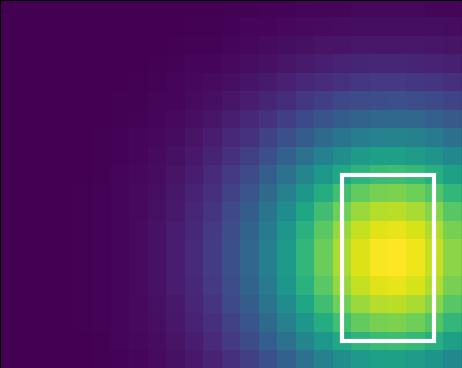}
	\caption[IBS - target distribution]{Target Distribution. The target distribution is calculated for $\sigma=0.2$ (screen size is $1\times1 $ in dimensionless units).}
	\label{fig:IBS - target_dist}
\end{figure}

We compare the virtual goals distributions generated from the different algorithms to the target distribution $q^\ast$ (Fig.\ref{fig:IBS - target_dist}) generated by equation (\ref{eq:optVG}). Table \ref{tbl:IBS - VG_dists} shows the effect of the different virtual goal selection strategies and the resulting distributions. Each figure shows the mean height at each point, over all the runs. The virtual goal distribution generated by \textit{HER-IBS} is closer to the target distribution as indicated by the KL distance in Table \ref{tbl:IBS - KL_distance}.

\begin{table}[H]
	\centering
	\begin{tabular}{m{0em}*3{C}}
		\toprule
		& \textbf{HER} & \textbf{HER-IBS}\\
		\midrule
		\rotatebox{90}{\textbf{Hand}} & 
		\includegraphics[width=8em]{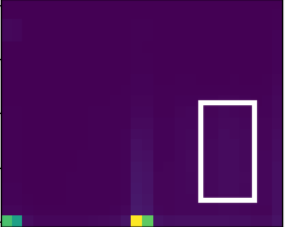} &
		\includegraphics[width=8em]{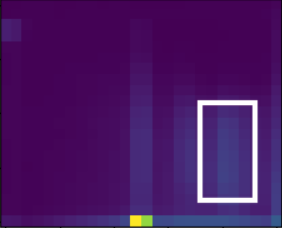}
		\\
		\cmidrule(lr{1em}){1-3}
		\rotatebox{90}{\textbf{Hand-Wall}} & 
		\includegraphics[width=8em]{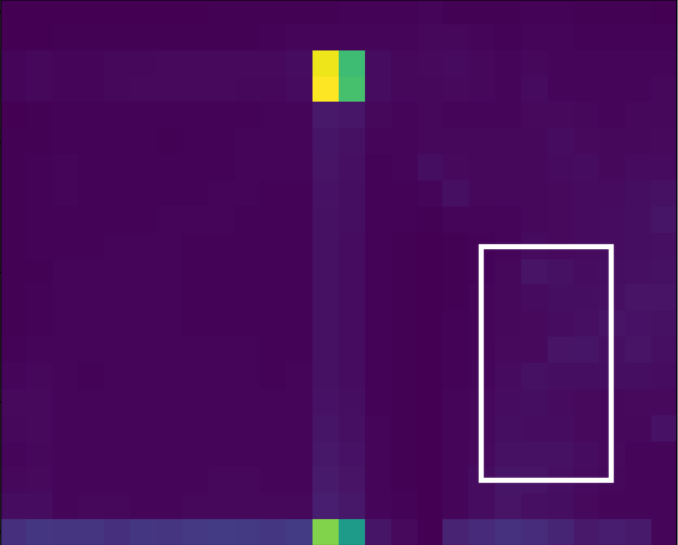} & 
		\includegraphics[width=8em]{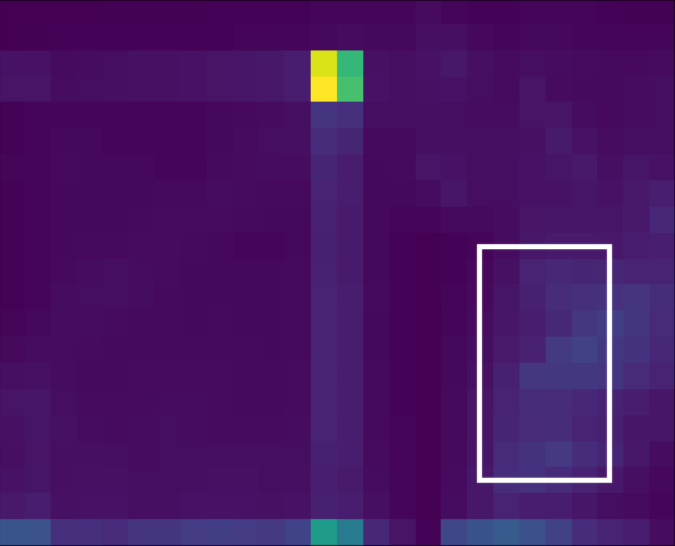}
		\\
		\cmidrule(lr{1em}){1-3}
		\rotatebox{90}{\textbf{Robot}} & 
		\includegraphics[width=8em]{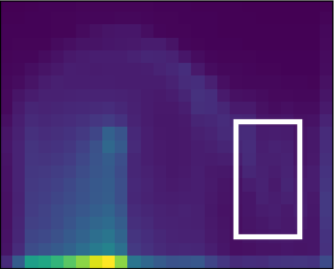} & 
		\includegraphics[width=8em]{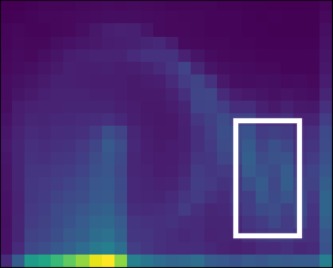}
		\\
		\bottomrule
	\end{tabular}
	\caption[IBS - Comparison between proposal and target distribution of virtual goals]{Comparison between proposal and target distribution of virtual goals}
	\label{tbl:IBS - VG_dists}
\end{table}

\begin{table}[H]
	\centering
	\resizebox{0.6\textwidth}{!}{
	\begin{tabular}{ccc}
		\toprule
		& HER & HER-IBS \\
		\cmidrule(lr){2-3}
		
		\textbf{Hand} &0.82727&\textbf{0.38977}\\
		\textbf{Hand-Wall} &1.1932&\textbf{0.8656}\\
		\textbf{Robot} &1.16236&\textbf{0.7134}\\
		\bottomrule
	\end{tabular}}
	\caption[IBS - KL distance]{KL Distance \label{tbl:IBS - KL_distance}}
\end{table}

\subsubsection{Success Rate and Distance from Goal}
\begin{figure}[H]%
	\centering
	\subfloat[Success rate]
	{{\includegraphics[width=0.7\textwidth]{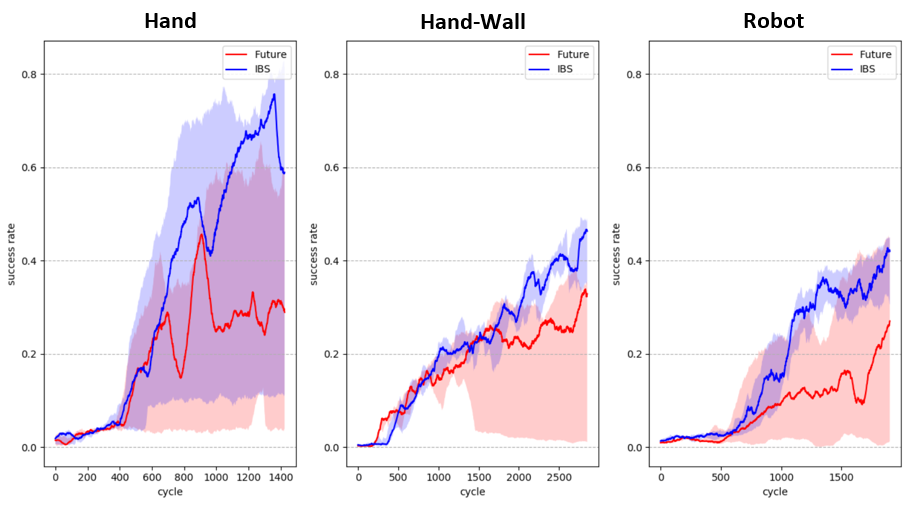} 
	\label{fig:IBS - Success rate}}} \\
	\subfloat[Distance from goal]
	{{\includegraphics[width=0.7\textwidth]{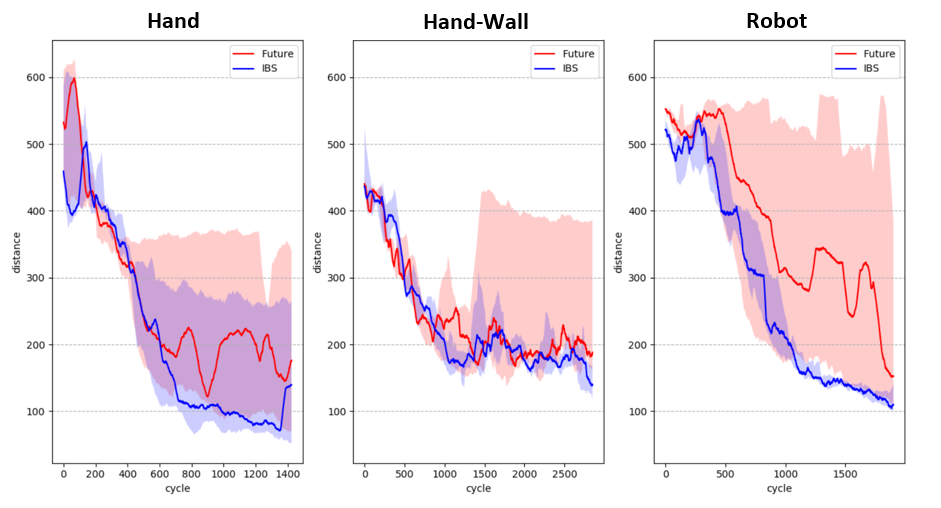} 
	\label{fig:IBS - Distance from goal}}}%
	\caption[IBS - Performance]{Learning curves for the multi-goal tasks. Results are shown over 15 independent runs. The bold line shows the median, and the light area indicates the range between the 33rd to 67th percentile.}
	\label{fig:IBS - performance}%
\end{figure}

As shown in Fig.\ref{fig:IBS - Success rate} and \ref{fig:IBS - Distance from goal}, the HER-IBS algorithm outperforms the vanilla-HER in all tasks. Furthermore, In both Hand-Wall and Robot tasks, the performances are significantly more consistent, as indicated by the shaded area, which represents the 33rd to 67th percentile of the performances.

\section{Conclusion}
In this chapter, we introduced a novel strategy for virtual-goals prioritization, called IBS. IBS improves HER's robustness and performances in two out of the three tasks we tested it on, but is still limited to simple tasks where the agent can easily manipulate the achieved-goal. We published the algorithm and performances described in this chapter in the following article \cite{manela2019bias}. In the next chapter, we introduce a new technique that solves this problem.

\chapter{Bias-reduced HER } 

\label{FilteredHER} 



In this chapter we present the motivation, theoretical background and performances of our new method to reduce bias in HER. As we apply a filtering technique we denoted our algorithm as \textit{Filtered-HER}.


\section{Motivation}
In this chapter, we discuss a fundamental problem within the original HER algorithm. As mentioned in \cite{plappert2018multi}, HER may insert bias to the learning process. Using the \textit{achieved-goal} as a virtual goal may lead in some cases to situations in which the agent performs poorly even though the agent receives repeatedly rewards indicating that it should continue to act in this way. consider the bit-flipping environment from \cite{andrychowicz2017hindsight}: The state- and action spaces are $S = \{0, 1\}^n$ and $A = \{0,1,...,n-1\}$ respectively for some length $n$. Executing the i-th action flips the i-th bit of the state. The initial and target states are sampled uniformly at the beginning of each episode, and each step has a cost of -1. To illustrate HER's problem, we add new action that has no effect and then terminates the game. It is clear that this action is useless, but the agent might think otherwise. Since the state stays the same, the virtual goal of this state will always be the state itself, thus the virtual reward of this action will always be positive (zero). As a result,
the agent might think this action is desired. Although this scenario may seem implausible, it happens frequently in manipulation tasks, for example, in the \textit{Push} task of OpenAI Gym. In this environment, a manipulator needs to push a box to a desired location. If the manipulator does not touch the box, the achieved-goal (i.e., the box position) will not change. Hence, when virtual goals for experience replay are sampled, they all will be the same and identical to all the achieved-goals and this will result in misleading positive virtual rewards\\ 

\section{Bias in Traditional Reinforcement Learning}
\begin{figure}[h]%
	\centering
	\subfloat[Game]{{\includegraphics[width=0.5\textwidth]{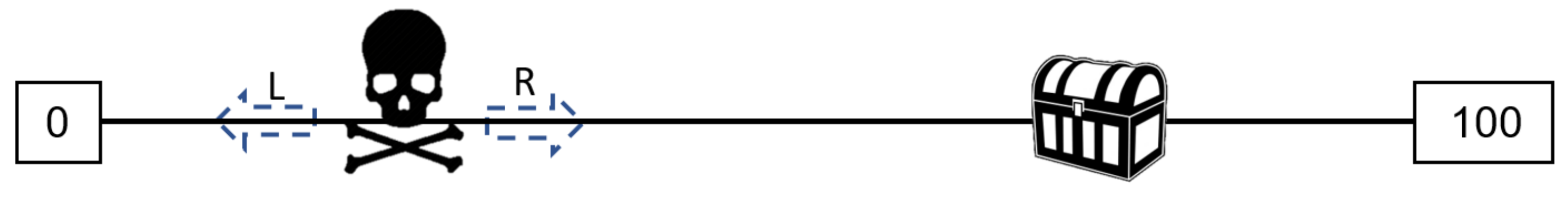} 
	\label{fig:pirate}}}
	\subfloat[Optimal Policy]{{\includegraphics[width=0.5\textwidth]{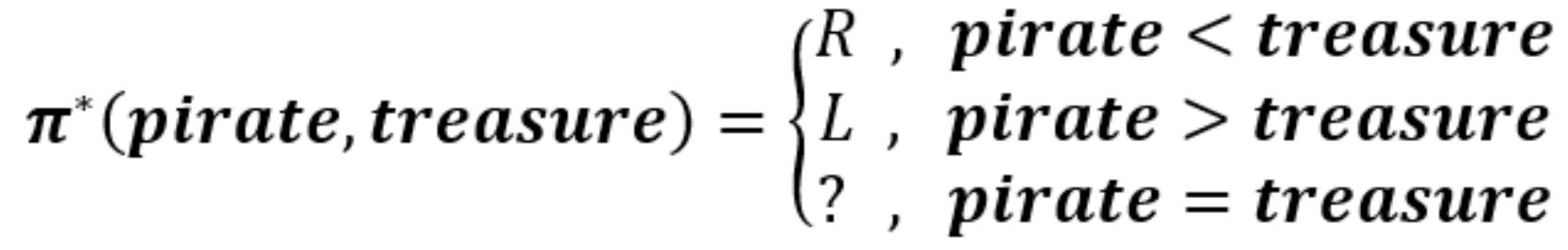} 
	\label{fig:pirates_policy}}}%
	\caption[Treasure Hunting game]{Treasure Hunting game: (a) In every episode, a treasure box is located randomly somewhere on a line between 0 to 100. The agent, a pirate, is also located at a random place within this range. The pirate's goal is to reach the treasure box using the left and right actions. (b) The pirate's optimal policy in the treasure hunting game. No meaningful action can be assigned in the terminal state.}
	\label{fig:pirate_example}%
\end{figure}

This drawback of HER is similar to the role of terminal states in bootstrapping, in which the values of all states are gradually updated except for terminal states. Terminal states are, by definition, states for which the achieved goal is identical to the desired goal. However, no actions are assigned to terminal states in bootstrapping, nor is any next-state observed (i.e., a tuple S,A,R,S$^\prime$), because assigning actions to terminal states will disturb the learning process. To illustrate the problem in more detail, consider the following treasure hunting game: for every episode of the game, a treasure box is placed randomly in a one-dimensional world. The pirate is also located at a random position, and his goal is to reach the treasure box by using the left and right actions (Figure \ref{fig:pirate}). The pirate's optimal policy is straightforward and shown in Figure \ref{fig:pirates_policy}. Assigning a \textit{stay} action to states which are equal to the goal is similar to memorizing irrelevant material for a coming exam. The irrelevant material is not wrong, but it is unnecessary and may make convergence harder, particularly when using function approximation methods such as neural networks. However, the latter is an inherent feature of the HER algorithm, and thus, will generate misleading samples.

\section{Method}
To resolve this problem, we apply a filter to remove misleading samples. Before storing the virtual sample in the replay buffer, the filter checks if the virtual goal has been already achieved in the current state. If so, the sample will be deleted, and the next virtual goal will be generated. See Alg. \ref{alg:Filtered-HER} for the pseudo-code of the \textit{Filtered-HER} algorithm. 

\begin{algorithm}
	\caption{Filtered-HER} \label{alg:Filtered-HER}
	\begin{algorithmic}[1]
		\Require
		\Statex \textbullet~ an off-policy RL algorithm $\mathbb{A}$,\Comment e.g. DQN, DDPG 
		\Statex \textbullet~ a reward function: $\mathcal{S}\times \mathcal{A}\times \mathcal{G} \rightarrow \mathcal{R}$,\Comment e.g. $r(s,a,g)=-1 \text{ if fail, } 0 \text{ if success}$
		\Statex
		
		\State Initialize $\mathbb{A}$
		\State Initialize replay buffer $R$
		
		\While{True}
		\For{$Episode \leftarrow 1, M$}
		\State Sample a goal $g$ and an initial state $s_0$.
		\For{$t \leftarrow 0, T-1$}
		\State Sample an action $a_t$ using the behavioral policy from $\mathbb{A}$: \par
		$\qquad \qquad a_t \leftarrow \pi(s_t||g)$ \Comment || denotes concatenation
		\State Execute the action $a_t$ and observe a new state $s_{t+1}$
		\EndFor
		
		\For{$t \leftarrow 0, T-1$}
		\State $r_t := rs_{t+1}, g)$
		\State Store the transition $(s_t||g, a_t, r_t, s_{t+1}||g)$ in $R$ \Comment standard experience replay
		
		\State Sample a set of virtual goals $\tilde{G}$ for replay $\tilde{G} := \mathbb{S}(\text{\textbf{current episode}})$
		\For {$\widetilde{g} \in \tilde{G}$}
		\State $\widetilde{r} = r(s_{t+1}, \widetilde{g})$
		\If{$r(s_{t}, \widetilde{g})<0$}\Comment \textbf{Filtered-HER}
		\State Store the transition $(s_t||\widetilde{g}, a_t, \widetilde{r}, s_{t+1}||\widetilde{g})$ in $R$\Comment HER
		\EndIf
		\EndFor
		\EndFor
		\EndFor
		
		\For{$t \leftarrow 1, N$}
		\State Sample a minibatch $B$ from the replay buffer $R$
		\State Perform one step of optimization using $\mathbb{A}$ and minibatch $B$
		\EndFor
		\EndWhile
		
	\end{algorithmic}
\end{algorithm}


\section{Experiments}
In this section, we evaluate Filtered-HER and the combination of Filtered-HER with IBS.\\
We tested our algorithms on the following environments:
\begin{enumerate}
\item \textbf{Hand\_v2} - The medium version of the Hand task, where the ball is initialized within the hand with a probability of $0.5$
\item \textbf{Hand\_wall\_v2} - The medium version of the Hand-Wall task, where the ball is initialized within the hand with a probability of $0.5$
\item \textbf{Robot\_v2} - The medium version of the Robot task, where the ball is initialized within the end-effector with a probability of $0.5$
\end{enumerate}
\noindent Training is performed using the DDPG algorithm \cite{lillicrap2015continuous}, in which the actor and the critic were represented using multi-layer perceptrons (MLPs). See Appendix \ref{AppendixB} for more details regarding networks architecture and hyperparameters. In order to test the performance of the algorithms, we ran on each environment all four combinations: \textbf{vanilla-HER}, \textbf{Filtered-HER}, \textbf{HER} with \textbf{IBS} and \textbf{Filtered-HER} with \textbf{IBS}. For all tasks, we set $\sigma = 0.2$ (equation \ref{eq:value}), when IBS is used. In all algorithms we used prioritized experience replay (PER) \cite{schaul2015prioritized}. The results of the algorithms are evaluated using four criteria:
\begin{itemize}
	\item Virtual goal distributions
	\item Success rate
	\item Distance-to-goal
	\item Estimated Q value
\end{itemize}
The first criterion analyzes the differences in virtual goal selection for the different algorithms. 
The second and third criteria evaluate the performances of the agent. 
The fourth criterion evaluate the bias induced with each combination.

\subsubsection{Virtual Goal Distributions}
\begin{figure}[H]
	\centering
	\centering
	\includegraphics[width=15em]{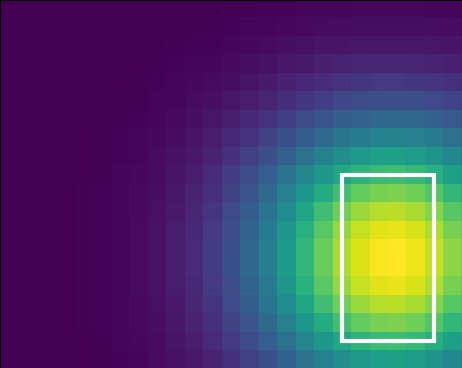}
	\caption[Filtered HER - Target Distribution]{Target Distribution. The target distribution is calculated for $\sigma=0.2$ (screen size is $1\times1 $ in dimensionless units).}
	\label{fig:Filtered HER - target_dist}
\end{figure}

We compare the virtual goals distributions generated from the different algorithms to the target distribution $q^\ast$ (Fig.\ref{fig:Filtered HER - target_dist}) generated by equation (\ref{eq:optVG}). Table \ref{tbl:Filtered HER - VG_dists} shows the effect of different virtual goal selection strategies and the resulting distributions. The virtual goal distribution generated by \textit{Filtered-HER-IBS} is the closest to the target distribution, as indicated by the KL distance in Table \ref{tbl:Filtered HER - KL_distance}. Notice that \textit{Filtered-HER} dramatically reduces the number of samples on the floor ($y=0$) by removing misleading samples.

\begin{table}[H]
	\centering
	\begin{tabular}{m{0em}*5{C}}
		\toprule
		& \textbf{vanilla HER} & \textbf{HER-IBS} & \textbf{Filtered-HER} & \textbf{Filtered-HER-IBS}\\
		\midrule
		\rotatebox{90}{\textbf{Hand}} & 
		\includegraphics[width=8em]{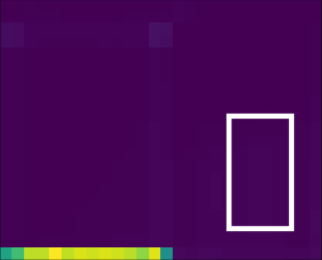} &
		\includegraphics[width=8em]{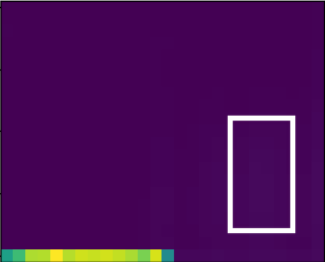}
		& \includegraphics[width=8em]{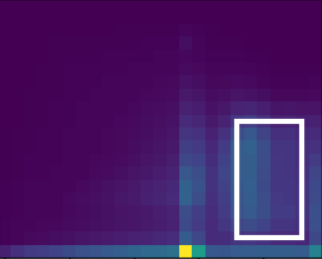} &
		\includegraphics[width=8em]{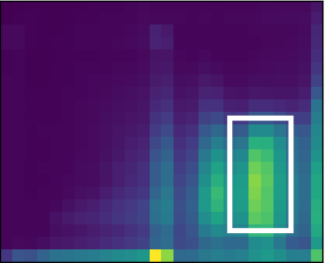} 
		\\
		\cmidrule(lr{1em}){1-5}
		\rotatebox{90}{\textbf{Hand-Wall}} & 
		\includegraphics[width=8em]{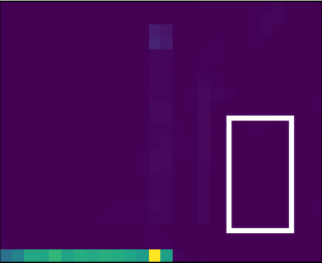} &
		\includegraphics[width=8em]{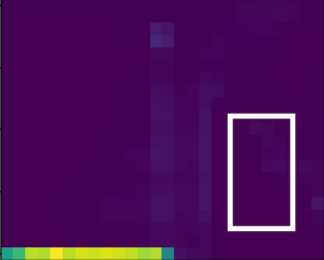}
		& \includegraphics[width=8em]{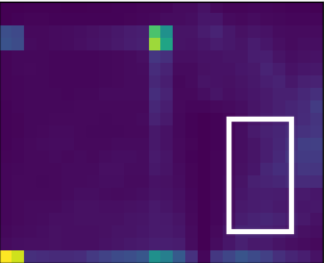} &
		\includegraphics[width=8em]{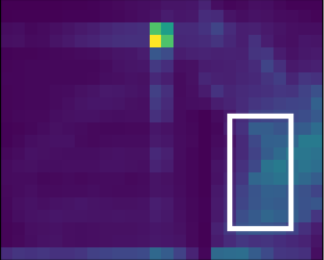} 
		\\
		\cmidrule(lr{1em}){1-5}
		\rotatebox{90}{\textbf{Robot}} & 
		\includegraphics[width=8em]{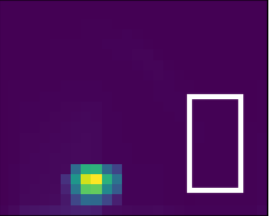} &
		\includegraphics[width=8em]{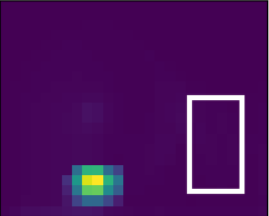}
		& \includegraphics[width=8em]{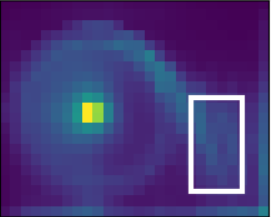} &
		\includegraphics[width=8em]{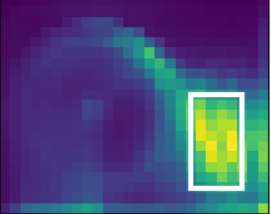} 
		\\
		\bottomrule
	\end{tabular}
	\caption[Filtered HER - Comparison between proposal and target distribution of virtual goals]{Comparison between proposal and target distribution of virtual goals}
	\label{tbl:Filtered HER - VG_dists}
\end{table}

\begin{table}[H]
	\centering
	\begin{tabular}{ccccc}
		\toprule
		& HER & HER-IBS & Filtered-HER & Filtered-HER-IBS \\
		\cmidrule(lr){2-5}
		
		\textbf{Hand} &2.0317&1.6623&0.3436&\textbf{0.2272}\\
		\textbf{Hand-Wall} &5.0574&4.6005&0.9606&\textbf{0.5971}\\
		\textbf{Robot} &2.3201&1.9909&0.8609&\textbf{0.3113}\\
		\bottomrule
	\end{tabular}
	\caption[Filtered HER - KL distance]{KL Distance \label{tbl:Filtered HER - KL_distance}}
\end{table}

\subsubsection{Success Rate and Distance from Goal}
\begin{figure}[H]
	\centering
	\subfloat[Success rate]
	{{\includegraphics[width=0.7\textwidth]{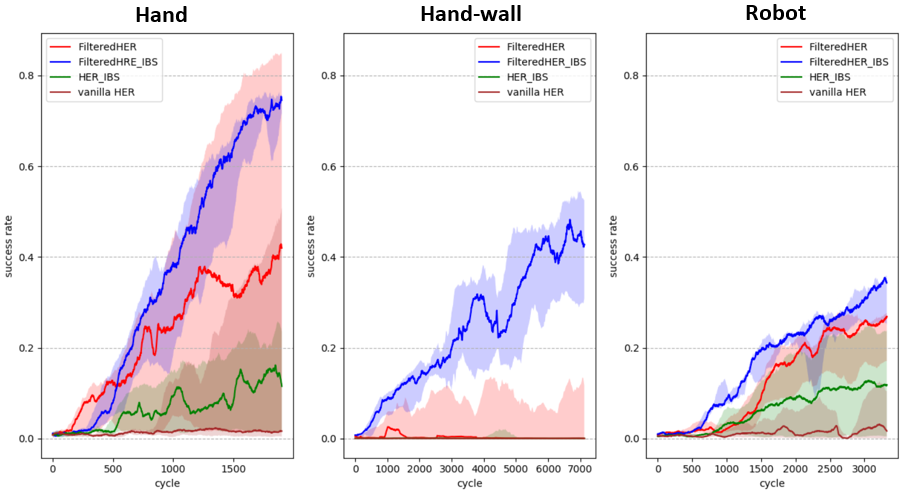} 
	\label{fig:Filtered HER - Success rate}}} \\
	\subfloat[Distance from goal]
	{{\includegraphics[width=0.7\textwidth]{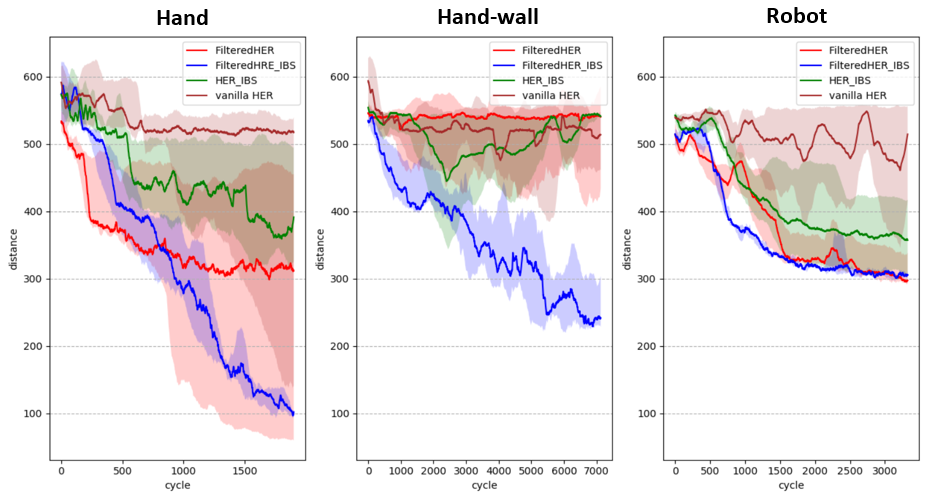} 
	\label{fig:Filtered HER - Distance from goal}}}%
	\caption[Filtered HER - Performance]{Learning curves for the multi-goal tasks. Results are shown over 15 independent runs. The bold line shows the median, and the light area indicates the range between the 33rd to 67th percentile.}
	\label{fig:Filtered HER - performance}%
\end{figure}

As shown in Fig.\ref{fig:Filtered HER - Success rate} and \ref{fig:Filtered HER - Distance from goal}, the vanilla-HER algorithm fails to solve these tasks with nearly zero success rate and almost no improvements in the distance-to-goal measure. For both tasks, it is relatively hard to affect the achieved-goal in the first place. Without using \textit{Filtered-HER}, the agent observes too many misleading samples and fails to learn. Although \textit{Filtered-HER} improved the success rates in all tasks, the performances can be further increased by using the instructional-based selection strategy. Moreover, IBS leads to more robust performances, as indicated by the reduced range of the 33rd to 67th percentile.

\subsubsection{Estimated Q value}
\begin{figure}[ht]
	\centering
	\includegraphics[width=0.7\textwidth]{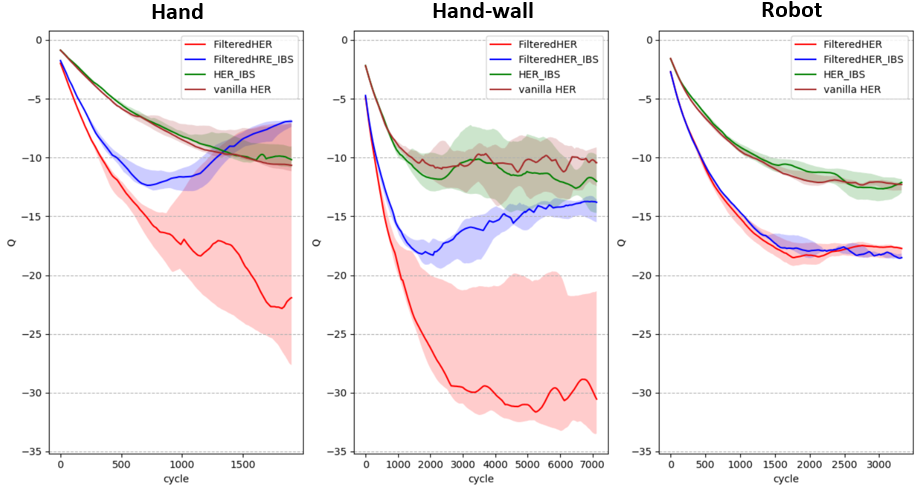}
	\caption[Filtered HER - Bias evaluation]{Bias evaluation. Results are shown over 15 independent runs. The bold line shows the median, and the light area indicates the range between the 33rd to 67th percentile.}
	\label{fig:Filtered HER - Estimated Q value}
\end{figure}

The misleading samples let the agent think it performs better than its real performances in practice. To evaluate the bias, we compare the agent's estimations of the initial state.
As shown in figure \ref{fig:Filtered HER - Estimated Q value}, the filter reduces the bias consistently and leads to a better and more realistic evaluation of the future return.
Figure \ref{fig:Filtered HER - Estimated Q value} presents the agents' evaluation for the Q-value of the initial state and action. This graph shows that when not using our filter, the agent gets too optimistic and is over-estimating its performances. Although both vanilla-HER and HER with IBS performs poorly compared to Filtered-HER and Filtered-HER with IBS respectively, almost in all cases, their estimations are higher than their parallels' evaluations.

\section{Conclusion}
In this chapter, we introduced a novel technique to reduce the bias induced by HER, called Filtered-HER. Filtered-HER detects experience that may induce bias and filters it out from the learning process. Using Filtered-HER, we have been able to train the agent on the full task (pick and throw), but we had to use a simplified version where the ball is initialized within the hand half of the time. This solution is not optimal since it requires the adaptation of the environments to the algorithm's needs. We published the algorithm and performances described in this chapter in the following article \cite{manela2019bias}. In the next chapter, we introduce a new algorithm that extends HER and can solve our real task with no modifications to the environment.


\chapter{Curriculum Learning with HER} 

\label{CurriculumHER} 


{\large\emph{learn to walk before you run}}.\\ \\
Just like humans, any ML-algorithm is facing difficulties when trying to learn a complex task from tabula rasa. Thus, it will make more sense to start with a simplified version of a task and gradually increase the task complexity. This approach is also known as \emph{Curriculum Learning}. Inspired by this idea, we built our algorithm \textit{Curriculum Hindsight Experience Replay} (\emph{CHER}). Our algorithm uses curriculum learning combined with Hindsight Experience Replay to solve complex tasks with a sparse reward function. We also present a novel technique, allowing the entire learning process to use the same simulation environment without the need to adjust it for each sub-task individually.\\
In this chapter we will first present the fundamentals for curriculum learning, followed by a description of the algorithm and the experiments' results. 

\section{Motivation}\label{s:cher motivation}
Learning from a sparse reward function is particularly difficult since the agent must solve the entire task, while following a random policy, before receiving any reward other than -1. The time complexity (expected time of solving the task), using traditional reinforcement learning algorithms, is exponential in the task complexity. This problem is magnified for multi-layered tasks (sequential manipulation tasks), such as our ball throwing tasks in which the agent needs first to reach the ball and only then throw it. HER enables solving single-layered tasks with sparse reward function in a reasonable time by generating an auto-curriculum that reduces the expected time significantly. Nevertheless, as shown in chapter \ref{FilteredHER}, HER is often useless for multi-layered tasks since it is hard to affect the achieved goal in the first place. In the paper \cite{andrychowicz2017hindsight}, the authors used HER for a double-layered task (\emph{pick and place}) by artificially simplifying the first layer in the game, so that the algorithm will be able to learn. Thus, in this approach, the environment is adapted to suit the algorithm and can only be used for environments with a maximum of two layers. In this chapter, we develop a more fundamental approach in terms of an algorithm, which can solve multi-layered tasks by applying HER on each layer sequentially, without requiring any adaptation to the environment. Our algorithm shows vast improvement for multi-layered tasks, compared to the original HER.


\section{Fundamentals}
In this section, we present the curriculum learning approach and its connection to transfer learning.

\subsection{Transfer learning}
\begin{figure}[th]%
	\centering
	\subfloat[supervised learning setup]{{\includegraphics[width=0.5\textwidth]{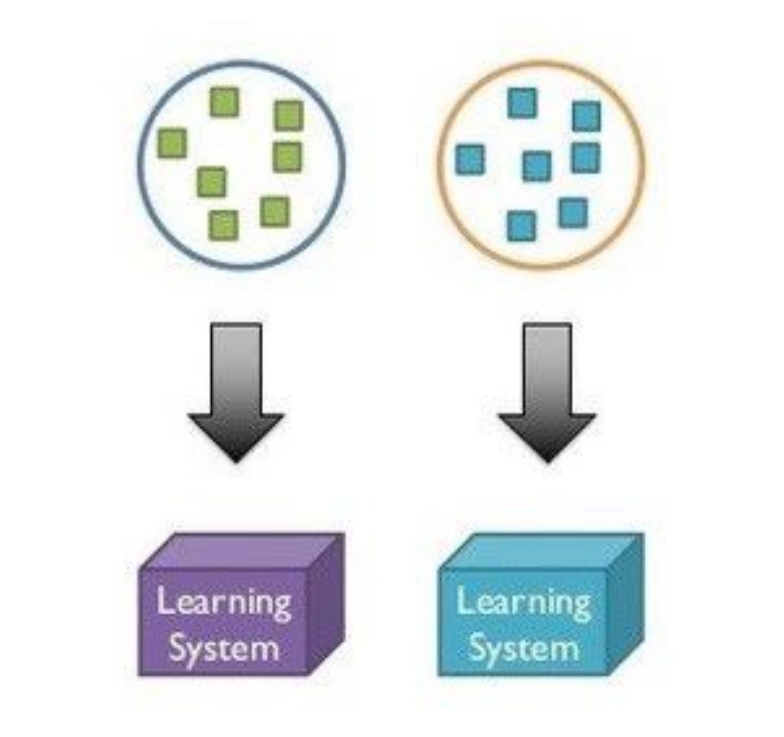} 
	\label{fig:supervised learning}}}
	\subfloat[transfer learning setup]{{\includegraphics[width=0.5\textwidth]{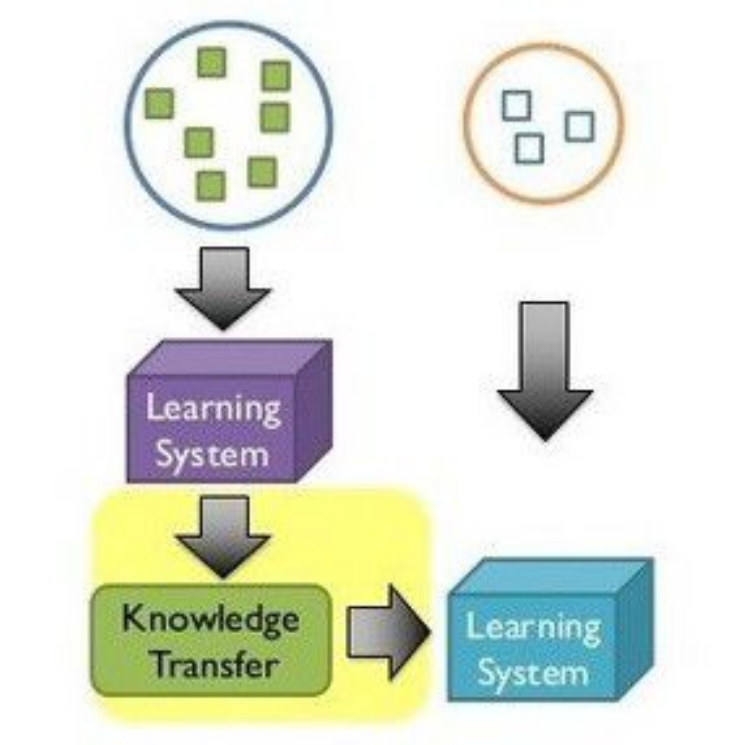}
	\label{fig:transfer learning}}}%
	
	\caption[Supervised learning vs transfer learning]{Supervised learning vs transfer learning\footnotemark.}
	\label{fig:Supervised learning vs transfer learning}%
\end{figure}

In the classic supervised machine learning framework, shown in figure \ref{fig:supervised learning}, a model for some domain A is trained by using a set of labeled data for the same domain. We can now train a model on this dataset A and expect it to generalize to unseen data from the same domain. When given data for some other domain B, we require again labeled data of the domain B, which can be used to train a new model that is expected to perform well on this type of data.\\
This model breaks down when there are not sufficient labeled data to train our model on. Transfer learning enables leveraging knowledge learned in one task, also known as \emph{source task} (figure \ref{fig:transfer learning}), to learn a new, related task.

\footnotetext{Source: \url{https://medium.com/data-science-101/transfer-learning-57ce3b98650}}
\subsection{Curriculum learning}
Curriculum learning \cite{bengio2009curriculum}, also known as \emph{sequential transfer learning}, is an approach where instead of training the model on the desired task from the beginning, we train the agent first an a source task, which is a simplified version of the task, and its complexity is gradually increasing. The goal in curriculum learning is to design a sequence of source tasks for an agent to train on, such that final performance or learning speed is improved. This approach is inspired by human learning during childhood, which follows a curriculum defined by parents and education systems by exposing children to simple concepts first and then gradually increasing them.


\section{Time Complexity of curriculum-HER}
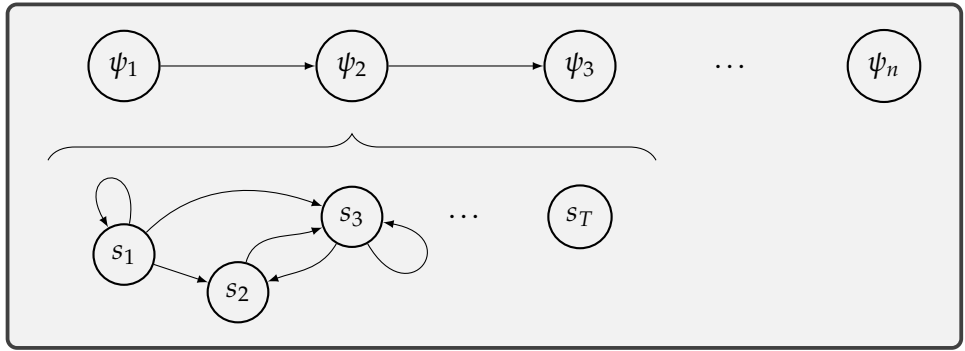
\begin{figure}
    \centering
    \tcbox{
    \begin{tikzpicture}[auto,node distance=8mm,>=latex,font=\small]
        \tikzstyle{round}=[thick,draw=black,circle]
        \node[round]  at (0, 0) (t1) {$\psi_1$};
        \node[round]  at (3, 0) (t2) {$\psi_2$};
        \node[round]  at (6, 0) (t3) {$\psi_3$};
        \node[round]  at (10, 0) (t4) {$\psi_n$};
        \node at ($(t3)!.5!(t4)$) {\ldots};

        \draw[->] (t1) -- (t2);
        \draw[->] (t2) -- (t3);
        
        \node[round]  at (0, -2.5) (s1) {$s_1$};
        \node[round]  at (1.5, -3) (s2) {$s_2$};
        \node[round]  at (3, -2) (s3) {$s_3$};
        \node[round]  at (6, -2) (sT) {$s_T$};
        \node at ($(s3)!.5!(sT)$) {\ldots};

        \draw[->] (s1) [out=80,in=120,loop] to (s1);
        \draw[->] (s1) -- (s2);
        \draw[->] (s1) to [out=45,in=160] (s3);
        \draw [->] (s2) to [out=75,in=200] (s3);
        \draw [->] (s3) to [out=240,in=20] (s2);
        \draw[->] (s3) [out=300,in=350,loop] to (s3);

        
        \draw [decorate,decoration={brace,amplitude=10pt},yshift=0.75cm]
        (-1, -2) -- (7, -2);
    \end{tikzpicture}
    }
    \caption[multi-layered task]{multi-layered task}
    \label{fig:multi-layered task}
\end{figure}

We define a sequential manipulation task as a multi-layered task. Let $\Psi$ be a multi-layered task, consisting of $\{\psi_1, ..., \psi_n\}$ sub-tasks. In order to complete the task $\Psi$, the agent needs to solve sequentially each sub-task, i.e,. $\psi_1 \rightarrow \psi_2 \rightarrow \dots \rightarrow  \psi_n$ (see figure \ref{fig:multi-layered task}). An example of a two-layered task is our hand-throwing task. First, the agent needs to grab the ball with the hand and then throw it towards the target. The agent cannot move the ball before grabbing it with the hand.\\
Let $\mathcal{O}(\psi_i)$ be the state-space complexity (from now will be denoted as \emph{complexity}) of sub task $\psi_i$ and $\mathcal{O}_t(\psi_i)$ the time complexity of sub task $\psi_i$. 
A task's complexity is the sum of all its sub-tasks' complexities:
\begin{equation}
\label{eq:task complexity}
    \mathcal{O}(\Psi) = \sum_i\mathcal{O}(\psi_i)\,.
\end{equation}
As shown in \cite{whitehead1991complexity,koenig1993complexity}, when following a random policy, the task's time complexity is an exponential function of its complexity:
\begin{equation}
\label{eq:task time complexity}
    \mathcal{O}_t(\Psi) = exp(\mathcal{O}(\Psi))\,.
\end{equation}
Using equations \ref{eq:task complexity} and \ref{eq:task time complexity}, the task's time complexity can be written as follows:
\begin{equation}
    \mathcal{O}_t(\Psi) = exp(\sum_i\mathcal{O}(\psi_i)) = \prod_i exp(\mathcal{O}(\psi_i)) = \prod \mathcal{O}_t(\psi_i)\,.
\end{equation}
When using sparse reward function without HER, the agent needs to solve the task, following a random policy. Hence, the time complexity of a task with sparse reward function is the product of all its sub-tasks' time complexities.\\
When using a curriculum, the task's time complexity reduces to about the sum of all sub-tasks' time complexities, since the agent solves only one sub-task at a time. By applying HER to each sub-task $\psi_i$, we reduced its time complexity (approximately) to a polynomial function of its complexity. 
\begin{equation}
    \mathcal{O}_t(\Psi) \approx \sum_i \mathcal{O}_t(\psi_i)) = \sum_i poly(\mathcal{O}(\psi_i)))\,.
\end{equation}
We refer to appendix \ref{AppendixC} for a proof.\\

\section{Method}
Our approach consists of two curricula types applied sequentially to each sub-task:
\begin{enumerate}
\item Use a layer-based HER to solve the current sub-task (\ref{ss:Layer-based HER}).
\item Transfer the knowledge learned in the current sub-task to the next sub-task (\ref{ss:Knowledge transfer between sub-tasks}).
\end{enumerate}

\subsection{Layer-based HER} \label{ss:Layer-based HER}
\begin{figure}[H]
	\centering
	\subfloat[Sub-task 1]
	{{\includegraphics[width=0.4\textwidth]{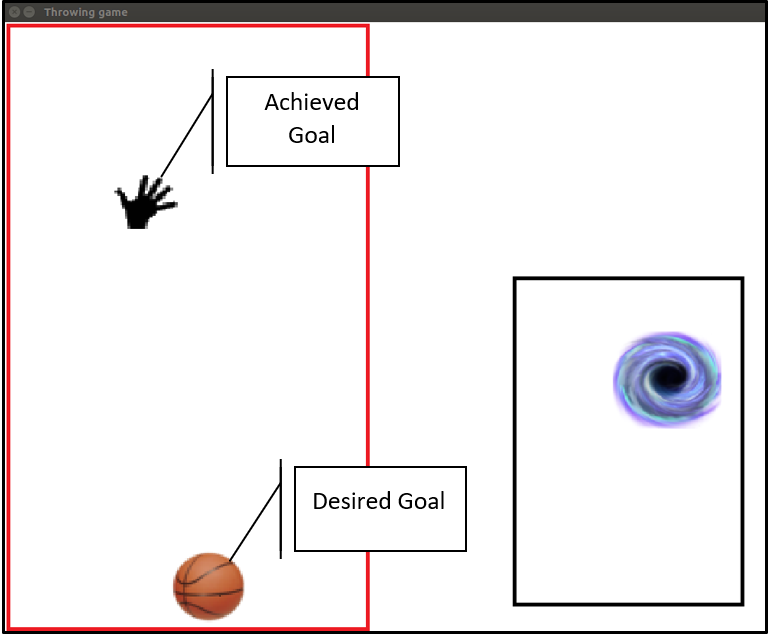} 
	\label{fig:cher - subtask1}}}
	\hskip 3ex
	\subfloat[Sub-task 2]
	{{\includegraphics[width=0.4\textwidth]{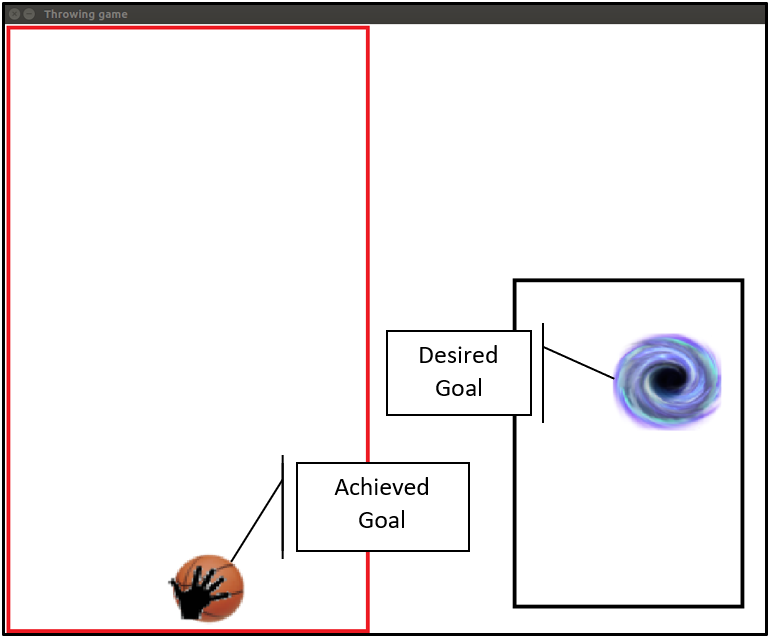} 
	\label{fig:cher - subtask2}}}%
	\caption[CHER - Sub Tasks]{An illustration of the sub-tasks in the task Hand. Figure \ref{fig:cher - subtask1} shows the first sub-task and figure \ref{fig:cher - subtask2}, the second. The objects' dimensions were enlarged for demonstration purposes}
	\label{fig:cher - subtasks}%
\end{figure}

Each sub-task consists of an observation dictionary of the same structure as used for the original HER:
\begin{itemize}
\item \textbf{observation} - the current state of the environment
\item \textbf{achieved goal} - the goal achieved in the current state
\item \textbf{desired goal} - the goal of the current episode
\end{itemize}
The achieved goal is the position of the current sub-task's object, and the desired goal is the object's desired position. Our approach assumes the desired goal for the object of sub-task $\psi_i$ is the location of the object from sub-task $\psi_{i+1}$. For each sub-task, we predefined the state space as well as the achieved and desired goals. Furthermore, the tolerance area of the reward function around the target can vary between sub-tasks, depending on the objects. For example, in take our hand-throwing task. In this task we need to reach the target with the ball, thus the desired goal is the target, and the reached goal is the ball's position. This task consists of two sub-tasks:
\paragraph{First sub-task:} The object is the hand, the achieved goal is the hand position, the desired goal is the ball position, and the reward function's threshold is the ball's radius (see figure \ref{fig:cher - subtask1}). By manipulating the hand, we can reach the target (the ball position).
\paragraph{Second sub-task:} Now, when the agent can get the hand to the ball, it can start manipulating the ball and reaching the target (the black-hole). Thus, the object is now the ball, the achieved goal is the ball position, the desired goal is the black-hole's position, and the reward function's threshold is the black-hole's radius (see figure \ref{fig:cher - subtask1}).

\subsection{Knowledge transfer between sub-tasks} \label{ss:Knowledge transfer between sub-tasks}
In order to leverage the knowledge learned in the previous sub-task, we use the learned policy as the initial policy for the current sub-task. The problem is that the input dimension, that is, the state space, is different for each sub-task. In order to adjust the input dimensions of the network without affecting the output, we add the new dimensions with all weights equal to 0.
With weights equal to zero, the new dimensions do not affect the policy, and keep the changes as smooth as possible.
In this way the agent can easily manipulate the current sub-task's object, and generate a useful experience, from which the new task can be learned from. CHER must use Filtered-HER (see chapter \ref{FilteredHER}). Otherwise, when facing the new sub-task, the agent will "forget" what it learned due to all the misleading samples. In section \ref{s:cher experiments} we show the performances of CHER with- and without Filtered-HER.

\subsection{Implementation}\label{s:cher implementation}
In order to train all the sub-tasks on the same simulation, we define the procedure \textit{state\_to\_obs}. This procedure gets the current state of the environment (the current observation concatenated with the goal) and returns the relevant observation dictionary, which is a reduced version of the real task. 
Before starting the learning process, we call the function \textit{get\_curriculum} (see algorithm \ref{alg:get_curriculum}) that returns a predefined list of \textit{state\_to\_obs} procedures for all the sub-task. While training, we call the current \textit{state\_to\_obs} procedure at each step, and the agent plays according to the returned observation. See algorithm \ref{alg:state_to_obs} for the formal pseudo-code.

\begin{algorithm}[H]
	\caption{Build the layer-based obs dictionary}
    \label{alg:state_to_obs}
    \begin{algorithmic}[1]
		\Require
            \Statex $obs\_idx$ - indices for the observation
            \Statex $achieved\_idx$ - indices for the achieved goal
            \Statex $desired\_idx$ - indices for the desired goal
        \Input $S$ (current state) - The real observation concatenated with the real goal
        \Output $modified\_obs$ - The modified $obs$ dictionary
	    \Procedure{state\_to\_obs}{$S$}
            \State $modified\_obs[observation] = S[obs\_idx]$
            \State $modified\_obs[achieved\_goal] = S[achieved\_idx]$
            \State $modified\_obs[desired\_goal] = S[desired\_idx]$
        \EndProcedure
	\end{algorithmic}
\end{algorithm}

\begin{figure}
	\centering	
	\tikzset{%
		neuron missing/.style={
			draw=none, 
			scale=4,
			text height=0.2cm,
			execute at begin node=\color{black}$\vdots$
		},
	}
	
	\begin{tikzpicture}[x=1.5cm, y=1.5cm, >=stealth]
	
	\node[text width=1cm] at (0,2.5){Input};
	\foreach \m/\l [count=\y] in {1,...,4}
	{
		\node [circle,fill=brown,minimum size=0.4cm] (input-\m) at (0,2.5-0.5*\y) {};
	}

	\foreach \m/\l [count=\y] in {5,...,6}
	{
		\node [circle,fill=red,minimum size=0.4cm] (input-\m) at (0,2.5-0.5*\m) {};
	}
	
	\foreach \m/\l [count=\y] in {7,...,8}
	{
		\node [circle,fill=blue,minimum size=0.4cm] (input-\m) at (0,2.5-0.5*\m) {};
	}
		
	\node[text width=1cm] at (2,2.5){Hidden};
	\foreach \m [count=\y] in {1,...,6}
	\node [circle,fill=black!50,minimum size=0.4cm ] (hidden-\m) at (2,2-0.5*\y) {};

	\node[text width=1cm] at (4,2.5){Output};
	\foreach \m [count=\y] in {1}
	\node [circle,fill=black!50,minimum size=0.4cm ] (output-\m) at (4,1.0) {};
	
	\foreach \m [count=\y] in {2}
	\node [circle,fill=black!50,minimum size=0.4cm ] (output-\m) at (4,-0.5) {};
	
	\node [neuron missing]  at (4,0.1) {};
	
	\foreach \l [count=\i] in {1,...,8}
	\foreach \l [count=\i] in {1,...,6}
	\node [above] at (hidden-\i.north){};
	\foreach \l [count=\i] in {1,n}
	\node [above] at (output-\i.east){};
	
	\foreach \i in {1,...,4}
	\foreach \j in {1,...,6}
	\draw [->] (input-\i) -- (hidden-\j);
	
	\foreach \i in {1,...,6}
	\foreach \j in {1,...,2}
	\draw [->] (hidden-\i) -- (output-\j);
	\foreach \i in {5,...,8}
	\foreach \j in {1,...,6}
	\draw [dashed,->] (input-\i) -- (hidden-\j);
	
	\draw[decoration={brace,mirror,raise=5pt},decorate]
  (-0.8,2.2) -- node[left=6pt] {$\Psi$} (-0.8,-1.7);
	\draw[decoration={brace,mirror,raise=5pt},decorate]
  (-0.2,2.2) -- node[left=6pt] {$\psi_1$} (-0.2,0.3);
  \draw[decoration={brace,mirror,raise=5pt},decorate]
  (-0.2,0.2) -- node[left=6pt] {$\psi_2$} (-0.2,-0.7);
  \draw[decoration={brace,mirror,raise=5pt},decorate]
  (-0.2,-0.8) -- node[left=6pt] {$\psi_3$} (-0.2,-1.7);
	\end{tikzpicture}
		
    \caption[Neural network for curriculum]{Neural network for curriculum. \textcolor{brown}{Brown} neurons are for the first sub-task ($\psi_1$), \textcolor{red}{red} are for the second sub-task ($\psi_2$), and \textcolor{blue}{blue} are for the third sub-task ($\psi_3$). The dashed weights are initialized with zeros.}
    \label{fig:neural network for curriculum}
\end{figure}
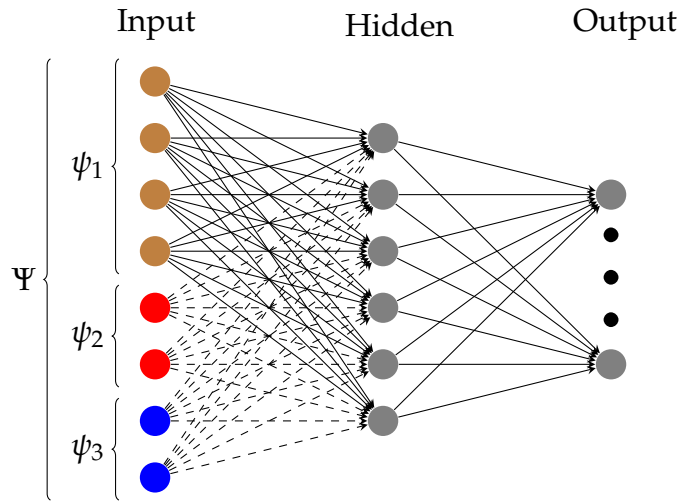

The agent trains on each sub-task until it reaches a predefined level of expertise on this task. The agent's expertise is measured by a moving average over the agent's success rate. When the agent reaches an average of $90\%$ success-rate with a window size of $k$, the algorithm automatically switches to the next sub-task (see algorithm \ref{alg:learned_task}). We investigated task-overfitting, and policy's re-maneuverability for different window sizes $k$ in section \ref{ss:cher experiments task overfitting}.\\
As explained in section \ref{ss:Knowledge transfer between sub-tasks}, when switching between sub-task, the weights of the new dimensions must be initialized with zeros. In order to improve the algorithm implementation and efficiency, we initialized the agent's networks in their final architecture and initialized all the weights that are unused for the first sub-task to zero (see figure \ref{fig:neural network for curriculum}). In order to match the experience- and networks' dimensions, we pad the experience with zeros. By doing so, the local gradients of the un-used weights are always zeros, and the weights do not change. We compared different networks initialization methods in section \ref{ss:cher experiments Networks Initialization Methods}. In addition, the experience buffer is emptied for each sub-task. See algorithm \ref{alg:CHER} for the formal pseudo-code.

\begin{algorithm}
	\caption{get curriculum}
	\label{alg:get_curriculum}
	\begin{algorithmic}[1]
		\Require
            \Statex $state\_to\_obs$ function for each layer
            \Statex padding vector for each layer
            \Statex reward function threshold for each layer
        \Output $curriculum$ - A list containing all the information for each layer
	    \Procedure{get\_curriculum}{}
	        \State curriculum $\leftarrow$ []
	        \For{each layer}
	            \State l[sto] $\leftarrow$ $state\_to\_obs$
	            \State l[rft] $\leftarrow$ $reward\_function\_threshold$
	            \State l[pad] $\leftarrow$ padding
                \State curriculum += l \Comment{+= denotes for append}
            \EndFor
        \EndProcedure
	\end{algorithmic}
\end{algorithm}

\begin{algorithm}
	\caption{check if the agent learned the task}
	\label{alg:learned_task}
	\begin{algorithmic}[1]
		\Require
            \Statex c - success-rate threshold \Comment{we used c = 90\%}
            \Statex k - window size
        \Input $B$ - The success rate buffer
        \Output $learned$ - A Boolean. True if the performances reached the threshold
	    \Procedure{learned\_task}{$R$}
	        \State results $\leftarrow$ average($R$[-k]) \Comment{last k samples}
	        \State $learned \leftarrow results \ge c$
        \EndProcedure
	\end{algorithmic}
\end{algorithm}

\begin{algorithm}
	\caption{Curriculum HER (CHER)}
	\label{alg:CHER}
	\begin{algorithmic}[1]
		\Require
		\Statex \textbullet~ an off-policy RL algorithm $\mathbb{A}$,\Comment e.g. DQN, DDPG 
		
		\State Initialize $\mathbb{A}$
		\State curriculum $\leftarrow$ get\_curriculum() \Comment{See algorithm \ref{alg:get_curriculum}}
		\State Initialize unused weights to zero

		\For{each $layer$ in curriculum}
    		\State pad, sto, rft $\leftarrow$ $layer[pad],\, layer[sto],\, layer[rft]$ \footnotemark
    		\State initialize replay buffer $R$
    		\State initialize success rate buffer $B$
    		\State  $learned$ $\leftarrow$ False
    		\While{$learned$ is false}
        		\For{$Episode \leftarrow 1, M$}
            		\State Sample a goal $g$ and an initial state $s_0$.
            		\State obs $\leftarrow$ sto($s_0||g$) \Comment || denonts concatenation
            		\For{$t \leftarrow 0, T-1$}
                		\State Sample an action $a_t$ using the behavioral policy from $\mathbb{A}$: \par
                		$\qquad \qquad a_t \leftarrow \pi(\hat{s_t}||\hat{g}||pad)$ \Comment hat stands for modified state / goal
                		\State Execute the action $a_t$ and observe a new state $s_{t+1}$
            		    \State obs $\leftarrow$ sto($s_{t+1}||g$)
            		\EndFor
        		\EndFor
        		\State Preform filtered hindsight experience replay \Comment{see chapter \ref{FilteredHER}}
        		\State  $learned$ $\leftarrow$ learned\_task($B$)  \Comment{see algorithm \ref{alg:learned_task}}
    		\EndWhile
		\EndFor
	\end{algorithmic}
\end{algorithm}
\footnotetext{sto and rft stands for state\_to\_obs and reward\_function\_threshold}


\section{Related Work}
Applying curricula to reinforcement learning is not a new idea, but has received much attention recently due to its intuitive and elegant solution for complex tasks. Curriculum approaches for reinforcement learning can be clustered into two main classes:\\
\begin{itemize}
\item \textbf{Simplify the current task} - This class of algorithms modifies the environment in a way that the source tasks is a simpler version of the target task, but stays fundamentally the same, for example, by choosing target states which are more straightforward to achieve \cite{saito2018curriculum, held2018automatic, andrychowicz2017hindsight}, or by generating initial states closer to the goal \cite{florensa2017reverse}. The algorithms from this class use, most of the time, auto-curriculum approaches, which is a curriculum generated automatically, without the need to manually specify it by hand.

\item \textbf{Learn sub-skill} - This class of algorithms trains the agent on a fundamentally different source task, which provides the agent with essential knowledge to learn the real task \cite{narvekar2019learning, narvekar2017autonomous}. For example, learn how to play Ms. Pac-Man by first playing with no ghosts, and gradually introduce the agent with the different types of ghosts (as introduced by \cite{narvekar2019learning}). Most curriculum approaches from this class rely on the ability to provide the agent with adjusted simulations to train the source tasks on.
\end{itemize}
Our approach combines the two classes by splitting the full task into sequential sub-tasks and simplify each sub-task using HER. Unlike existing algorithms from the second class, our method does not require an adjusted simulation for each sub-task.


\section{Experiments} \label{s:cher experiments}
In this section, we evaluate Curriculum HER.\\
We tested our algorithms on the following environments:
\begin{enumerate}
\item \textbf{Hand\_v1} - The full version of the Hand task, where the ball is always initialized on the floor
\item \textbf{Hand\_wall\_v1} - The full version of the Hand-Wall task, where the ball is always initialized on the floor
\item \textbf{Robot\_v1} - The full version of the Hand task, where the ball is always initialized in the air.
\end{enumerate}
\noindent Training is performed using the DDPG algorithm \cite{lillicrap2015continuous}, in which the actor and the critic were represented using multi-layer perceptrons (MLPs). See Appendix \ref{AppendixB} for more details regarding networks architecture and hyperparameters. 
\subsection{Performances}
In order to test the performance of the algorithms, we ran on each environment the following combinations: \textbf{vanilla-HER}, \textbf{Filtered-HER} with \textbf{IBS}, \textbf{Unfiltered-CHER}\footnotemark, \textbf{Unfiltered-CHER}\footnotemark[\value{footnote}] with \textbf{IBS}, \textbf{CHER}, \textbf{CHER} with \textbf{IBS}. For all tasks, we set $\sigma = 0.2$ (equation \ref{eq:value}), when IBS is used. In all algorithms we used prioritized experience replay (PER) \cite{schaul2015prioritized}.
\footnotetext{As explained is section \ref{ss:Knowledge transfer between sub-tasks}\,, CHER uses Filter by default.\label{refnote}}
The results of the algorithms are evaluated using four criteria:
\begin{itemize}
	\item Success rate
	\item Distance-to-goal
	\item Positive Rewards
\end{itemize}
The first and second criteria evaluate the performances of the agent.
The third criterion shows the collection pace of useful data for each algorithm.

\subsubsection{Success Rate and Distance from Goal}
\begin{figure}[H]
	\centering
	\subfloat[Success rate]
	{{\includegraphics[width=0.7\textwidth]{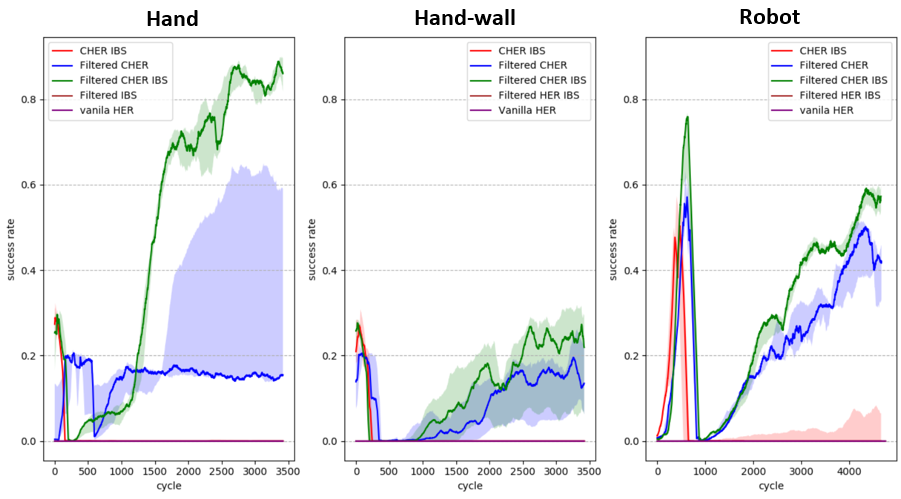} 
	\label{fig:cher - Success rate}}} \\
	\subfloat[Distance from goal]
	{{\includegraphics[width=0.7\textwidth]{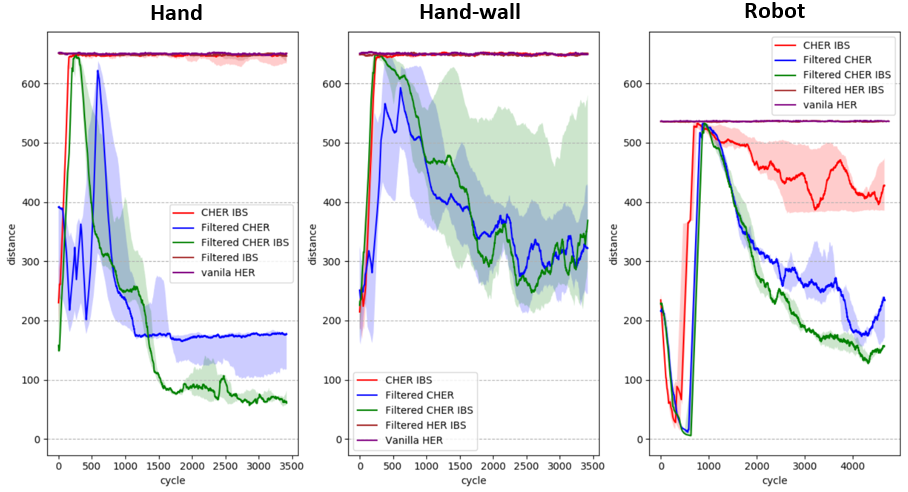} 
	\label{fig:cher - Distance from goal}}}%
	\caption[CHER - Performance]{Learning curves for the multi-goal tasks. Results are shown over 15 independent runs. The bold line shows the median, and the light area indicates the range between the 33rd to 67th percentile.}
	\label{fig:cher - performance}%
\end{figure}

As shown in figure \ref{fig:cher - Success rate} and \ref{fig:cher - Distance from goal}, the vanilla-HER algorithm fails to solve these tasks with zero success rate and no improvements in the distance-to-goal measure. For both tasks, it is relatively hard to affect the achieved-goal in the first place. Without using \textit{Filtered-HER}, the agent observes too many misleading samples and fails to learn. Although \textit{Filtered-HER} improved the success rates in both tasks, the performances can be further increased by using the instructional-based selection strategy. Moreover, IBS leads to more robust performances, as indicated by the reduced range of the 33rd to 67th percentile.

\subsubsection{Positive Rewards}
\begin{figure}[H]
	\centering
	\includegraphics[width=0.7\textwidth]{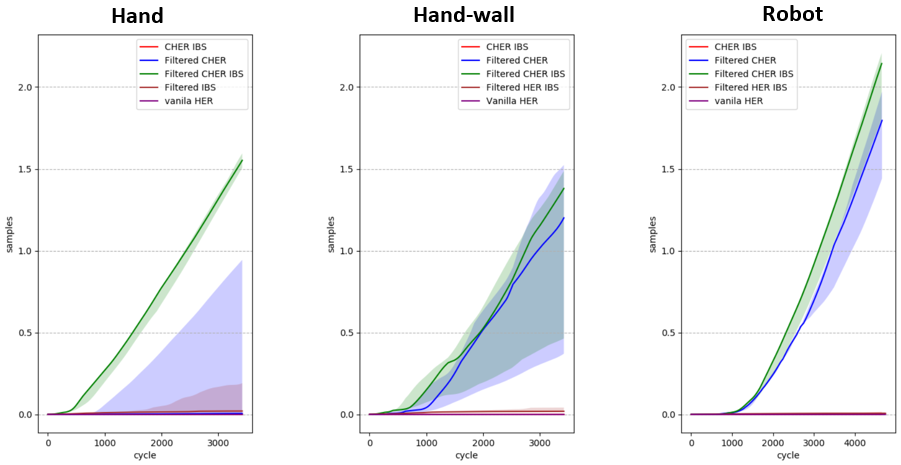}
	\caption[CHER - Positive rewards]{Positive rewards over time. The positive rewards are counted in tens of thousands. Results are shown over 15 independent runs. The bold line shows the median, and the light area indicates the range between the 33rd to 67th percentile.}
	\label{fig:cher - Positive Rewards}
\end{figure}

As explained in section \ref{s:cher motivation}, the problem when trying to apply HER on sequential tasks comes from the fact that the agent does not affect the achieved-goal, and thus don't get any useful experience. To learn, the agent must see a sufficient amount of positive rewards. As shown by figure  \ref{fig:cher - Positive Rewards}, In all tasks the curriculum approaches collect positive reward in a significantly faster rate. Notice that the samples are counter in tens of thousands

\subsection{Task Overfitting} \label{ss:cher experiments task overfitting}
\begin{figure}[H]
	\centering
	\includegraphics[width=0.7\textwidth]{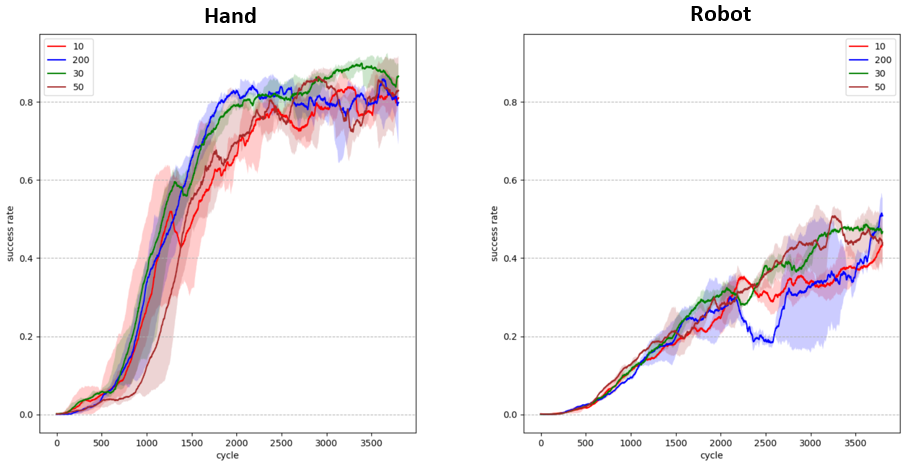}
	\caption[CHER - Task Overfitting]{Policy’s re-maneuverability for different amount of sub-task expertise. Results are shown over 10 independent runs. The bold line shows the median, and the light area indicates the range between the 33rd to 67th percentile.}
	\label{fig:cher - Task Overfitting}
\end{figure}

As explained in section \ref{s:cher implementation}, CHER trains the agent on each sub-task to some expertise and then move to the next sub-task. The agent's knowledge is measured by a moving average on the success-rate with a window size of k. Using different window sizes may affect policy's re-maneuverability due to task over- or under-fitting.\\
In this section, we investigated task-overfitting and policy's re-maneuverability for different window sizes.
To isolate the policy's adaptation time, we trained the agent on the sub-tasks until it reached the desired expertise and saved the trained policy. Then we loaded the trained policy and trained the agent on the real task (with no curriculum). As shown in figure \ref{fig:cher - Task Overfitting}, the algorithm is relatively robust to different window sizes, and the difference in success-rate is negligible. Nevertheless, we use a window size of 20, since it consistently had a slight advantage, and it is faster to achieve compared to larger window sizes.

\subsection{Networks Initialization Methods} \label{ss:cher experiments Networks Initialization Methods}
As explained in section \ref{ss:Knowledge transfer between sub-tasks}, to transfer the knowledge between sub-tasks, we initialize the new dimensions of the actor-network with all weights equal to zeros. By initializing the new dimensions with zeros, we maintain the policy's updates as smooth as possible. In this section, we investigate the effect of different initialization methods for the critic network on the Hand task. We first initialize the new weights regularly (same initialization as the rest of the weights) and then multiply it by a factor $\alpha$. We compared three different methods:
\begin{enumerate}
\item \textbf{Regular}: $\alpha = 1$
\item \textbf{Decreased weights}: $\alpha = 0.1$
\item \textbf{Reset weights}: $\alpha = 0$
\end{enumerate}

\begin{figure}[H]
	\centering
	\includegraphics[width=0.7\textwidth]{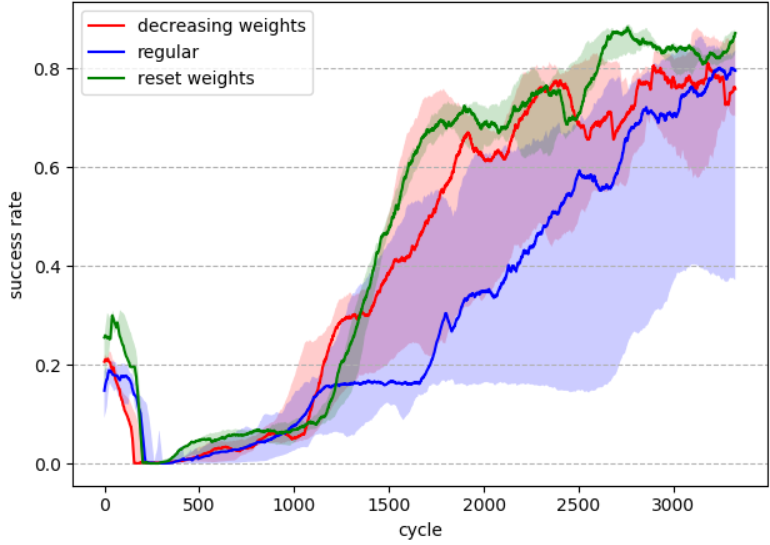}
	\caption[CHER - Critic initialization methods]{Success rate for different initialization methods. Results are shown over 10 independent runs. The bold line shows the median, and the light area indicates the range between the 33rd to 67th percentile.}
	\label{fig:cher - Networks Initialization Methods}
\end{figure}
As shown in figure \ref{fig:cher - Networks Initialization Methods}, we got best results with the \textbf{Reset} method. It means we don't change the critic's function when adding more dimensions.


\section{Conclusion}
In this chapter, we introduced a novel algorithm to enable HER on sequential manipulation tasks, called CHER. CHER applies HER on each sub-task sequentially using a curriculum-learning approach. Using CHER, we have been able to train the agent on the full task (pick and throw) with no simplification and with no need for modified environments to learn from.

\chapter{Conclusion and Future Work} 

\label{conclusion} 

In this work, we aimed to solve challenging manipulation tasks with sparse feedback using deep reinforcement learning. 
We restricted our solution to sparse feedback to enable an easy adaptation of our algorithms to new tasks. We applied our algorithms to throwing tasks and compared their performances to 
an existing state-of-art algorithm (Hindsight Experience Replay).
Solving throwing tasks requires many skills such as understanding fundamental physics (e.g., gravity), learning sequential controls, and generalize over different task conditions, while receiving insufficient feedback from a sparse reward function.
Many of the mentioned requirements are currently not provided by existing algorithms.
We based our work on the algorithm \emph{Hindsight Experience Replay} (HER).\\
\indent In the first stage, we built a varied set of 2D simulated throwing tasks with different levels of complexities to test our algorithms. We built two classes of tasks: hand manipulation tasks, where the agent controls the Cartesian velocity of the hand and robot manipulation tasks where the agent controls the angular velocities of the manipulator's joints.\\
\indent In the second stage, we presented an \emph{Instructional Based Strategy} (IBS) to improve the virtual-goal generating process in HER by exploiting the generalization capabilities of neural-networks. This led to an improved performance in simple and more challenging manipulation tasks.\\
\indent In the third stage, we augmented the vanilla-HER algorithm with a filter that reduces the bias induced in the learning process by HER. This \textit{Filtered-HER} algorithm improved the performances significantly in all of our tasks. 
\indent Finally, we developed a new extension of HER, called Curriculum-HER (CHER), that can solve sequential control tasks. CHER learns complicated tasks using a curriculum-learning approach that applies HER on each sub-task sequentially. \\
Using our three algorithms, we have been able to solve the robotic throwing task and reach a success rate of 80\% on the full hand manipulation task and 40\% on the full joints manipulation task. In conclusion, we managed to outperform existing algorithms, thus providing a proof of concept and opened new research directions for learning from sparse feedback.
\section{Future Work}
The research presented in this thesis has several possible directions of extensions for future work. 
\subsection{Random Network Distillation for virtual goals prioritization:}
In our IBS algorithm, we measure the inventiveness of a virtual-goal using the difference between the proposed- and target distributions of the virtual-goals. This measure may be improved using \textit{Random network distillation} (RND).
RND is a method to measure the familiarity of the agent with each input \cite{burda2018exploration}. This method is training a network to predict the output of a second, static network for a given input. The error of the learning network measures the familiarity of the agent with this input and other inputs nearby.
Replacing the current measure with RND may lead to better prioritizing virtual goals with which the agent is not familiar.\\

\subsection{Automatic CHER:}
Another direction of research may be to automate the curriculum in CHER. This may be done using a teacher-student curriculum \cite{matiisen2019teacher}, where the hard-coded $Get\_Curriculum()$ procedure is replaced by an agent (the teacher) that specifies the sub-task in each episode by choosing the suitable indices of the achieved- and desired goal.



\bibliographystyle{unsrt}
\bibliography{references}

\begin{thebibliography}{10}

\bibitem{mayor2018gods}
Adrienne Mayor.
\newblock {\em Gods and Robots: Myths, Machines, and Ancient Dreams of
  Technology}.
\newblock Princeton University Press, 2018.

\bibitem{fujiwara1995floor}
Yoshimori Fujiwara, Kazuhiro Hiratsuka, Yoshiya Yamaue, Hiroaki Arakawa, Daizo
  Takaoka, and Ryuji Suzuki.
\newblock Floor cleaning robot and method of controlling same, March~28 1995.
\newblock US Patent 5,402,051.

\bibitem{chang2011floor}
Wen-Chung Chang.
\newblock Floor washing robot, August~23 2011.
\newblock US Patent 8,001,651.

\bibitem{mortimer2003mix}
John Mortimer.
\newblock Mix of robots used for jaguar's aluminium-bodied xj luxury car.
\newblock {\em Industrial Robot: An International Journal}, 30(2):145--151,
  2003.

\bibitem{kitano1998robocup}
Hiroaki Kitano.
\newblock {\em RoboCup-97: robot soccer world cup I}, volume 1395.
\newblock Springer Science \& Business Media, 1998.

\bibitem{muller2007making}
Heiko M{\"u}ller, Martin Lauer, Roland Hafner, Sascha Lange, Artur Merke, and
  Martin Riedmiller.
\newblock Making a robot learn to play soccer using reward and punishment.
\newblock In {\em Annual Conference on Artificial Intelligence}, pages
  220--234. Springer, 2007.

\bibitem{zhang2018behavioral}
Xin Zhang, Maolin Chen, and Xingqun Zhan.
\newblock Behavioral cloning for driverless cars using transfer learning.
\newblock In {\em 2018 IEEE/ION Position, Location and Navigation Symposium
  (PLANS)}, pages 1069--1073. IEEE, 2018.

\bibitem{abbeel2010autonomous}
Pieter Abbeel, Adam Coates, and Andrew~Y Ng.
\newblock Autonomous helicopter aerobatics through apprenticeship learning.
\newblock {\em The International Journal of Robotics Research},
  29(13):1608--1639, 2010.

\bibitem{kober2009learning}
Jens Kober and Jan Peters.
\newblock Learning motor primitives for robotics.
\newblock In {\em 2009 IEEE International Conference on Robotics and
  Automation}, pages 2112--2118. IEEE, 2009.

\bibitem{sutton2018reinforcement}
Richard~S Sutton and Andrew~G Barto.
\newblock {\em Reinforcement learning: An introduction}.
\newblock MIT press, 2018.

\bibitem{goodfellow2016deep}
Ian Goodfellow, Yoshua Bengio, and Aaron Courville.
\newblock {\em Deep learning}.
\newblock MIT press, 2016.

\bibitem{mnih2013playing}
Volodymyr Mnih, Koray Kavukcuoglu, David Silver, Alex Graves, Ioannis
  Antonoglou, Daan Wierstra, and Martin Riedmiller.
\newblock Playing atari with deep reinforcement learning.
\newblock {\em arXiv preprint arXiv:1312.5602}, 2013.

\bibitem{todorov2012mujoco}
Emanuel Todorov, Tom Erez, and Yuval Tassa.
\newblock Mujoco: A physics engine for model-based control.
\newblock In {\em 2012 IEEE/RSJ International Conference on Intelligent Robots
  and Systems}, pages 5026--5033. IEEE, 2012.

\bibitem{lillicrap2015continuous}
Timothy~P Lillicrap, Jonathan~J Hunt, Alexander Pritzel, Nicolas Heess, Tom
  Erez, Yuval Tassa, David Silver, and Daan Wierstra.
\newblock Continuous control with deep reinforcement learning.
\newblock {\em arXiv preprint arXiv:1509.02971}, 2015.

\bibitem{schulman2017proximal}
John Schulman, Filip Wolski, Prafulla Dhariwal, Alec Radford, and Oleg Klimov.
\newblock Proximal policy optimization algorithms.
\newblock {\em arXiv preprint arXiv:1707.06347}, 2017.

\bibitem{silver2018general}
David Silver, Thomas Hubert, Julian Schrittwieser, Ioannis Antonoglou, Matthew
  Lai, Arthur Guez, Marc Lanctot, Laurent Sifre, Dharshan Kumaran, Thore
  Graepel, et~al.
\newblock A general reinforcement learning algorithm that masters chess, shogi,
  and go through self-play.
\newblock {\em Science}, 362(6419):1140--1144, 2018.

\bibitem{kober2013reinforcement}
Jens Kober, J~Andrew Bagnell, and Jan Peters.
\newblock Reinforcement learning in robotics: A survey.
\newblock {\em The International Journal of Robotics Research},
  32(11):1238--1274, 2013.

\bibitem{andrychowicz2017hindsight}
Marcin Andrychowicz, Filip Wolski, Alex Ray, Jonas Schneider, Rachel Fong,
  Peter Welinder, Bob McGrew, Josh Tobin, OpenAI~Pieter Abbeel, and Wojciech
  Zaremba.
\newblock Hindsight experience replay.
\newblock In {\em Advances in Neural Information Processing Systems}, pages
  5048--5058, 2017.

\bibitem{bellman1957markovian}
Richard Bellman.
\newblock A markovian decision process.
\newblock {\em Journal of mathematics and mechanics}, pages 679--684, 1957.

\bibitem{watkins1992q}
Christopher~JCH Watkins and Peter Dayan.
\newblock Q-learning.
\newblock {\em Machine learning}, 8(3-4):279--292, 1992.

\bibitem{rumelhart1988learning}
David~E Rumelhart, Geoffrey~E Hinton, Ronald~J Williams, et~al.
\newblock Learning representations by back-propagating errors.
\newblock {\em Cognitive modeling}, 5(3):1, 1988.

\bibitem{mnih2015human}
Volodymyr Mnih, Koray Kavukcuoglu, David Silver, Andrei~A Rusu, Joel Veness,
  Marc~G Bellemare, Alex Graves, Martin Riedmiller, Andreas~K Fidjeland, Georg
  Ostrovski, et~al.
\newblock Human-level control through deep reinforcement learning.
\newblock {\em Nature}, 518(7540):529, 2015.

\bibitem{uhlenbeck1930theory}
George~E Uhlenbeck and Leonard~S Ornstein.
\newblock On the theory of the brownian motion.
\newblock {\em Physical review}, 36(5):823, 1930.

\bibitem{schaul2015prioritized}
Tom Schaul, John Quan, Ioannis Antonoglou, and David Silver.
\newblock Prioritized experience replay.
\newblock {\em arXiv preprint arXiv:1511.05952}, 2015.

\bibitem{salimans2018learning}
Tim Salimans and Richard Chen.
\newblock Learning montezuma's revenge from a single demonstration.
\newblock {\em arXiv preprint arXiv:1812.03381}, 2018.

\bibitem{schaul2015universal}
Tom Schaul, Daniel Horgan, Karol Gregor, and David Silver.
\newblock Universal value function approximators.
\newblock In {\em International Conference on Machine Learning}, pages
  1312--1320, 2015.

\bibitem{zhang2016understanding}
Chiyuan Zhang, Samy Bengio, Moritz Hardt, Benjamin Recht, and Oriol Vinyals.
\newblock Understanding deep learning requires rethinking generalization.
\newblock {\em arXiv preprint arXiv:1611.03530}, 2016.

\bibitem{zhao2018energy}
Rui Zhao and Volker Tresp.
\newblock Energy-based hindsight experience prioritization.
\newblock {\em arXiv preprint arXiv:1810.01363}, 2018.

\bibitem{manela2019bias}
Binyamin Manela and Armin Biess.
\newblock Bias-reduced hindsight experience replay with virtual goal
  prioritization.
\newblock {\em arXiv preprint arXiv:1905.05498}, 2019.

\bibitem{plappert2018multi}
Matthias Plappert, Marcin Andrychowicz, Alex Ray, Bob McGrew, Bowen Baker,
  Glenn Powell, Jonas Schneider, Josh Tobin, Maciek Chociej, Peter Welinder,
  et~al.
\newblock Multi-goal reinforcement learning: Challenging robotics environments
  and request for research.
\newblock {\em arXiv preprint arXiv:1802.09464}, 2018.

\bibitem{bengio2009curriculum}
Yoshua Bengio, J{\'e}r{\^o}me Louradour, Ronan Collobert, and Jason Weston.
\newblock Curriculum learning.
\newblock In {\em Proceedings of the 26th annual international conference on
  machine learning}, pages 41--48. ACM, 2009.

\bibitem{whitehead1991complexity}
Steven~D Whitehead.
\newblock A complexity analysis of cooperative mechanisms in reinforcement
  learning.
\newblock In {\em AAAI}, pages 607--613, 1991.

\bibitem{koenig1993complexity}
Sven Koenig and Reid~G Simmons.
\newblock Complexity analysis of real-time reinforcement learning.
\newblock In {\em AAAI}, pages 99--107, 1993.

\bibitem{saito2018curriculum}
Atsushi Saito.
\newblock Curriculum learning based on reward sparseness for deep reinforcement
  learning of task completion dialogue management.
\newblock In {\em Proceedings of the 2018 EMNLP Workshop SCAI: The 2nd
  International Workshop on Search-Oriented Conversational AI}, pages 46--51,
  2018.

\bibitem{held2018automatic}
David Held, Xinyang Geng, Carlos Florensa, and Pieter Abbeel.
\newblock Automatic goal generation for reinforcement learning agents.
\newblock 2018.

\bibitem{florensa2017reverse}
Carlos Florensa, David Held, Markus Wulfmeier, Michael Zhang, and Pieter
  Abbeel.
\newblock Reverse curriculum generation for reinforcement learning.
\newblock {\em arXiv preprint arXiv:1707.05300}, 2017.

\bibitem{narvekar2019learning}
Sanmit Narvekar and Peter Stone.
\newblock Learning curriculum policies for reinforcement learning.
\newblock In {\em Proceedings of the 18th International Conference on
  Autonomous Agents and MultiAgent Systems}, pages 25--33. International
  Foundation for Autonomous Agents and Multiagent Systems, 2019.

\bibitem{narvekar2017autonomous}
Sanmit Narvekar, Jivko Sinapov, and Peter Stone.
\newblock Autonomous task sequencing for customized curriculum design in
  reinforcement learning.
\newblock In {\em IJCAI}, pages 2536--2542, 2017.

\bibitem{burda2018exploration}
Yuri Burda, Harrison Edwards, Amos Storkey, and Oleg Klimov.
\newblock Exploration by random network distillation.
\newblock {\em arXiv preprint arXiv:1810.12894}, 2018.

\bibitem{matiisen2019teacher}
Tambet Matiisen, Avital Oliver, Taco Cohen, and John Schulman.
\newblock Teacher-student curriculum learning.
\newblock {\em IEEE transactions on neural networks and learning systems},
  2019.

\bibitem{sweigart2012making}
Albert Sweigart.
\newblock {\em Making Games with Python \& Pygame}.
\newblock CreateSpace North Charleston, 2012.

\end{thebibliography}


\appendix 



\chapter{PyGame Simulation Classes} 

\label{AppendixA} 

In this appendix we provide a full description of our simulation classes, including the classes' variables and methods.

\section{Pygame}

\begin{figure}[th]
	\centering
	\includegraphics[width=10cm]{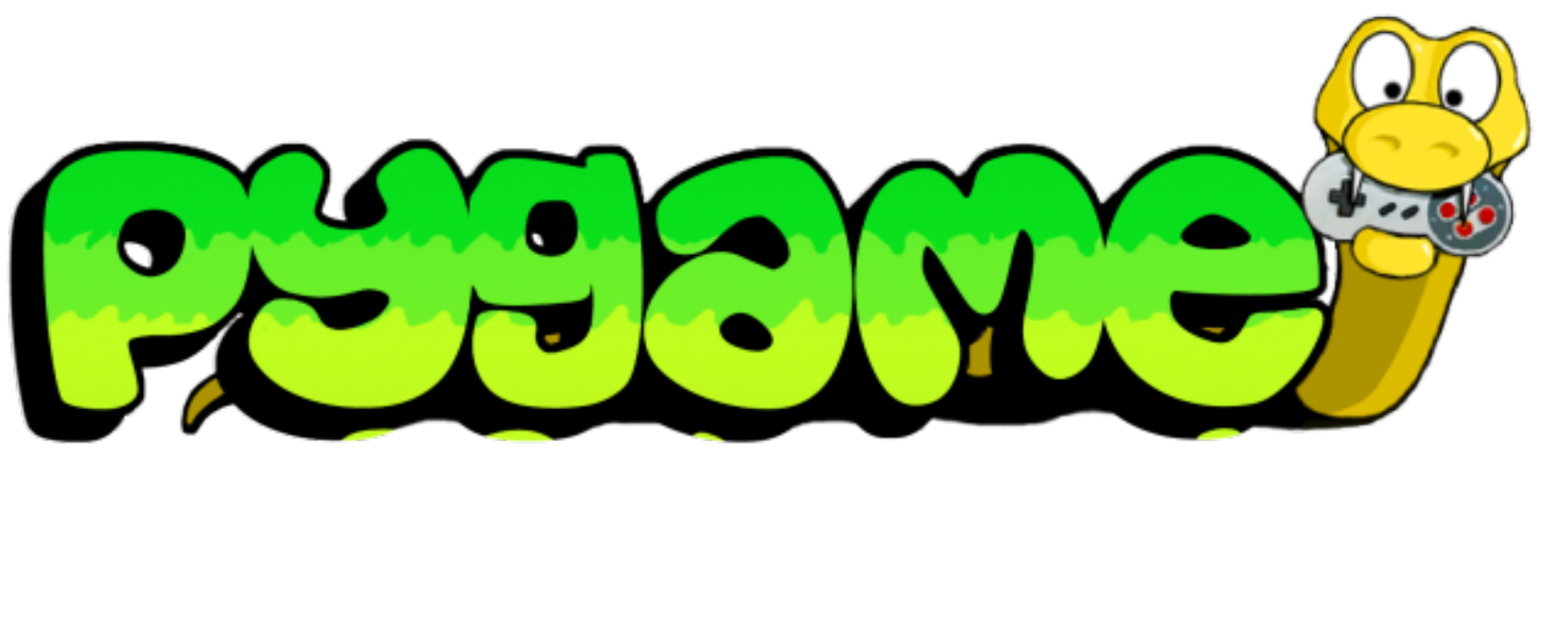}
	\decoRule
	\caption[Pygame's logo]{Pygame's logo \footnotemark}
	\label{fig:pygame}
\end{figure}

All our environments were built in Python using the package Pygame \cite{sweigart2012making} (figure \ref{fig:pygame}). PyGame is an open-source package for building games. Pygame has an object-oriented programming style, where the screen and all objects in the game are python-objects. 
\footnotetext{Source: \quad\url{http://www.pygame.org/docs/logos.html}}
\newpage

\section{Object classes}

\subsection{Ball}

\begin{figure}[th]
	\centering
	\includegraphics[width=3cm]{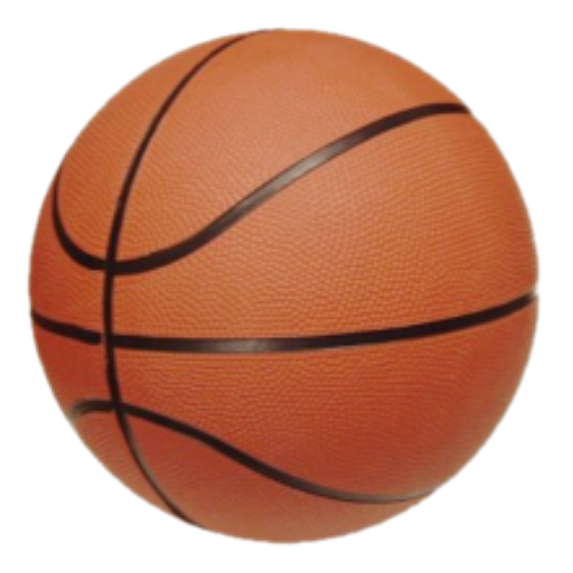}
	\decoRule
	\caption[Ball object]{Ball object: The image of the ball object}
	\label{fig:ball object}
\end{figure}

The class \textit{Ball} is the ball object. This object is represented in the game using a basket-ball image (see figure \ref{fig:ball object}). The ball can be picked and thrown. The class \textit{Ball} also contains all the relevant physics, such as gravity, friction, and bouncing. 
The ball is initialized at a random location on the floor, within the reachable area of the hand.

\subsubsection{Ball Physics}
The ball is initialized on the ground, with no velocity. If the hand (see subsection \ref{ss:Hand}) is holding the ball, it will move in the same velocity as the hand. \\
In a free fall, gravity is applied on the ball, and the ball accelerates towards the ground with a constant acceleration factor of $10\frac{m^2}{s}$.\\
Every time the ball bounces from the floor or the walls, it looses $30\%$ of its velocity.\\
The ball mass is 3 Kilograms and has a kinetic friction factor of 0.1.

\subsubsection{Class variables}
The \textit{Ball}'s variables are as follows:
\begin{itemize}
	\item{\makebox[3.5cm][l]{\textit{img}} The image of the object}
	\item{\makebox[3.5cm][l]{\textit{pos\_col\_left}} The left column of the object}
	\item{\makebox[3.5cm][l]{\textit{pos\_col\_right}} The right column of the object}
	\item{\makebox[3.5cm][l]{\textit{pos\_row\_top}} The top row of the object}
	\item{\makebox[3.5cm][l]{\textit{pos\_row\_bottom}} The bottom row of the object}
	\item{\makebox[3.5cm][l]{\textit{vel\_col}} The velocity over the col axis}
	\item{\makebox[3.5cm][l]{\textit{vel\_row}} The velocity over the row axis}
	\item{\makebox[3.5cm][l]{\textit{scale}} Scaling factor pixels to meter}
	\item{\makebox[3.5cm][l]{\textit{s\_p\_s}} Steps per seconds}
	\item{\makebox[3.5cm][l]{\textit{gameDisplay}} The display object}
	\item{\makebox[3.5cm][l]{\textit{max\_col}} The ball's maximum reachable col}
	\item{\makebox[3.5cm][l]{\textit{max\_row}} The ball's maximum reachable row}
	\item{\makebox[3.5cm][l]{\textit{is\_held}} A Boolean. True if the ball is held by an hand}
	\item{\makebox[3.5cm][l]{\textit{on\_ground}} A Boolean. True if the ball is resting on the floor}
	\item{\makebox[3.5cm][l]{\textit{wall}} The wall's specifications. None if there is no wall.}
	\item{\makebox[3.5cm][l]{\textit{bounce\_count}} A dict counting all type of bouncing (wall/floor etc.).}
\end{itemize}

\subsubsection{Class methods}
The \textit{Ball}'s public methods are as follows:
\begin{itemize}
	\item{\makebox[3.5cm][l]{\textit{move}} Set the velocity for future steps}
	\item{\makebox[3.5cm][l]{\textit{step}} Execute a step}
	\item{\makebox[3.5cm][l]{\textit{display}} Add the ball image to the display}
	\item{\makebox[3.5cm][l]{\textit{hold}} Called when an hand grabs the ball}
	\item{\makebox[3.5cm][l]{\textit{release}} Called when the hand releases the ball}
	\item{\makebox[3.5cm][l]{\textit{center}} Returns the center coordinates of the ball}
	\item{\makebox[3.5cm][l]{\textit{velocity}} Returns the current velocity of the ball}
\end{itemize}
The \textit{Ball}'s private methods are as follows:
\begin{itemize}
	\item{\makebox[4.5cm][l]{\_\_touch\_ground\_\_} Check if the ball touches the ground}
	\item{\makebox[4.5cm][l]{\_\_apply\_gravity\_\_} Apply gravity to the ball's velocity, if necessary}
	\item{\makebox[4.5cm][l]{\_\_apply\_friction\_\_} Apply friction to the ball's velocity, if necessary}
	\item{\makebox[4.5cm][l]{\_\_calculate\_position\_\_} Calculate the new position of the ball}
	\item{\makebox[4.5cm][l]{\_\_border\_collision\_\_} Checks if the ball collides with the wall}
	\item{\makebox[4.5cm][l]{\_\_reverse\_\_} Reverse the velocity after collision}
	\item{\makebox[4.5cm][l]{\_\_gravity\_\_} Calculate the current gravity applied on the ball}
	\item{\makebox[4.5cm][l]{\_\_friction\_\_} Calculate the current friction applied on the ball}
\end{itemize}

\subsection{Ball\_manipulator} \label{ss:Ball manipulator}

The \textit{Ball\_manipulator} object is the same as the ball object but adjusted to the robotic environment. The ball in the robotic environments is initialized in the air; thus, it should not move until held by the manipulator. Furthermore, while the manipulator holds the ball, its movement is nonlinear. Thus the \textit{move} method can also get the next position of the ball, instead of the velocities. 
The ball\_manipulator, like the ball, is initialized at a random location, within a sub-area of the manipulator's workspace. Namely, $ball_\theta \in [75,105] \,, ball_{radius} \in [0.75\times robot\_length \,, robot\_length]$

\subsection{Hand} \label{ss:Hand}

\begin{figure}[h]%
	\centering
	\subfloat[]{{\includegraphics[width=3cm]{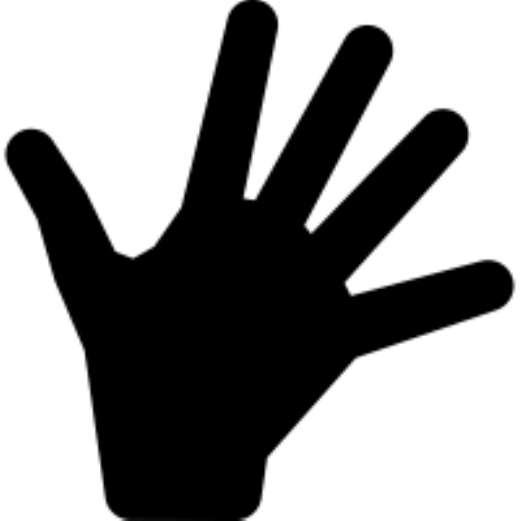} }}
	\hspace{3cm}%
	\subfloat[]{{\includegraphics[width=1.8cm]{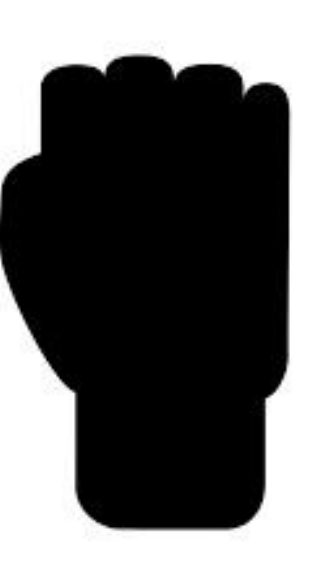} }}%
	\caption[Hand object]{Hand object: The image of the hand object. (A) open hand (B) close hand.}
	\label{fig:hand object}%
\end{figure}

The class \textit{Hand} is the hand object. This object is represented in the game using two hand images - open and close (see figure \ref{fig:hand object}). The agent can move the hand with $x$ and $y$ velocities. The agent can also open or close the hand. If the hand is closed while hovering above the ball, it will catch the ball. When the hand releases the ball, the ball keeps moving with the same velocity as the hand moved before the releasement. The hand movement is bounded to the right half of the screen. The hand is initialized at a random position within its reachable area.

\subsubsection{Class variables}
The \textit{Hand}'s variables are as follows:
\begin{itemize}
	\item{\makebox[3.5cm][l]{\textit{img\_close}} The image of the closed hand}
	\item{\makebox[3.5cm][l]{\textit{img\_open}} The image of the opened hand}
	\item{\makebox[3.5cm][l]{\textit{img}} The image of the current state of the hand}
	\item{\makebox[3.5cm][l]{\textit{pos\_col\_left}} The left column of the object}
	\item{\makebox[3.5cm][l]{\textit{pos\_col\_right}} The right column of the object}
	\item{\makebox[3.5cm][l]{\textit{pos\_row\_top}} The top row of the object}
	\item{\makebox[3.5cm][l]{\textit{pos\_row\_bottom}} The bottom row of the object}
	\item{\makebox[3.5cm][l]{\textit{vel\_col}} The velocity over the col axis}
	\item{\makebox[3.5cm][l]{\textit{vel\_row}} The velocity over the row axis}
	\item{\makebox[3.5cm][l]{\textit{scale}} Scaling factor pixels to meter}
	\item{\makebox[3.5cm][l]{\textit{s\_p\_s}} Steps per seconds}
	\item{\makebox[3.5cm][l]{\textit{gameDisplay}} The display object}
	\item{\makebox[3.5cm][l]{\textit{max\_col}} The ball's maximum reachable col}
	\item{\makebox[3.5cm][l]{\textit{max\_row}} The ball's maximum reachable row}
	\item{\makebox[3.5cm][l]{\textit{ball}} The ball object held by the hand. None if the hand does not hold any ball}
	\item{\makebox[3.5cm][l]{\textit{is\_close}} A Boolean. True if the hand is close}
\end{itemize}

\subsubsection{Class methods}
The \textit{Hand}'s public methods are as follows:
\begin{itemize}
	\item{\makebox[3.5cm][l]{move} Set the velocity for future steps}
	\item{\makebox[3.5cm][l]{step} Execute a step}
	\item{\makebox[3.5cm][l]{close} close the hand. can get ball to grab}
	\item{\makebox[3.5cm][l]{open} open the hand. release the ball if there is any}
	\item{\makebox[3.5cm][l]{display} Add the hand image to the display}
	\item{\makebox[3.5cm][l]{center} Returns the center coordinates of the hand}
	\item{\makebox[3.5cm][l]{velocity} Returns the current velocity of the hand}
\end{itemize}
The \textit{Hand}'s private methods are as follows:
\begin{itemize}
	\item{\makebox[4.5cm][l]{\_\_calculate\_position\_\_} Calculate the new position of the ball}
	\item{\makebox[4.5cm][l]{\_\_border\_collision\_\_} Checks if the hand collides with the walls}
\end{itemize}

\subsection{RoboticManipulator} \label{ss:RoboticManipulator}

\begin{figure}[th]
	\centering
	\includegraphics[width=8cm]{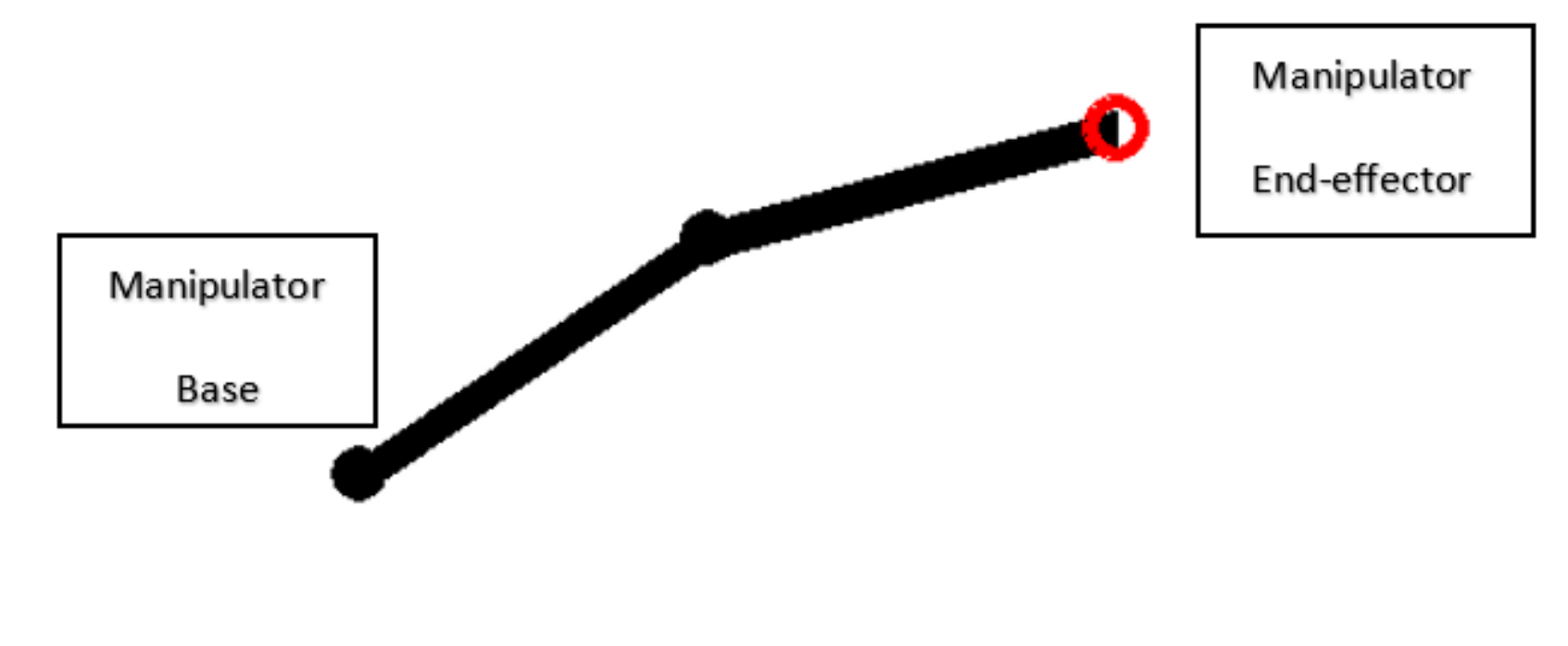}
	\decoRule
	\caption[Manipulator object]{Manipulator object: An image of a 2 DOF manipulator object. The left black circle is the base of the manipulator. The red circle is the end-effector.}
	\label{fig:manipulator object}
\end{figure}

The class \textit{RoboticManipulator} is the manipulator object. This object is represented in the game using a chain of lines and circles. Each link in the manipulator is represented by a line and each joint by a circle. A red circle represents the end-effector (see figure \ref{fig:manipulator object}). If the end-effector is close, the red circle if full. Otherwise, empty. The base of the manipulator is at $33\%$ on the $x$ axis and $50\%$ on the $y$ axis. The total length of the manipulator is 300 pixels, regardless of the number of joints. The agent can control the manipulator's joints' velocities. Like with the hand (\ref{ss:Hand}), the agent can also open or close the end-effector. If the end-effector is closed while hovering above the ball, it will catch the ball. When the end-effector releases the ball, the ball keeps moving with the same velocity as the end-effector moved before at the moment of releasement. For all forward-kinematics' calculations, we wrote an additional script, called \textit{Kinematics}.
The manipulator is initialized with random joints' angles $\theta$ at the range of $[180,360]$.

\subsubsection{Class variables}
The \textit{RoboticManipulator}'s variables are as follows:
\begin{itemize}
	\item{\makebox[3.5cm][l]{\textit{scale}} Scaling factor pixels to meter}
	\item{\makebox[3.5cm][l]{\textit{s\_p\_s}} Steps per seconds}
	\item{\makebox[3.5cm][l]{\textit{gameDisplay}} The display object}
	\item{\makebox[3.5cm][l]{\textit{zero\_point}} The $x$ and $y$ coordinates of the manipulator's base point}
	\item{\makebox[3.5cm][l]{\textit{num\_of\_links}} The DOF of the manipulator}
	\item{\makebox[3.5cm][l]{\textit{link\_length}} The length of all links}
	\item{\makebox[3.5cm][l]{\textit{theta}} The current angles of the joints}
	\item{\makebox[3.5cm][l]{\textit{theta\_dot}} The current joints' velocity}
	\item{\makebox[3.5cm][l]{\textit{pos}} The $x$ and $y$ coordinates of the manipulator's end-effector}
	\item{\makebox[3.5cm][l]{\textit{vel}} The $x$ and $y$ velocities of the manipulator's end-effector}
	\item{\makebox[3.5cm][l]{\textit{ball}} The ball object held by the end-effector. None if the end-effector does not hold any ball}
	\item{\makebox[3.5cm][l]{\textit{is\_close}} A Boolean. True if the end-effector is close}
\end{itemize}

\subsubsection{Class methods}
The \textit{RoboticManipulator}'s public methods are as follows:
\begin{itemize}
	\item{\makebox[3.5cm][l]{move} Set the joints' velocity for future steps}
	\item{\makebox[3.5cm][l]{step} Execute a step}
	\item{\makebox[3.5cm][l]{close} close the end-effector. can get ball to grab}
	\item{\makebox[3.5cm][l]{open} open the hand. release the ball if there is any}
	\item{\makebox[3.5cm][l]{display} Add the hand image to the display}
	\item{\makebox[3.5cm][l]{center} Returns the center coordinates of the hand}
	\item{\makebox[3.5cm][l]{velocity} Returns the current velocity of the hand}
	\item{\makebox[3.5cm][l]{get\_theta} Returns the current joints' angles}
	\item{\makebox[3.5cm][l]{get\_theta\_dot} Returns the current joints' velocities}
\end{itemize}
The \textit{RoboticManipulator}'s private methods are as follows:
\begin{itemize}
	\item{\makebox[4.5cm][l]{\_\_updateTheta\_\_} Updates the joints' angles}
\end{itemize}

\subsection{BlackHole} \label{ss:BlackHole}

\begin{figure}[th]
	\centering
	\includegraphics[width=3cm]{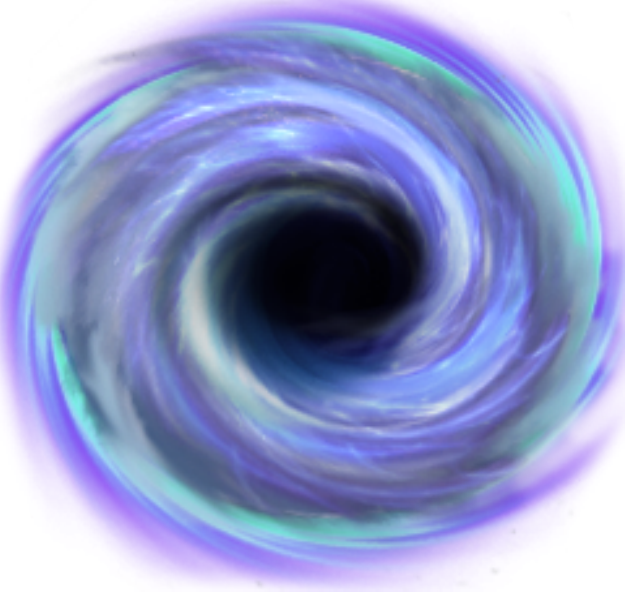}
	\decoRule
	\caption[Black-hole object]{Black-hole object: The image of the target, represented by a black-hole.}
	\label{fig:black-hole object}
\end{figure}

The class \textit{BlackHole} is the target object. This object is represented in the game using a black-hole image (see figure \ref{fig:black-hole object}). The black-hole position is fixed and set at the beginning of each game.

\subsubsection{Class variables}
The \textit{BlackHole}'s variables are as follows:
\begin{itemize}
	\item{\makebox[3.5cm][l]{\textit{img}} The image of the object}
	\item{\makebox[3.5cm][l]{\textit{pos\_col\_left}} The left column of the object}
	\item{\makebox[3.5cm][l]{\textit{pos\_col\_right}} The right column of the object}
	\item{\makebox[3.5cm][l]{\textit{pos\_row\_top}} The top row of the object}
	\item{\makebox[3.5cm][l]{\textit{pos\_row\_bottom}} The bottom row of the object}
	\item{\makebox[3.5cm][l]{\textit{center\_col}} The col coordinate of the goal}
	\item{\makebox[3.5cm][l]{\textit{center\_row}} The row coordinate of the goal}
	\item{\makebox[3.5cm][l]{\textit{gameDisplay}} The display object}
\end{itemize}

\subsubsection{Class methods}
The \textit{BlackHole}'s public methods are as follows:
\begin{itemize}
	\item{\makebox[3.5cm][l]{display} Add the hand image to the display}
	\item{\makebox[3.5cm][l]{goal} Returns goal coordinates}
\end{itemize}


\section{Abstract Classes}
\begin{figure}[th]
	\centering
	\includegraphics[width=14cm]{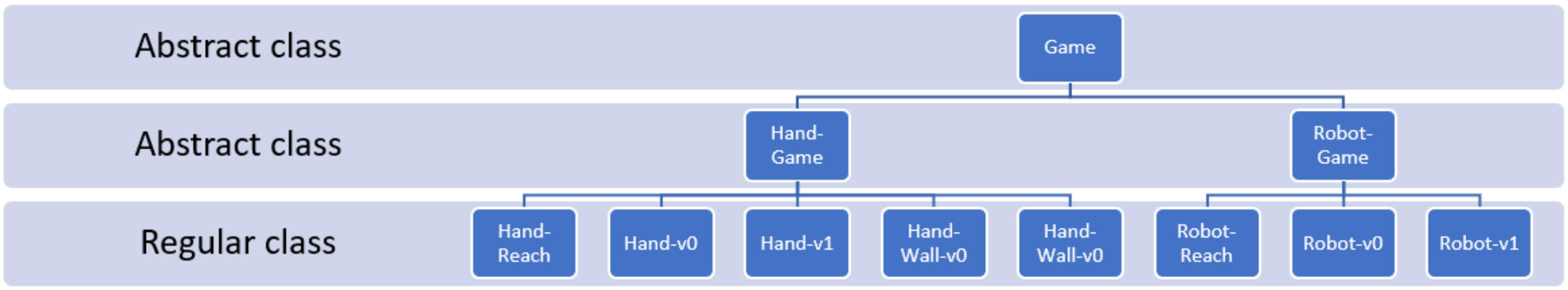}
	\decoRule
	\caption[Environments classes]{Environments architecture}
	\label{fig:environments classes}
\end{figure}

As shown in figure \ref{fig:environments classes}, we built our environments with an object-oriented architecture.
Our environments have the same format as in OpenAI's simulations. That is, all environments have the following methods: \textit{reset}, \textit{step}, \textit{render} and \textit{compute\_reward}. See appendix \ref{AppendixA} for the full classes' description.

\subsection{Game} \label{ss:Game}
The class \textit{Game} is an abstract class, which includes the basic concepts shared by all environments.
\subsubsection{Class variables}
The class \textit{Game}'s variables are as follows:
\begin{itemize}
	\item{\makebox[3.5cm][l]{\textit{clock}} Manage time in the simulation}
	\item{\makebox[3.5cm][l]{\textit{gameDisplay}} The display object to print the game on}
	\item{\makebox[3.5cm][l]{\textit{scale}} Scaling factor pixels to meter}
	\item{\makebox[3.5cm][l]{\textit{to\_render}} If true, render the game}
	\item{\makebox[3.5cm][l]{\textit{ball}} Ball object}
	\item{\makebox[3.5cm][l]{\textit{b\_pos}} Ball position}
	\item{\makebox[3.5cm][l]{\textit{b\_vel}} Ball velocity}
	\item{\makebox[3.5cm][l]{\textit{obs}} The observation dict}
	\item{\makebox[3.5cm][l]{\textit{goal}} The goal's position}
	\item{\makebox[3.5cm][l]{\textit{done}} True when the game is over}
	\item{\makebox[3.5cm][l]{\textit{reward}} the latest reward}
	\item{\makebox[3.5cm][l]{\textit{step\_count}} Counts the agent's steps}
	\item{\makebox[3.5cm][l]{\textit{max\_step}} Maximum steps before the game is over}
	\item{\makebox[3.5cm][l]{\textit{action\_max}} The maximum action's value in the game}
	\item{\makebox[3.5cm][l]{\textit{action\_min}} The minimum action's value in the game}
	\item{\makebox[3.5cm][l]{\textit{action\_dim}} The actions' dimensions}
	\item{\makebox[3.5cm][l]{\textit{observation\_space}} The observation's dimension}
\end{itemize}

\subsubsection{Class methods}
The class \textit{Game}'s public methods are as follows:
\begin{itemize}
	\item{\makebox[3.5cm][l]{\textit{reset}} Reset all the objects in the environment}
	\item{\makebox[3.5cm][l]{\textit{update\_obs}} Update the obs dictionary}
	\item{\makebox[3.5cm][l]{\textit{render}} Rendering the display}
	\item{\makebox[3.5cm][l]{\textit{step}} Execute step according to the agent's action}
	\item{\makebox[3.5cm][l]{\textit{compute\_reward}} Return the reward}
	\item{\makebox[3.5cm][l]{\textit{check\_if\_done}} Return True if game is over}
	\item{\makebox[3.5cm][l]{\textit{action\_sample}} Returns a random action}
	\item{\makebox[3.5cm][l]{\textit{distance}} return the distance between two points}
	\item{\makebox[3.5cm][l]{\textit{goal\_dist}} Returns the goal distribution}\\
\end{itemize}
The class \textit{Game}'s private methods are as follows:
\begin{itemize}
	\item{\makebox[3.5cm][l]{\textit{\_\_get\_reward\_\_}} Calculates the reward}
	\item{\makebox[3.5cm][l]{\textit{\_\_display\_\_}} Set the display}
	\item{\makebox[3.5cm][l]{\textit{\_\_scaler\_\_}} Scales the input to a similar range}
	\item{\makebox[3.5cm][l]{\textit{\_\_reverse\_scaler\_\_}} Un-scale}
\end{itemize}


\subsection{Hand Game} \label{ss:HandGame}
The class \textit{HandGame} is an abstract class that inherits the class \textit{Game} (\ref{ss:Game}). This class includes the objects and methods shared by the Hand environments.\\
Important: In \textbf{all} Hand-environments, the hand movement is bounded to the left half of the screen. Thus, the agent must throw the ball in order to reach the target.
\subsubsection{Class variables}
The class \textit{HandGame} adds several variables:
\begin{itemize}
	\item{\makebox[3.5cm][l]{\textit{hand}} Hand objec}
	\item{\makebox[3.5cm][l]{\textit{h\_pos}} Hand position}
	\item{\makebox[3.5cm][l]{\textit{h\_vel}} Hand velocity}
\end{itemize}

\subsubsection{Class methods}
The class \textit{HandGame}'s public methods are as follows:
\begin{itemize}
	\item{\makebox[3.5cm][l]{\textit{close\_hand}} Close the hand. Check if the hand grabbed a ball}
\end{itemize}


\subsection{Robot Game} \label{ss:RobotGame}
The class \textit{RobotGame} is an abstract class that inherits the class \textit{Game} (\ref{ss:Game}). This class includes the objects and methods shared by the Robot environments. Joint Angles $\theta$ are mapped via a transformation $\phi$ to $\phi(\theta) = (sin(\theta), cos(\theta))$. This representation is better suited to neural networks since it defines angles ($\phi(0) = \phi(360)$) as continuous variables\\ 
In all the experiments with the robotic environments the manipulator had two joints.
\subsubsection{Class variables}
The class \textit{RobotGame} adds several variables:
\begin{itemize}
	\item{\makebox[3.5cm][l]{\textit{num\_of\_links}} The manipulator's number of joints}
	\item{\makebox[3.5cm][l]{\textit{Manipulator}} Manipulator object}
	\item{\makebox[3.5cm][l]{\textit{Theta}} Joints' angles}
	\item{\makebox[3.5cm][l]{\textit{theta\_dot}} Joints' velocities}
	\item{\makebox[3.5cm][l]{\textit{m\_pos}} End-effector position}
	\item{\makebox[3.5cm][l]{\textit{m\_vel}} End-effector velocity}
\end{itemize}

\subsubsection{Class methods}
The class \textit{RobotGame}'s public methods are as follows:
\begin{itemize}
	\item{\makebox[3.5cm][l]{\textit{close\_manipulator}} Close the hand. Check if the hand grabbed a ball}
\end{itemize}

\chapter{Experiments Layout} 

\label{AppendixB} 

In this appendix we provide a full description of our experiments setup, including network's- and algorithm's parameters.

\section{Training algorithm}
All the training was done using the DDPG algorithm with the following parameters:
\begin{center}
 \begin{tabular}{||c|c||} 
 \hline
 hyper-parameters & value \\ [0.5ex] 
 \hline\hline
 discount factor ($\gamma$) & 0.98\\ 
 \hline
 target-networks smoothing ($\tau$) & 7 \\
 \hline
 buffer size & 1e6 \\
 \hline
 $\epsilon$ initial value & 1 \\
 \hline
 $\epsilon$ decay rate & 0.95 \\
 \hline
 $\epsilon$ final value & 0.05 \\
 \hline
\end{tabular}
\end{center}
For exploration we used a decaying epsilon-greed policy:

\begin{equation*}
a =
\begin{cases*}
  a^{*} & with probability $1-\epsilon$ \\
  a^{*}+\mathcal{N}(0,\,I\cdot\sigma)   & with probability $0.8\cdot\epsilon$ \\
  rand(a) & with probability $0.2\cdot\epsilon$
\end{cases*}
\end{equation*}
Where $\sigma=0.05\cdot action\_range$ and $\epsilon$ decays at the beginning of every epoch.\\
For experience replay we used \textit{prioritize experience replay} \cite{schaul2015prioritized}.

\section{Neural networks}
We used the same neural network layout for all the experiments:
\subsection{Actor:}
\begin{center}
 \begin{tabular}{||c|c|c|c|c|c||} 
 \hline
 layer & size & type & activation & BN  & additional info\\ [0.5ex] 
 \hline\hline
input&input dim&Input&relu&No&No \\
\hline
hidden 1&64&FC&relu&No&No \\
\hline
hidden 2&64&FC&relu&No&No \\
\hline
hidden 3&64&FC&relu&No&No \\
\hline
output&action dim&FC&tanh&No&No \\
\hline
\end{tabular}
\end{center}
\begin{center}
 \begin{tabular}{||c|c||} 
 \hline
 hyper-parameter & value \\ [0.5ex] 
 \hline\hline
 learning rate & 0.001 \\
 \hline
 gradient clipping & 3 \\
 \hline
 batch size & 64 \\ 
 \hline
\end{tabular}
\end{center}

\subsection{Critic:}
\begin{center}
 \begin{tabular}{||c|c|c|c|c|c||} 
 \hline
 layer & size & type & activation & BN  & additional info\\ [0.5ex] 
 \hline\hline
input&input dim&Input&relu&Yes&No \\
\hline
hidden 1&64&FC&relu&Yes&concat the layer to the action \\
\hline
hidden 2&64&FC&relu&Yes&No \\
\hline
hidden 3&64&FC&relu&Yes&No \\
\hline
output&1&FC&linear&Yes&No \\
\hline
\end{tabular}
\end{center}
\begin{center}
 \begin{tabular}{||c|c||} 
 \hline
 hyper-parameter & value \\ [0.5ex] 
 \hline\hline
 learning rate & 0.001 \\
 \hline
 gradient clipping & 3 \\
 \hline
 batch size & 64 \\ 
 \hline
\end{tabular}
\end{center}

\chapter{Curriculum HER} 

\label{AppendixC} 

In this appendix we sketch a proof for the time complexities of traditional RL algorithms, HER and Curriculum-HER on tasks with a sparse reward function. For this purpose consider the following toy problem: An agent needs to go from an initial state $s_1$ to a goal $s_n$, which is $n$ steps away (see figure \ref{fig:toy problem}).
\begin{figure}[H]
    \centering
    \tcbox{
    \begin{tikzpicture}[auto,node distance=8mm,>=latex,font=\small]
        \tikzstyle{round}=[thick,draw=black,circle]
        \node[round]  at (0, 0) (s1) {$s_1$};
        \node[round]  at (1.5, 0) (s2) {$s_2$};
        \node[round]  at (3, 0) (s3) {$s_3$};
        \node[round]  at (4.5, 0) (s4) {$s_4$};
        \node[round]  at (7, 0) (sn) {$s_n$};
        \node at ($(s4)!.5!(sn)$) {\ldots};
        
        \draw[->] (s1) to [out=315,in=225] (s2);
        \draw[->] (s2) to [out=315,in=225] (s3);
        \draw[->] (s3) to [out=315,in=225] (s4);
        \draw[->] (s2) to [out=135,in=45] (s1);
        \draw[->] (s3) to [out=135,in=45] (s2);
        \draw[->] (s4) to [out=135,in=45] (s3);

    \end{tikzpicture}
    }
    \caption[Toy problem]{Toy problem}
    \label{fig:toy problem}
\end{figure}
\noindent For simplicity we make the following assumptions:
\begin{itemize}
\item \textit{premise 1:} Under a random policy, the agent goes forward with probability $p$.
\item \textit{premise 2:} At each episode, the agent will first go the best known target, and will continue with a random policy.
\item \textit{premise 3:} The agent can learn how to reach a goal from a single successful example.
\item \textit{premise 4:} There a limited number of steps, hence the agent cannot waist moves.
\end{itemize}

\section{Time complexity of traditional RL algorithms} \label{s:Time complexity of traditional RL algorithms}
\begin{figure}[th]
    \centering
    \tcbox{
    \begin{tikzpicture}[auto,node distance=8mm,>=latex,font=\small]
        \tikzstyle{round}=[thick,draw=black,circle]
        \node[round]  at (0, 0) (s1) {$s_1$};
        \node[round]  at (1.5, 0) (s2) {$s_2$};
        \node[round]  at (3, 0) (s3) {$s_3$};
        \node[round]  at (4.5, 0) (s4) {$s_4$};
        \node[round]  at (7, 0) (sn) {$s_n$};
        \node at ($(s4)!.5!(sn)$) {\ldots};
        
        \draw[->] (s1) to [out=315,in=225] (s2);
        \draw[->] (s2) to [out=315,in=225] (s3);
        \draw[->] (s3) to [out=315,in=225] (s4);
        \draw[->] (s2) to [out=135,in=45] (s1);
        \draw[->] (s3) to [out=135,in=45] (s1);
        \draw[->] (s4) to [out=135,in=45] (s1);

    \end{tikzpicture}
    }
    \caption[Toy problem - reset state]{Toy problem - reset state}
    \label{fig:toy problem - reset state}
\end{figure}
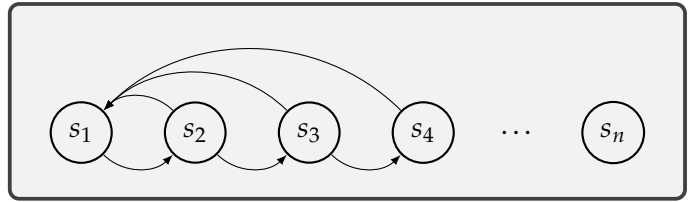

Under \textit{premise 4}, the agent cannot make any mistakes. Thus, the game can be treated as a \textit{reset-state game} (from \cite{koenig1993complexity}). In order to solve the game, the agent must choose the right action at every state (see figure \ref{fig:toy problem - reset state}). Under \textit{premise 3}, the agent needs to reach the goal once in order to learn the task. Under \textit{premise 1}, in each state, the agent goes forward with a probability of $p$. Hence, the probability of the agent to reach the goal is $p^n$. Thus, the time complexity of solving the task is $\frac{1}{p}^n$.

\section{Time complexity of HER} \label{s:Time complexity of HER}
By learning from virtual goals, the agent memorizes every state it has seen.
Under \textit{premise 2}, the agent can then reach the known states, and start exploring from there. Therefore, in order to reach the goal, the agent can learn state by state. Thus, the time complexity of solving the task using HER is the sum of the time complexities of all states: $\sum_{i=1}^n \frac{1}{p_i}$. That is, under \textit{premise 1}, $n\cdot\frac{1}{p}$.

\section{Time complexity of Curriculum-HER}\label{s:Time complexity of Curriculum-HER}
In a multilayered task, the agent must solve all sub-tasks in order manipulate the object. In other words, the time complexity of getting from $s_1$ to $s_2$ is equal to the time complexity of all the sub-tasks. Since HER is not applied on the sub-task level, their time complexity is exponential in their total state-space-complexity (see \ref{s:Time complexity of traditional RL algorithms}).
By applying HER on each sub-task level, the time complexity of getting from $s_1$ to $s_2$ is instead polynomial in the sub-tasks' total state-space complexity (\ref{s:Time complexity of HER}).

\end{document}